\documentclass[10pt]{article} 
\usepackage[accepted]{tmlr}

\usepackage{hyperref}
\usepackage{url}
\usepackage{enumitem}

\usepackage[dvipsnames,svgnames]{xcolor}

\hypersetup{
  colorlinks=true,
  linkcolor=magenta,
  citecolor=ForestGreen,
  urlcolor=blue!80!black,
  filecolor=blue!80!black,
  }

\usepackage{dsfont}

\usepackage{colortbl} 
\usepackage{graphicx}
\usepackage{tikz}
\usepackage{tikz-cd}
\usepackage{hf-tikz}
\usepackage{pgfplots} 
\pgfplotsset{compat=1.17} 
\pgfplotsset{
        table/search path={figures/drawings},
    }
\usetikzlibrary{fadings}
\usetikzlibrary{shapes, arrows, fit, backgrounds, arrows.meta}

\usetikzlibrary{matrix}
\usetikzlibrary{shadows.blur}
\usetikzlibrary{patterns, tikzmark}
\usetikzlibrary{decorations.pathreplacing, calc, decorations.markings,}
\usetikzlibrary{positioning}

\usepgfplotslibrary{groupplots}
\usepgfplotslibrary{patchplots}
\definecolor{bluegray}{rgb}{0.4, 0.6, 0.8}
\definecolor{bluebell}{rgb}{0.64, 0.64, 0.82}
\definecolor{etonblue}{rgb}{0.59, 0.78, 0.64}
\definecolor{junglegreen}{rgb}{0.16, 0.67, 0.53}
\definecolor{bg}{gray}{0.97}
\definecolor{olive}{rgb}{0.6, 0.6, 0.2}
\definecolor{sand}{rgb}{0.8666666666666667, 0.8, 0.4666666666666667}
\definecolor{wine}{rgb}{0.5333333333333333, 0.13333333333333333, 0.3333333333333333}
\definecolor{deblue}{RGB}{11,132,147}
\definecolor{ocra}{RGB}{204, 119, 34}

\pgfplotsset{%
            mesh line legend/.style={legend image code/.code=\meshlinelegend#1},%
}
\makeatletter
\long\def\meshlinelegend#1{%
    \scope[%
        #1,
        /pgfplots/mesh/rows=1,
        /pgfplots/mesh/cols=4,
        /pgfplots/mesh/num points=,
        /tikz/x={(0.44237cm,0cm)}, 
        /tikz/y={(0cm,0.23932cm)},
        /tikz/z={(0.0cm,0cm)},
        scale=0.4,
    ]
    \let\pgfplots@metamax=\pgfutil@empty
    \pgfplots@curplot@threedimtrue

    \pgfplotsplothandlermesh
    \pgfplotstreamstart

    \def\simplecoordinate(##1,##2,##3){%
        \pgfmathparse{1000*(##3)}%
        \pgfmathfloatparsenumber\pgfmathresult
        \let\pgfplots@current@point@meta=\pgfmathresult
        \pgfplotstreampoint{\pgfqpointxyz@orig{##1}{##2}{##3}}%
    }%

    \pgfplotsforeachungrouped \x in {0,...,\pgfkeysvalueof{/pgfplots/samples}}{
        \pgfmathsetmacro\y{\x/\pgfkeysvalueof{/pgfplots/samples}}
        \pgfmathsetmacro\x{\x/\pgfkeysvalueof{/pgfplots/samples}*3}
        \simplecoordinate(\x,0,\y)
    }

    \pgfplotstreamend
    \pgfusepath{stroke}
    \endscope
}%
\makeatother

\usepackage{wrapfig}
\usepackage{booktabs} 
\usepackage{microtype}
\usepackage{graphicx}

\usepackage{amsmath, amssymb, amsfonts, mathtools, amsthm}
\usepackage{physics}
\usepackage{cancel}
\usepackage{accents}
\usepackage{bm}
\usepackage{bbm}  
\usepackage{float}

\usepackage{xcolor}
\usepackage{multirow}
\usepackage{makecell}  
\usepackage{newfloat}
\usepackage{listings}
\usepackage[font=footnotesize]{caption} 
\usepackage{subcaption}
\usepackage{enumitem}

\usepackage{pifont} %
\newcommand{\cmark}{\ding{51}}%
\usepackage[capitalize,noabbrev,nameinlink]{cleveref}

\theoremstyle{plain}

\theoremstyle{definition}

\theoremstyle{remark}

\newcommand{\rebuttal}[1]{{#1}}

\newcommand{\tmlr}[1]{{#1}} 

\def \our{\textsc{RouteFinder}}
\def \papertitle{\our{}: Towards Foundation Models for Vehicle \\ Routing Problems}

\title{\papertitle{}}

\author{Federico Berto\thanks{Equal contributions. $^\ddagger$Authors are members of the AI4CO open research community.}$^{~1,5}$, Chuanbo Hua$^{*1,5}$, Nayeli Gast Zepeda$^{*2}$, André Hottung$^{2}$, Niels A. Wouda$^{3}$, Leon Lan$^{4}$, Junyoung Park$^{1}$, 
Kevin Tierney$^{2}$, Jinkyoo Park$^{1,5}$
\AND \addr $^1$KAIST ~ $^2$Bielefeld University ~ $^3$Rotterdam School of Management ~ $^4$VU Amsterdam ~ $^5$Omelet ~ AI4CO$^\ddagger$ \AND
}

\def\month{09}  
\def\year{2025} 
\def\openreview{\url{https://openreview.net/forum?id=QzGLoaOPiY}} 

\begin{document}

\maketitle




\begin{abstract}
This paper introduces \our{}, a comprehensive foundation model framework to tackle different Vehicle Routing Problem (VRP) variants. Our core idea is that a foundation model for VRPs should be able to represent variants by treating each as a subset of a generalized problem equipped with different attributes. We propose a unified VRP environment capable of efficiently handling any combination of these attributes. The \our{} model leverages a modern transformer-based encoder and global attribute embeddings to improve task representation. Additionally, we introduce two reinforcement learning techniques to enhance multi-task performance: mixed batch training, which enables training on different variants at once, and multi-variant reward normalization to balance different reward scales. Finally, we propose efficient adapter layers that enable fine-tuning for new variants with unseen attributes. Extensive experiments on \rebuttal{48} VRP variants show \our{} outperforms recent state-of-the-art learning methods. Our code is publicly available at \url{https://github.com/ai4co/routefinder}.
\end{abstract}

\section{Introduction} \label{sec:introduction}


Vehicle Routing Problems (VRPs) are an important class of Combinatorial Optimization (CO) problems that have received much attention in Operations Research (OR) and Computer Science.
Since the VRP is an NP-hard problem, finding an optimal solution by exhaustively exploring the solution space is not possible for large instances.
Instead, heuristic methods that quickly generate good (but possibly suboptimal) solutions are commonly used. The OR community has developed many heuristics over the years, including the well-known Lin-Kernighan-Helsgaun (LKH) heuristic~\citep{helsgaun2017extension}, Fast Iterated Local Optimization (FILO)~\citep{accorsi2021fast, accorsi2024routing} and Hybrid Genetic Search (HGS)~\citep{vidal2022hybrid, wouda2024pyvrp}.
While these algorithms deliver state-of-the-art results for certain VRP variants, they often require expert knowledge and careful adaptation to be effectively applied in practice.
Recently, Neural Combinatorial Optimization (NCO) approaches have been developed to solve CO problems. By leveraging deep learning, these approaches seek to learn  from data, potentially providing more flexible and scalable solutions \citep{kool2018attention, hottung2019neural, kwon2020pomo, kim2022sym,  berto2024rl4co, hottung2024polynet}.

Similar to how the developments in natural language processing have resulted in Large Language Models (LLMs), research efforts in solving CO problems through machine learning are also trending toward foundation models \citep{liu2024_llm_gls, ye2024reevo, liu2024multi, zhou2024mvmoe}. However, despite the recent progress made in learning VRP variants, there is a lack of a unified approach that provides a platform for effectively finetuning unseen variants \citep{lin2024cross}. Such a foundation model for VRPs would have important implications for real-world applications as it can be easily adapted to new business requirements (constraints) outside of the training distribution. 

In this work, we introduce \our{}, a comprehensive foundation model framework for solving VRPs. We summarize our key contributions as follows:

\begin{itemize}
    \item We introduce a general framework to solve different VRP variants via a \textit{unified VRP environment} that can handle any number of attributes.
    \item We propose a modern \textit{Transformer-based encoder} and introduce \textit{Global Attribute Embeddings} to enable the model to better understand and differentiate between VRPs.
    \item We introduce two novel reinforcement learning techniques, \textit{Mixed Batch Training} and \textit{Multi-Variant Reward Normalization}, to ensure stable and effective training across multiple VRP variants.
    \item We present \textit{Efficient Adapter Layers}, a lightweight yet powerful mechanism for finetuning pre-trained \our{} models to tackle new variants with previously unseen attributes.
\end{itemize}

We evaluate \our{} through extensive experiments on \rebuttal{48} VRP variants, i.e., three times as many as previous works, assessing the impact of each novel component on performance. \our{} significantly outperforms recent multi-task learning models by reducing optimality gaps by more than $10\%$ across all variants.

\section{Related Works} \label{sec:relatedWork}

\paragraph{Neural combinatorial optimization for VRPs}
NCO has emerged as a pivotal solution approach for VRPs, leveraging advancements in machine learning and neural network architectures \citep{bengio2021machine, peng2021graph, mazyavkina2021reinforcement, bogyrbayeva2022learning}. While several works have explored enhancing exact solvers for other CO problems with deep learning \citep{gasse2019exact,prouvost2020ecole}, these are not generally scalable for real-time complex VRPs \citep{wu2024neural,kim2024neural}, which instead usually employ heuristics in practice. In this work, we aim to learn heuristics for fast solution generation following the seminal work of \citet{vinyals2015pointer} who paved the way in applying NCO to VRPs, further developed by \citet{bello2016neural} and \citet{nazari2018reinforcement}.
Subsequent works, including the transformer-based encoder with self-attention of \citet{kool2018attention} and the training methods of POMO \citep{kwon2020pomo} and Sym-NCO \citep{kim2022sym}, have significantly enhanced solution quality.
These advancements have been complemented by novel training algorithms, including learning with (partial) problem re-encoding at each step \citep{bdeir2022attention, drakulic2024bq, luo2024neural, luo2024self} and population-based approaches \citep{grinsztajn2024winner, chalumeau2024combinatorial,hottung2024polynet}.
Despite this progress, challenges remain in the form of requiring manual tuning for inductive bias and the need for problem-specific models which impact deployment and generalizability \citep{liu2023good, thyssens2023routing}.
The field has also explored non-autoregressive construction methods that predict promising edges \citep{joshi2020learning, fu2021generalize, kool2022deep, sun2024difusco}, improvement methods that iteratively refine solutions through local adjustments \citep{hottung2019neural, ma2021learning, ma2022efficient, ma2024learning,hottung2025neural}, and test-time adaptation methods \citep{hottung2021efficient, choo2022simulation} which allow for solution improvement given larger time budgets.
Recent works additionally explore alternative ways of solving VRPs, such as learning heuristics for Ant Colony Optimization \citep{ye2024deepaco, kim2024gfacs_ant_colony_optimization} and divide-and-conquer \citep{kim2021learning,  li2021learning, hou2022generalize, ye2023glop, chen4679437extnco, zheng2024udc}.

\paragraph{Multi-task learning for VRPs} 
In this work, we develop a unified VRP solver to solve multiple tasks that can be efficiently fine-tuned to new ones. Due to its promise, multi-task learning for VRPs has garnered much attention recently. \citet{wang2023efficient} introduces a multi-armed bandit method that solves several VRP variants with limited training budget. \citet{lin2024cross} proposes training a \textit{backbone} model (i.e., deep layers) for VRPs that can then be adapted via low-dimensional layers, such as linear projections, to fine-tune different problems efficiently. \citet{drakulic2024goal} propose a multi-task model for CO problems trained via supervised learning, similar to Large Language Models (LLMs). \citet{jiang2024unco} introduce a method to transfer different problems to the embedding space via textual description through an LLM. Most related to this work are the works of \citet{liu2024multi} and \citet{zhou2024mvmoe}, which use attribute composition \citep{ruis2021independent} to achieve (zero-shot) generalization on several VRP variants. \citet{liu2024multi} builds on the Reinforcement-Learning-based POMO \citep{kwon2020pomo}, on top of which \citet{zhou2024mvmoe} employ a mixture-of-experts model to improve generalization.

\section{Preliminaries} \label{sec:background}

\subsection{Vehicle Routing Problems}
\label{sec:vrp_features}

We formulate the classic VRP, the foundation for more complex variants, on a graph $G = (N, E)$, where \rebuttal{$N = \{0, \dots, m-1, m, \dots, m + n - 1\}$ represents the nodes, with $N_d = \{0, \dots, m-1 \}$ denoting the $m$ depots ($m=1$ for the classic VRP) and $N_c = \{ m, \dots, m+n-1 \}$ denoting the $n$ customers.} The edges $E$ connect pairs of nodes, and each edge $(i, j) \in E$ has a travel cost $c_{ij}$ (e.g., distance or travel duration). 
Vehicles depart from the depot to serve each customer exactly once and then return to the depot(s), while minimizing the total travel cost.
\begin{figure*}[h!]
    \centering
    \includegraphics[width=\linewidth]{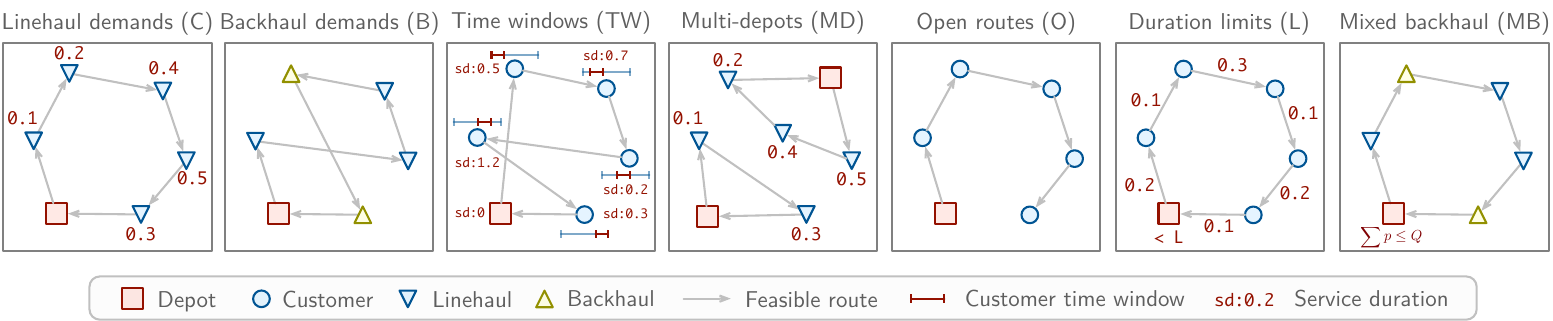}
    \caption{
    \rebuttal{VRP attributes. Linehaul demands (C), backhaul demands (B), time windows (TW), and multi-depot (MD) are \textit{node attributes}, whereas open routes (O), duration limits (L), and mixed backhaul (MB) mode are \textit{global attributes}. Attribute combinations can define new VRP variants.}
    }
    \label{fig:vrp_features}
\end{figure*}
Following~\cite{vidal2014uhgs}, we consider a collection of VRP \textit{variants} that extend the classic VRP by one or more \textit{attributes} (see \cref{fig:vrp_features}). This results in a rich set of routing problems with practical relevance.
Each of these variants offers a unique generalization task for \our{}.
We describe the attributes in the following, separating them into \textit{node attributes} and \textit{global attributes}. \cref{table:vrp_variants_asymmetric} in \cref{append:unified-vrp-env-details} provides a list of all \rebuttal{48} VRP variants considered in this work.

\subsubsection*{Node attributes}

\paragraph{Demand and Vehicle Capacity (C)} [$q \in [0, Q]$]: Every customer $i \in N_c$ has a linehaul demand $q_i$ and is serviced by vehicles with fixed capacity $Q > 0$. The total customer demand per vehicle cannot exceed its capacity at any point. 

\paragraph{Backhauls (B)} [$p \in [0, Q]$]:
Backhauls generalize demand to also account for return shipments. Customers are either linehaul or backhaul customers.
Linehaul customers require delivery of a demand $q_i$ that needs to be transported from the depot to customer $i$ (as in the CVRP), whereas backhaul customers need a pickup of an amount $p_i$ that is transported from the client back to the depot.
Vehicles can serve a combination of linehaul and backhaul customers on a single route, but any linehaul customers must precede the backhaul customers on the route.
An application with returnable bottles is presented in~\cite{ropke2006vrpb}. 

\paragraph{Time Windows (TW)} [$e, s, l  \in [0, T]^3$]:
Every customer $i \in N_c$ has a time window $[e_i, l_i]$ during which service must begin.
Service takes $s_i$ time.
The depot has a time window $[e_0, l_0] = [0, T]$ and a service duration of $s_0 = 0$.
Vehicles must reach node $i$ before $l_i$, but any early arrivals must wait at node $i$ until $e_i$ before service may start.

\subsubsection*{Global attributes}

\paragraph{Open Routes (O)} [$o \in \{0, 1\}$]: Vehicles are not required to return to the depot after serving all customers.
Open routes can be found in applications with third-party drivers, who are often only compensated until they have completed their last delivery~\citep{li2007ovrp}.

\paragraph{Duration Limits (L)} [$l \in [0, L]$]: Imposes a limit on the total travel duration (or length) of each route, balancing the workload across vehicles, and, e.g., limiting the duration of the work day. 

\rebuttal{\paragraph{Mixed Backhauls (MB)} [$\mu \in \{0, 1 \}$]:}
Relaxes the strict precedence constraint of linehaul customers preceding backhaul customers: with mixed backhauls, linehaul and backhaul customers may be mixed along a route.  The vehicle's capacity must still be respected at any point along the route. Since both the current linehaul and backhaul demands must be tracked per vehicle, this variant requires careful planning.

\paragraph{Multi-depot (MD)} [$m > 1$]: Generalizes single-depot ($m=1$) variants to multiple depot nodes $m>1$ from which vehicles can start their tour. Each vehicle must return to its starting depot. This variant requires decisions about depot-to-customer assignments, making the problem more realistic for organizations operating from multiple facilities \citep{karakativc2015survey}. 

\subsection{Learning Neural Solvers for VRPs}
\label{sec:learning_neural_solvers}

\paragraph{Solving VRPs using Autoregressive Sequence Generation}
Autoregressive (AR) methods address CO problems by constructing solutions sequentially. The process begins with encoding the problem instance $\bm{x}$ (e.g., node and global attributes) using a trainable encoder $f_\theta$  that maps $\bm{x}$ to an embedding $\bm{h} = f_\theta(\bm{x})$. The solution $\bm{a}$ is then decoded based on  $\bm{h}$ through a series of actions, where each action determines the next step in the solution based on the current partial sequence. This is achieved using a decoder $g_\theta$. The encoding and decoding process are formalized as:
\begin{subequations}
\label{eq:encoding_decoding}
    \begin{align}
        \pi_\theta(\bm{a}|\bm{x}) &\triangleq \prod_{t=1}^{T-1} g_\theta(a_{t} | a_{t-1}, ... ,a_0, \bm{h}),
    \end{align}
\end{subequations}
where $\bm{a} = (a_1, ..., a_T)$ represents a feasible solution to the CO problem, $T$ the steps in solution construction, and $\pi_\theta$ the stochastic solver mapping problem instance $\bm{x}$ to $\bm{a}$.

\paragraph{Training VRP Solvers via Reinforcement Learning}
The solver $\pi_\theta$ can be trained using either supervised learning (SL) or reinforcement learning (RL). This paper focuses on RL due to its ability to train solvers independently of the availability of optimal solutions. Under the RL framework, the training objective for neural combinatorial optimization solvers is defined as:
\begin{equation}
\label{eq:training_obj}
    \theta^{*} = \underset{\theta}{\text{argmax}}
    \left[
    \mathbb{E}_{\bm{x} \sim P(\bm{x})}\left[\mathbb{E}_{a \sim \pi_\theta(\bm{a}|\bm{x})}[R(\bm{a},\bm{x})]\right]
    \right],
\end{equation}
where $P(\bm{x})$ is the distribution of problem instances, and $R(\bm{a}, \bm{x})$ represents the reward (i.e., the negative cost), associated with the solution $\bm{a}$ for the given $\bm{x}$. The above training problem can be tackled using various RL algorithms such as REINFORCE and its modern variants \citep{kool2018attention, kwon2020pomo}.

\section{The \our{} Recipe} 
\label{sec:method}

\begin{figure*}
    \centering
    \includegraphics[width=0.95\linewidth]{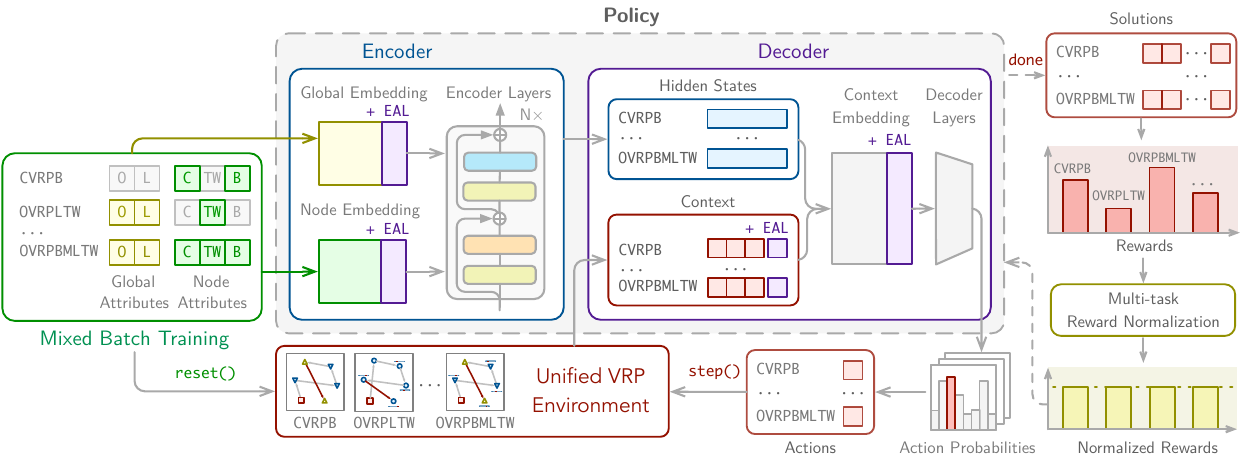}
    \caption{\our{} overview. \rebuttal{The unified VRP environment is used for data generation and solution construction (\cref{sec:vrp_env}). Our Transformer-based encoder (\cref{subsec:transformer-architecture}) processes node and global embeddings (\cref{subsec:gloabal-attribute-embeddings}) of problem instances. During training, we sample multiple variants in the same batch (\cref{subsec:mixed-batch-training}) whose multi-task reward is then normalized (\cref{sec:reward-normalization}). Efficient Adapter Layers (EAL) are employed for efficient fine-tuning to new variants (\cref{subsec:efficient-adapter-layers-eal}). }}
    \label{fig:overview}
\end{figure*}

\our{} leverages attribute composition from \citet{liu2024multi, zhou2024mvmoe} to solve multiple VRP variants. 
We treat different variants of the VRP as combinations of fundamental attributes (\cref{sec:vrp_features}) and use a common network to learn their representations. We go further than previous works and consider different combinations of attributes within training batches (see \cref{subsec:mixed-batch-training}). 
\cref{fig:overview} provides an overview of \our{}'s architecture.

\subsection{Unified VRP Environment}
\label{sec:vrp_env}

In previous works proposing multi-task learning for VRPs, like \mbox{MTPOMO} \citep{liu2024multi} and MVMoE \citep{zhou2024mvmoe}, instance variants (CVRP, VRPTW, etc.) are sampled out of the set of available variants during training. Every instance within that batch is then of the same problem category. This can bias the optimization at each gradient step toward a specific task, potentially hindering stable and effective training for a foundation model. We thus propose to learn across problems throughout training and include instances of various variants within each training batch. 
 
We define an environment capable of modeling all of the previously discussed VRP attributes (see \cref{sec:vrp_features}) simultaneously. Essentially, we build an \rebuttal{MDOVRPMBLTW environment: a multi-depot open route vehicle routing problem with linehauls, (mixed) backhauls, distance limit, and time windows.}
The environment supports subsets of the \rebuttal{MDOVRPMBLTW} by turning attributes ``on'' or ``off''. 
For example, if an instance does not have time window constraints, the time windows attribute of each customer is set to $[0, \infty]$, rendering them irrelevant during solution construction.
This modular attribute composition allows us to model up to \rebuttal{48} different problem types with one single environment. This approach can be easily extended, e.g., by including different location sampling mechanisms and new constraints, allowing for even more future problem variants to be modeled within the same environment.

\begin{wrapfigure}[13]{r}{0.45\textwidth}
\vspace{-19.4mm}
  \centering
  \includegraphics[width=\linewidth]{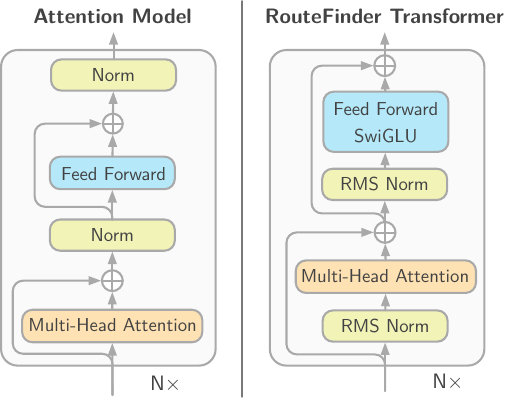}
  \caption{Attention model structure v.s. \our{} transformer structure.}
  \label{fig:routefinder-transformer}
\end{wrapfigure}

\subsection{Model} 
\label{sec:routefinder_model}

\subsubsection{Transformer-based Architecture}
\label{subsec:transformer-architecture}

The \our{} transformer encoder architecture shown in \cref{fig:routefinder-transformer} introduces key enhancements to the Attention Model (AM) from \citet{kool2018attention}, the de-facto standard in recent works \citep{liu2024multi, zhou2024mvmoe}. 
%
%
Firstly, the \our{} transformer encoder employs Root Mean Square (RMS) normalization \citep{zhang2019root}, improving stability and training speed. 
Secondly, we transition from post-norm to pre-norm in the transformer layers, applying normalization before the residual connections, which enhances gradient flow and promotes faster convergence \citep{jiang2024pre}. Thirdly, \our{} uses a Feed Forward SwiGLU, \citep{shazeer2020glu}, an extension of the Gated Linear Unit (GLU) \citep{dauphin2017language}, instead of the AM's ReLU-based feed-forward network. This enhances the model's capacity to capture complex relationships in the data. Finally, we employ FlashAttention \citep{dao2022flashattention, dao2023flashattention} in the Multi-Head Attention layer of all models to enhance overall performance. These improvements build on recent advances in foundation models in areas such as language modeling and biology \citep{dubey2024llama, nguyen2024sequence}. 
By building on modern architectures, we create a robust foundation model for VRPs.
Further details on architecture and modeling are provided in \cref{append:model}.

\subsubsection{Global Attribute Embeddings}
\label{subsec:gloabal-attribute-embeddings}

Global attributes as outlined in \cref{sec:vrp_features} are essential for modeling VRPs. For instance, given an open (O) attribute, the solver may find optimal routes that do not necessarily loop back to the starting depot. Previous multi-task learning models for VRPs \citep{liu2024multi, zhou2024mvmoe} project such features on the shallow decoder as dynamic features. However, such a design can be suboptimal, since the deep transformer layers carry out most of the learning and can enable effective attribute mixing, which is essential for understanding a (new) problem.  We therefore design Global Attribute Embeddings for effective problem representation, which incorporate problem variants and help the deep layers understand which problem is being faced.  Global attributes $\phi_0, \dots, \phi_k$ are projected via a projection layer:
\begin{equation}
    h_g^0 = f_\theta([\phi_0, \dots, \phi_k]), \quad f_\theta : \mathbb{R}^k \rightarrow \mathbb{R}^d 
\end{equation}
into $d$-dimensional space. Given our unified VRP representation, some attributes, such as the duration limit $l$ for unconstrained VRPs, might be $\infty$. These attributes are padded as $0$s before being processed by the deep transformer layers. \rebuttal{We highlight the significance of Global Attribute Embeddings in \cref{append:t-sne-visualization-interpretability}, where an analysis of the t-SNE latent space \citep{van2008visualizing} provides insights into their interpretability and importance.
}

\subsection{Training}

\subsubsection{Mixed Batch Training}
\label{subsec:mixed-batch-training}

 Optimizing a neural solver for tackling multiple tasks requires careful consideration of its training scheme, which needs to be robust against different variant distributions. We introduce a flexible approach which we call Mixed Batch Training (MBT) to efficiently reuse a single dataset to generate multiple problem variants. This optimizes data storage and processing. 

Let $\bm{X}$ be a dataset of \rebuttal{MDOVRPMBLTW} problem instances and $V$ be the set of attributes, where each attribute $\nu \in V$ is associated with a sampling probability $\mathbf{p}_{\nu}$.
For each instance $x \in \bm{X}$ we can write $x((\mathds{1}_{1})_{\nu \in V})$ to conveniently express using indicator functions $\mathds{1}_1$ for each attribute $\nu \in V$ that the instance $x$ has attribute $\nu$.
The sampling procedure of MBT can be defined as follows:
\[
    \bm{X}_{\text{subsampled}} = \{ x((\mathbf{1}_{\text{rand}(0,1) < \mathbf{p}_{\nu}})_{\nu \in V}) \}_{x \in \bm{X}},
\]
where $\text{rand}(0,1)$ draws an independent sample from $U[0, 1]$.
To sample uniformly across all problem variants, we set $\mathbf{p}_{\nu} = \frac12, \forall \nu \in V$.

 Note that the MDOVRPMBLTW problem variant is the most general problem variant we study in this paper and can be used to generate any of the other variants by selectively removing any combination of the global attributes (MD), (O), (MB) and (L), and node attributes (B) and (TW).
 We first consider OVRPBLTW as the most general problem variant from which we sample other variants. Then, for zero-shot generalization and few-shot learning, we additionally sample with multiple depots (MD) and mixed backhauls (MB). This increases the number of distinct variants from 16 variants that can be generated from OVRPBLTW to 48 variants that can be generated from MDOVRPMBLTW.

\begin{figure}[htb!]
    \centering
    \begin{minipage}[b]{0.65\linewidth}
        \centering 
        \includegraphics[width=\linewidth]{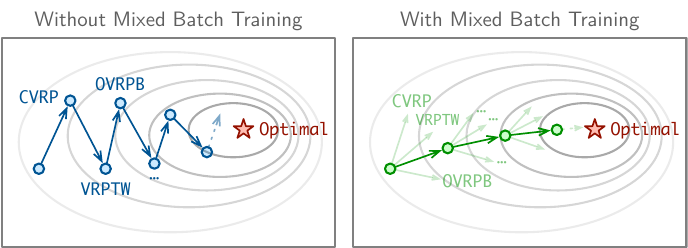}
    \end{minipage}
    \hspace{5mm}
    \begin{minipage}[b]{0.31\linewidth}
        \centering 
        \raisebox{-3.5mm}{\includegraphics[width=\linewidth]{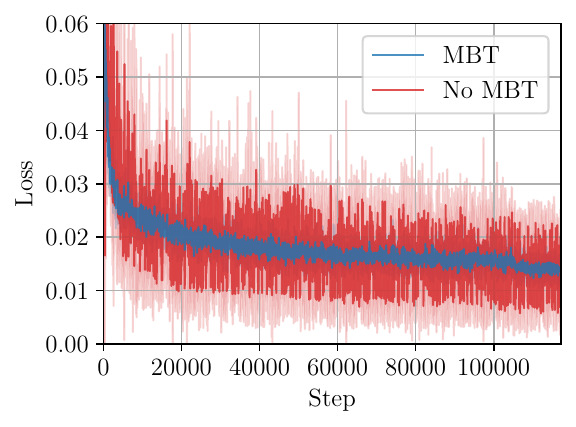}}
    \end{minipage}
\caption{\textbf{[Left]} Training without MBT leads to instability, since at each step the optimization is biased toward a single task. \textbf{[Middle]} Training \our{} with MBT allows for stable training. \textbf{[Right]} Effect of MBT on the loss during training.}
    \label{fig:mixed-batch-training}
\end{figure}

MBT is a flexible and scalable approach, capable of adapting to any problem where different constraints or features might be selectively activated or deactivated. 
\Cref{fig:mixed-batch-training} illustrates the stabilizing effect of MBT during training. Specifically, MBT significantly reduces the variance of the training loss, leading to faster and more stable convergence, as demonstrated in our experimental results in \cref{subsec:ablation-studies-main}.


\vspace{-2mm}

\subsubsection{Multi-task Reward Normalization}
\label{sec:reward-normalization}

As explained in \cref{sec:learning_neural_solvers}, the objective for RL-based NCO solvers is to maximize the expected reward. However, in multi-task learning settings different problems can yield rewards on different scales. 
To counteract potential biases during learning, we propose to apply reward normalization {per problem variant}. 
We implement four simple normalization techniques to calculate the normalized rewards $r_{\text{norm}, t}^{(k)}$ for all problem variants $k \in \{1, ... , K\}$ at training steps $t \geq 1$:
a) subtraction of the simple mean reward, b)  division through the simple mean reward, c) subtraction of the exponentially smoothed mean, and d) division through the exponentially smoothed mean. 
We calculate the average reward $\hat{r}_t^{(k)}$ \textit{up to} training step $t$ using the average \rebuttal{batch} reward $\bar{r}_t^{(k)}$ \textit{at} \rebuttal{training} step $t$ \rebuttal{(see \cref{app:avg_batch_reward})}.
The simple mean reward at step $t$ is calculated as:
\begin{equation}
\label{eq:simple_mean}
     \hat{r}_{t}^{(k)} = \left( (t-1) \cdot \hat{r}_{t-1}^{(k)} + \bar{r}_{t}^{(k)} \right) / t, \quad t \geq 1.
\end{equation}
For the exponential moving average we set $\hat{r}_1^{(k)} = \bar{r}_1^{(k)}$ and calculate the values for $t>1$ based on \cite{Hunter1986} using a smoothing factor $\alpha$:
\begin{equation}
\label{eq:exponential_smoothing}
    \hat{r}_{t}^{(k)} = (1 - \alpha) \cdot \hat{r}_{t-1}^{(k)} + \alpha \cdot \bar{r}_{t}^{(k)}, \quad 0 < \alpha < 1, \quad t > 1.
\end{equation}
Normalized rewards a)---d) can be calculated from original rewards $r_t^{(k)}$ with $r_{\text{norm}, t}^{(k)} = r_t^{(k)} - \hat{r}_t^{(k)}$ and $r_{\text{norm}, t}^{(k)} = r_t^{(k)}/|\hat{r}_t^{(k)}|$ for subtraction and division variants, respectively. 
Let $\xi(\bm{a}, \bm{x}) = r_{\text{norm}}^{(k)}(\bm{a}, \bm{x})$ be a function calculating the normalized reward for instance $\bm{x}$ that additionally maps instance $\bm{x}$ to variant $k$. The multi-task reward-normalized gradient becomes:
{
    \begin{equation}
    \label{eq:compact_normalized_loss_function}
    \nabla_{\theta} J(\theta) \approx \frac{1}{N} \sum_{i=1}^N \left(\xi(\bm{a}^i, \bm{x}) - \frac{1}{N} \sum_{j=1}^N \xi(\bm{a}^j, \bm{x})\right) \nabla_{\theta} \log p_{\theta}(\bm{a}^i|\bm{x}),
    \end{equation}
}
i.e., we employ the REINFORCE loss function with the POMO \citep{kwon2020pomo} shared mean baseline (right side of the parenthesis) to improve convergence, where both the reward and the shared baseline are normalized by $\xi$.

\subsection{Efficient Adapter Layers: Finetuning to Unseen Attributes}
\label{subsec:efficient-adapter-layers-eal}

Previous multi-task learning works \citep{liu2024multi, zhou2024mvmoe} train in environments of single-attribute VRP variants and, using compositionality \citep{ruis2021independent}, achieve promising results on zero-shot generalization to VRP variants combining these individual attributes. In \our{}, we go a step further and investigate how to efficiently generalize our pre-trained foundation model to variants with \textit{unseen} attributes. 
\citet{lin2024cross} propose pretraining a backbone model, on top of which specific Adapter Layers (AL) can be applied for finetuning to new problems. The rationale is that the backbone (i.e., the encoder layers) may capture transferable knowledge. However, doing so excludes previous information accumulated in the projection layers from the raw attribute features to the hidden space, complicating optimization. For example, if the first two out of $k$ dimensions encoded the Euclidean locations of nodes as $(x, y)$, re-initializing a new adapter layer from scratch will eliminate such transferable knowledge. Therefore, we propose Efficient Adapter Layers (EAL), an effective approach to few-shot learning for VRP foundation models.

Consider a linear projection layer $\mathbf{W} \in \mathbb{R}^{k \times d}$ as the original weight matrix for the projection from the raw attribute to latent space, where $k$ is the number of attributes and $d$ is the hidden dimension.
For simplicity, we consider unbiased linear projections to the latent space.
This can be readily extended to general affine projections using a bias term.
To accommodate $l$ new attributes, EAL augments $\mathbf{W}$ with zeros. 
The new matrix $\mathbf{W}' = \begin{bmatrix}
    \mathbf{W} & \mathbf{0} 
\end{bmatrix}^\top$ can be written as:

\pgfkeys{tikz/mymatrixenv/.style={decoration={brace},every left delimiter/.style={xshift=8pt},every right delimiter/.style={xshift=-8pt}}}
\pgfkeys{tikz/mymatrix/.style={matrix of math nodes,nodes in empty cells,left delimiter={[},right delimiter={]},inner sep=1pt,outer sep=1.5pt,column sep=8pt,row sep=8pt,nodes={minimum width=20pt,minimum height=10pt,anchor=center,inner sep=0pt,outer sep=0pt}}}
\pgfkeys{tikz/mymatrixbrace/.style={decorate,thick}}
\newcommand*\mymatrixbraceright[4][m]{
    \draw[mymatrixbrace] (#1.west|-#1-#3-1.south west) -- node[left=2pt] {#4} (#1.west|-#1-#2-1.north west);
}
\newcommand*\mymatrixbraceleft[4][m]{
    \draw[mymatrixbrace] (#1.east|-#1-#2-1.north east) -- node[right=2pt] {#4} (#1.east|-#1-#2-1.south east);
}
\newcommand*\mymatrixbracetop[4][m]{
    \draw[mymatrixbrace] (#1.north-|#1-1-#2.north west) -- node[above=2pt] {#4} (#1.north-|#1-1-#3.north east);
}
\newcommand*\mymatrixbracebottom[4][m]{
    \draw[mymatrixbrace] (#1.south-|#1-1-#2.north east) -- node[below=2pt] {#4} (#1.south-|#1-1-#3.north west);
}
\tikzset{style green/.style={
    set fill color=green!50!lime!60,draw opacity=0.4,
    set border color=green!50!lime!60,fill opacity=0.1,
  },
  style cyan/.style={
    set fill color=cyan!90!blue!60, draw opacity=0.4,
    set border color=blue!70!cyan!30,fill opacity=0.1,
  },
  style orange/.style={
    set fill color=orange!90, draw opacity=0.8,
    set border color=orange!90, fill opacity=0.3,
  },
  style brown/.style={
    set fill color=brown!70!orange!40, draw opacity=0.4,
    set border color=brown, fill opacity=0.3,
  },
  style purple/.style={
    set fill color=violet!90!pink!20, draw opacity=0.5,
    set border color=violet, fill opacity=0.3,    
  },
  kwad/.style={
    above left offset={-0.1,0.23},
    below right offset={0.10,-0.36},
    #1
  },
  pion/.style={
    above left offset={-0.07,0.2},
    below right offset={0.07,-0.32},
    #1
  },
  poz/.style={
    above left offset={-0.03,0.18},
    below right offset={0.03,-0.3},
    #1
  },set fill color/.code={\pgfkeysalso{fill=#1}},
  set border color/.style={draw=#1}
}
\vspace*{-6mm}
\[
{\mathbf{W}'^\top 
}=
\begin{tikzpicture}[baseline={-0.5ex},mymatrixenv]
\matrix [mymatrix,inner sep=4pt] (m)  
{
\tikzmarkin[kwad=style cyan]{Bis} w_{00} & \cdots & w_{0k} & \tikzmarkin[kwad=style green]{Prime} 0 & \tikzmarkend{Prime} \cdots & \tikzmarkend{Prime} 0 \\
\vdots & \ddots & \vdots & \tikzmarkend{Prime} \vdots & \tikzmarkend{Prime} \ddots & \tikzmarkend{Prime} \vdots \\
w_{d0} & \cdots & w_{dk} \tikzmarkend{Bis} & \tikzmarkend{Prime} 0 & \tikzmarkend{Prime} \cdots & 0 \tikzmarkend{Prime} \\
};
\mymatrixbracetop{1}{3}{\small$k$}
\mymatrixbracetop{4}{6}{\small$l$} 
\mymatrixbraceright{1}{3}{\small$d$}
\end{tikzpicture}
\]
\vspace*{1mm}


where $\mathbf{0} \in \mathbb{R}^{l \times d}$ is a matrix of zeros. The augmented matrix $\mathbf{W}'$ retains the original $k$ attributes and adds $l$ new attributes, which are initialized to zero. Doing so does not affect the model for seen attributes like AL does. Instead, these $l$ dimensions have no effect until fine-tuning on new variants occurs. This allows for new attributes to be included in any part of the model via EAL, as shown in \cref{fig:overview}.

\section{Experiments} \label{sec:experiments}

\begin{table*}[!t]
    \caption{Performance on 1000 test VRP instances. The lower, the better ($\downarrow$). ``$*$'' represents the best-known solutions. \our{} (\textsc{RF}) models outperform state-of-the-art neural baselines in all settings.
    }
    \vspace{-3.5mm}
    \label{tab:main-results}
    \begin{center}
    \renewcommand\arraystretch{1.05}
    \resizebox{0.98\textwidth}{!}{ 
    \begin{tabular}{ll|cccccc|ll|cccccc}
      \toprule
      \multicolumn{2}{c|}{\multirow{2}{*}{Solver}} & \multicolumn{3}{c}{\textbf{$n=50$}} & \multicolumn{3}{c|}{$n=100$} & \multicolumn{2}{c|}{\multirow{2}{*}{Solver}} &
      \multicolumn{3}{c}{\textbf{$n=50$}} & \multicolumn{3}{c}{$n=100$} \\
      \cmidrule(lr){3-5} \cmidrule(lr){6-8} \cmidrule(lr){11-13} \cmidrule(lr){14-16}
  
       & & Obj. & Gap & Time & Obj. & Gap & Time & & & Obj. & Gap & Time & Obj. & Gap & Time \\
      \midrule
\multirow{7}*{\rotatebox{90}{CVRP}} 
& HGS-PyVRP & 10.372 & * & 10.4m & 15.628 & * & 20.8m  & \multirow{7}*{\rotatebox{90}{VRPTW}} & HGS-PyVRP & 16.031 & * & 10.4m & 25.423 & * & 20.8m  \\
& OR-Tools & 10.572 & 1.907\% & 10.4m & 16.280 & 4.178\% & 20.8m      & & OR-Tools & 16.089 & 0.347\% & 10.4m & 25.814 & 1.506\% & 20.8m  \\
& MTPOMO           & 10.518 & 1.411\% & 2s     & 15.934 & 1.988\% & 7s     & 
                                                                             & MTPOMO         & 16.410 & 2.364\%  & 1s     & 26.412 & 3.873\%  & 7s   \\
& MVMoE            & 10.501 & 1.242\% & 2s     & 15.888 & 1.694\% & 9s     & & MVMoE          & 16.404 & 2.329\% & 2s     & 26.389 & 3.788\% & 9s     \\
& \textsc{RF}-POMO          & 10.508 & 1.314\% & 2s     & 15.908 & 1.826\% & 7s     & & \textsc{RF}-POMO        & 16.367 & 2.094\% & 1s     & 26.336 & 3.575\% & 7s     \\
& \textsc{RF}-MoE           & \textbf{10.499} & \textbf{1.226}\% & 2s     & 15.876 & 1.622\% & 9s     & & \textsc{RF}-MoE         & 16.389 & 2.234\% & 2s     & 26.322 & 3.519\% & 9s     \\
& \textsc{RF}-TE            & 10.504 & 1.274\% & 2s     & \textbf{15.857} & \textbf{1.505}\% & 7s     & & \textsc{RF}-TE          & \textbf{16.364} & \textbf{2.077}\% & 1s     & \textbf{26.235} & \textbf{3.178}\% & 7s     \\
      \midrule
\multirow{7}*{\rotatebox{90}{OVRP}} 
& HGS-PyVRP & 6.507 & * & 10.4m & 9.725 & * & 20.8m  & \multirow{7}*{\rotatebox{90}{VRPL}} & HGS-PyVRP & 10.587 & * & 10.4m & 15.766 & * & 20.8m \\
& OR-Tools & 6.553 & 0.686\% & 10.4m & 9.995 & 2.732\% & 20.8m  & & OR-Tools       & 10.570 & 2.343\% & 10.4m     & 16.466 & 5.302\% & 20.8m     \\
& MTPOMO         & 6.718  & 3.209\% & 1s    & 10.210 & 4.965\% & 6s     &    & MTPOMO         & 10.775 & 1.734\% & 1s     & 16.149 & 2.434\% & 7s     \\
& MVMoE          & 6.702  & 2.965\% & 2s    & 10.177 & 4.621\% & 9s     & & MVMoE          & 10.751 & 1.505\% & 2s     & 16.099 & 2.115\% & 9s     \\
& \textsc{RF}-POMO        & 6.698  & 2.904\% & 1s    & 10.180 & 4.659\% & 6s     & & \textsc{RF}-POMO        & 10.751 & 1.523\% & 1s     & 16.107 & 2.174\% & 6s     \\
& \textsc{RF}-MoE         & 6.697  & 2.886\% & 2s    & 10.139 & 4.229\% & 9s     & & \textsc{RF}-MoE         & \textbf{10.737} & \textbf{1.388}\% & 2s     & 16.070 & 1.941\% & 9s     \\
& \textsc{RF}-TE          & \textbf{6.684}  & \textbf{2.687}\% & 1s    & \textbf{10.121} & \textbf{4.055}\% & 6s     & & \textsc{RF}-TE          & 10.749 & 1.502\% & 1s     & \textbf{16.051} & \textbf{1.827}\% & 6s     \\
      \midrule
\multirow{7}*{\rotatebox{90}{VRPB}} 
& HGS-PyVRP & 9.687 & * & 10.4m & 14.377 & * & 20.8m   & \multirow{7}*{\rotatebox{90}{OVRPTW}} & HGS-PyVRP & 10.510 & * & 10.4m & 16.926 & * & 20.8m   \\
& OR-Tools & 9.802 & 1.159\% & 10.4m & 14.933 & 3.853\% & 20.8m 
& & OR-Tools & 10.519 & 0.078\% & 10.4m & 17.027 & 0.583\% & 20.8m  \\
& MTPOMO         & 10.033 & 3.564\% & 1s     & 15.082 & 4.922\% & 6s     &      & MTPOMO         & 10.668 & 1.479\% & 1s     & 17.420 & 2.892\% & 7s     \\
& MVMoE          & 10.005 & 3.270\% & 2s     & 15.023 & 4.508\% & 8s     & & MVMoE          & 10.669 & 1.492\% & 2s     & 17.416 & 2.872\% & 10s     \\
& \textsc{RF}-POMO        & 9.996 & 3.174\% & 1s     & 15.016 & 4.468\% & 6s     & & \textsc{RF}-POMO        & 10.657 & 1.378\% & 1s     & 17.391 & 2.720\% & 7s     \\
& \textsc{RF}-MoE         & 9.980 & 3.015\% & 2s     & 14.973 & 4.164\% & 8s     & & \textsc{RF}-MoE         & 10.674 & 1.539\% & 2s     & 17.387 & 2.697\% & 10s     \\
& \textsc{RF}-TE          & \textbf{9.977} & \textbf{2.989}\% & 1s     & \textbf{14.942} & \textbf{3.952}\% & 6s     & & \textsc{RF}-TE          & \textbf{10.652} & \textbf{1.326}\% & 1s     & \textbf{17.327} & \textbf{2.346}\% & 7s     \\

      \midrule
\multirow{7}*{\rotatebox{90}{VRPBL}} 
& HGS-PyVRP & 10.186 & * & 10.4m & 14.779 & * & 20.8m 
& \multirow{7}*{\rotatebox{90}{VRPBLTW}} 
                                                                             &   HGS-PyVRP      & 18.361 & *       & 10.4m     & 29.026 & *       & 20.8m     \\
& OR-Tools & 10.331 & 1.390\% & 10.4m & 15.426 & 4.338\% & 20.8m 
& & OR-Tools       & 18.422 & 0.332\% & 10.4m     & 29.830 & 2.770\% & 20.8m     \\
& MTPOMO         & 10.672 & 4.697\% & 1s     & 15.712 & 6.251\% & 7s     &  & MTPOMO         & 18.990 & 2.128\% & 1s     & 30.898 & 3.624\% & 7s     \\
& MVMoE          & 10.637 & 4.354\% & 2s     & 15.640 & 5.758\% & 9s     & & MVMoE          & 18.985 & 2.100\% & 2s     & 30.892 & 3.608\% & 10s     \\
& \textsc{RF}-POMO        & 10.593 & 3.942\% & 1s     & 15.628 & 5.695\% & 6s     & & \textsc{RF}-POMO        & \textbf{18.937} & \textbf{1.851}\% & 1s     & 30.796 & 3.284\% & 7s     \\
& \textsc{RF}-MoE         & \textbf{10.575} & \textbf{3.765}\% & 2s     & 15.541 & 5.121\% & 9s     & & \textsc{RF}-MoE         & 18.957 & 1.960\% & 2s     & \rebuttal{30.808} & 3.323\% & 10s     \\
& \textsc{RF}-TE          & 10.578 & 3.803\% & 1s     & \textbf{15.528} & \textbf{5.039}\% & 6s     & & \textsc{RF}-TE          & 18.941 & 1.877\% & 1s     & \textbf{30.688} & \textbf{2.923}\% & 7s     \\
      \midrule
\multirow{7}*{\rotatebox{90}{VRPBTW}} 
& HGS-PyVRP & 18.292 & * & 10.4m & 29.467 & * & 20.8m & \multirow{7}*{\rotatebox{90}{VRPLTW}} & HGS-PyVRP & 16.356 & * & 10.4m & 25.757 & * & 20.8m \\
& OR-Tools & 18.366 & 0.383\% & 10.4m & 29.945 & 1.597\% & 20.8m & & OR-Tools & 16.441 & 0.499\% & 10.4m & 26.259 & 1.899\% & 20.8m    \\
& MTPOMO         & 18.639 & 1.878\% & 1s     & 30.437 & 3.285\% & 7s     &  & MTPOMO         & 16.824 & 2.823\% & 1s     & 26.891 & 4.368\% & 7s     \\
& MVMoE          & 18.640 & 1.883\% & 2s     & 30.436 & 3.281\% & 9s     & & MVMoE          & 16.811 & 2.750\% & 2s     & 26.868 & 4.277\% & 9s     \\
& \textsc{RF}-POMO        & 18.601 & 1.670\% & 1s     & 30.341 & 2.961\% & 7s     & & \textsc{RF}-POMO        & \textbf{16.750} & \textbf{2.382}\% & 1s     & 26.783 & 3.948\% & 7s     \\
& \textsc{RF}-MoE         & 18.616 & 1.757\% & 2s     & 30.341 & 2.954\% & 9s     & & \textsc{RF}-MoE         & 16.777 & 2.550\% & 2s     & 26.774 & 3.912\% & 9s     \\
& \textsc{RF}-TE          & \textbf{18.600} & \textbf{1.676}\% & 1s     & \textbf{30.241} & \textbf{2.619}\% & 7s     & & \textsc{RF}-TE          & 16.762 & 2.454\% & 1s     & \textbf{26.689} & \textbf{3.579}\% & 7s     \\
      \midrule
\multirow{7}*{\rotatebox{90}{OVRPB}} 
& HGS-PyVRP & 6.898 & * & 10.4m & 10.335 & * & 20.8m  & \multirow{7}*{\rotatebox{90}{OVRPBL}} & HGS-PyVRP & 6.899 & * & 10.4m & 10.335 & * & 20.8m \\
& OR-Tools & 6.928 & 0.412\% & 10.4m & 10.577 & 2.315\% & 20.8m 
& & OR-Tools & 6.927 & 0.386\% & 10.4m & 10.582 & 2.363\% & 20.8m  \\
& MTPOMO         & 7.108  & 3.005\% & 1s     & 10.878 & 5.224\% & 7s     &    & MTPOMO         & 7.112  & 3.055\% & 1s     & 10.884 & 5.276\% & 6s     \\
& MVMoE          & 7.089  & 2.741\% & 2s     & 10.840 & 4.861\% & 9s     & & MVMoE          & 7.098  & 2.846\% & 2s     & 10.847 & 4.928\% & 9s     \\
& \textsc{RF}-POMO        & 7.086  & 2.688\% & 1s     & 10.836 & 4.821\% & 7s     & & \textsc{RF}-POMO        & 7.087  & 2.693\% & 1s     & 10.837 & 4.830\% & 6s     \\
& \textsc{RF}-MoE         & 7.080  & 2.513\% & 2s     & 10.805 & 4.522\% & 9s     & & \textsc{RF}-MoE         & 7.083  & 2.635\% & 2s     & 10.806 & 4.534\% & 9s     \\
& \textsc{RF}-TE          & \textbf{7.071}  & \textbf{2.479}\% & 1s     & \textbf{10.772} & \textbf{4.208}\% & 6s     & & \textsc{RF}-TE          & \textbf{7.074}  & \textbf{2.508}\% & 1s     & \textbf{10.778} & \textbf{4.262}\% & 6s     \\
      \midrule
\multirow{7}*{\rotatebox{90}{OVRPBLTW}} & HGS-PyVRP & 11.668 & * & 10.4m & 19.156 & * & 20.8m 
& \multirow{7}*{\rotatebox{90}{OVRPBTW}} & HGS-PyVRP & 11.669 & * & 10.4m & 19.156 & * & 20.8m  \\
& OR-Tools & 11.681 & 0.106\% & 10.4m & 19.305 & 0.767\% & 20.8m & & OR-Tools & 11.682 & 0.109\% & 10.4m & 19.303 & 0.757\% & 20.8m  \\
& MTPOMO         & 11.817 & 1.260\% & 1s     & 19.637 & 2.496\% & 7s     & 
                                                                             & MTPOMO         & 11.814 & 1.229\% & 1s     & 19.635 & 2.485\% & 7s     \\
& MVMoE          & 11.822 & 1.301\% & 2s     & 19.641 & 2.518\% & 10s     & & MVMoE          & 11.819 & 1.271\% & 2s     & 19.638 & 2.503\% & 10s     \\
& \textsc{RF}-POMO        & 11.805 & 1.157\% & 1s     & 19.609 & 2.344\% & 8s     & & \textsc{RF}-POMO        & 11.804 & 1.148\% & 1s     & 19.607 & 2.339\% & 7s     \\
& \textsc{RF}-MoE         & 11.824 & 1.312\% & 2s     & 19.607 & 2.334\% & 10s    & & \textsc{RF}-MoE         & 11.823 & 1.304\% & 2s     & 19.606 & 2.328\% & 10s    \\
& \textsc{RF}-TE          & \textbf{11.805} & \textbf{1.150}\% & 1s     & \textbf{19.551} & \textbf{2.048}\% & 7s     & & \textsc{RF}-TE          & \textbf{11.805} & \textbf{1.151}\% & 1s     & \textbf{19.550} & \textbf{2.042}\% & 7s     \\

\midrule
\multirow{7}*{\rotatebox{90}{OVRPL}} 
& HGS-PyVRP & 6.507 & * & 10.4m & 9.724 & * & 20.8m 
 & \multirow{7}*{\rotatebox{90}{OVRPLTW}} & HGS-PyVRP & 10.510 & * & 10.4m & 16.926 & * & 20.8m \\
& OR-Tools & 6.552 & 0.668\% & 10.4m & 10.001 & 2.791\% & 20.8m  & & OR-Tools       & 10.497 & 0.114\% & 10.4m     & 17.023 & 0.728\% & 20.8m     \\
& MTPOMO         & 6.719  & 3.227\% & 1s     & 10.214 & 5.002\% & 6s     &  & MTPOMO         & 10.670 & 1.500\% & 1s     & 17.420 & 2.889\% & 7s     \\
& MVMoE          & 6.707  & 3.030\% & 2s     & 10.184 & 4.696\% & 9s     & & MVMoE          & 10.671 & 1.511\% & 2s     & 17.419 & 2.885\% & 10s     \\
& \textsc{RF}-POMO        & 6.701  & 2.949\% & 1s     & 10.180 & 4.659\% & 6s     & & \textsc{RF}-POMO        & 10.657 & 1.375\% & 1s     & 17.393 & 2.731\% & 7s     \\
& \textsc{RF}-MoE         & 6.696  & 2.864\% & 2s     & 10.140 & 4.249\% & 9s     & & \textsc{RF}-MoE         & 10.673 & 1.532\% & 2s     & 17.386 & 2.693\% & 10s     \\
& \textsc{RF}-TE          & \textbf{6.686}  & \textbf{2.721}\% & 1s     & \textbf{10.120} & \textbf{4.052}\% & 6s     & & \textsc{RF}-TE          & \textbf{10.653} & \textbf{1.341}\% & 1s     & \textbf{17.327} & \textbf{2.347}\% & 7s     \\
      \bottomrule
    \end{tabular}}
    \end{center}
  \end{table*}


In this section, we empirically demonstrate the state-of-the-art performance of \our{} in extensive experiments on 48 VRP variants. Our code is publicly available\footnote{\url{https://github.com/ai4co/routefinder}}.
We address the following research questions:
\vspace{-2mm}
\begin{description}
    \item[\textbf{(RQ1)}] Does \our{} outperform state-of-the-art foundation models on different VRP variants?
    \vspace{-2mm}
    \item[\textbf{(RQ2)}] How do the individual novel components of \our{} contribute to its performance?
    \vspace{-2mm}
    \item[\textbf{(RQ3)}] Is the proposed EAL effective in finetuning \our{} models to unseen VRP variants?
\end{description}
\vspace{-3mm}
\paragraph{Hardware} All training runs are conducted on NVIDIA A100 GPUs and take between 9 to 24 hours per model. Evaluation is conducted on an AMD Ryzen Threadripper 3960X 24-core CPU with a single RTX 3090 GPU.
\paragraph{Baselines}
\textit{Traditional solvers}: 
We use PyVRP~\citep{wouda2024pyvrp}, an open-source, state-of-the-art heuristic VRP solver built on top of HGS-CVRP~\citep{vidal2022hybrid}, and the popular Google OR-Tools~\citep{perron2023ortools}. 
Both solve each instance on a single CPU core with a time limit of 10 and 20 seconds for instances with 50 and 100 nodes, respectively. We parallelize traditional solvers across 16 CPU cores as in \citet{zhou2024mvmoe}.
\textit{Neural solvers}:
We consider recent multi-task learning baselines for the VRP, including the recent MTPOMO \citep{liu2024multi}, which is based on POMO \citep{kwon2020pomo}, and MVMoE \citep{zhou2024mvmoe}, which introduces mixture-of-experts \citep{fedus2022review} to improve the model performance. 
\textit{Our models}:
We consider three versions of \our{}, denoted as \textsc{RF} in the tables: one considering the (MT)POMO encoder (\textsc{RF}-POMO), one with the MVMoE model with four experts and hierarchical gating (\textsc{RF}-MoE), and one with our modern Transformer-based Encoder (\textsc{RF}-TE). We use Reward Normalization with division through the exponentially smoothed mean with $\alpha=0.25$. 
Further details are available in \cref{append:model}.

\tmlr{
\paragraph{Data Generation}
To train and evaluate RouteFinder across a diverse set of VRP variants, we employ a unified data generation process detailed in \cref{sec:mtvrp-data-generation}. Problem instances are generated using our modular environment that supports all combinations of seven core VRP attributes: capacity (C), open routes (O), backhauls (B), duration limits (L), time windows (TW), mixed backhauls (MB), and multi-depots (MD). Node locations are uniformly sampled in $[0,1]^2$, while vehicle capacity is set relative to the number of nodes, i.e., $C=40$ for $n=50$ and $C=50$ for $n=100$. Linehaul and backhaul demands are integers uniformly sampled from $\{1,\dots,9\}$, with each customer having either a linehaul or backhaul demand with probability 0.8 and 0.2, respectively. Time windows are generated using a randomized offset procedure ensuring feasibility with lengths sampled uniformly from $[0.18, 0.2]$, and service times are sampled uniformly from $[0.15, 0.18]$. Duration limits are drawn from the distribution described in \cref{sec:mtvrp-data-generation}, i.e., ensuring reachability while introducing meaningful constraints. For multi-depot problems, three depots are used by default. Attribute combinations are sampled uniformly using mixed batch training with with $\mathbf{p}_{\nu} = 0.5, \forall \nu \in V$ to cover all variants equally.
}

\paragraph{Training}
Each model is trained for 300 epochs on $100{,}000$ VRP instances that are generated on the fly and include all attributes -- except for (MB) and (MD) which are only used in finetuning. We use the Adam optimizer \citep{adam} with a learning rate of $3 \times 10^{-4}$ and batch size of $256$. At epochs 270 and 295, the learning rate is multiplied by $0.1$. 
Note that our setup differs from the one in \citet{liu2024multi} and \citet{zhou2024mvmoe} in that we do not artificially restrict the variants with single attributes (such as only (B) or (TW)) but train on \textit{all} available variants, similarly to how LLMs are trained on all available data. This is readily available through our unified VRP environment (more details in \cref{append:unified-vrp-env-details}).

\paragraph{Evaluation}
\tmlr{We evaluate all approaches on $1{,}000$ instances of held-out test data for each size $n$ of each variant.} We roll out greedy solutions for all NCO approaches using multi-starts and $8 \times$ symmetric augmentations resulting in $n \times 8$ solutions per instance from which the best is selected \citep{liu2024multi,zhou2024mvmoe}.

\subsection{(RQ1) Main Results}
\label{subsec:experiments-main-results}

\paragraph{In-distribution}
\cref{tab:main-results} shows the in-distribution testing results for $50$ and $100$ nodes $n$. \our{} models consistently outperform neural baselines across all variants by more than $10\%$. While changing the encoder to the MVMoE's structure (\textsc{RF}-MoE) may slightly improve the performance in limited settings -- with a higher inference cost due to the more complex structure of mixture-of-experts -- the proposed Transformer Encoder (\textsc{RF}-TE) outperforms the other models in almost all settings, particularly for $n=100$. 

\paragraph{Out-of-distribution}
For CVRP, the training distribution with $100$ nodes considers a vehicle capacity $C = 50$. We study generalization over different capacities $C \in \{ 30, 50, 70, 90, 110, 130, 150, 200 \}$ and show the results in \cref{tab:ood-cvrp}. POMO trained on CVRP performs best for capacities close to the training distribution, but \our{} demonstrates superior generalization and finds the best solutions for larger capacities.

\begin{table*}[h!]
\caption{\rebuttal{Comparison of our model with single-task POMO on out-of-distribution CVRP instances.}}
\label{tab:ood-cvrp}
\begin{center}
\large %
\renewcommand{\arraystretch}{1.05} %
\rebuttal{
\resizebox{\textwidth}{!}{ 
\begin{tabular}{l|cccccccccccccccc}

\toprule

Vehicle Capacity & \multicolumn{2}{c}{30} & \multicolumn{2}{c}{50} & \multicolumn{2}{c}{70} & \multicolumn{2}{c}{90} & \multicolumn{2}{c}{110} & \multicolumn{2}{c}{130} & \multicolumn{2}{c}{150} & \multicolumn{2}{c}{200} \\
\cmidrule(lr){2-3} \cmidrule(lr){4-5} \cmidrule(lr){6-7} \cmidrule(lr){8-9} \cmidrule(lr){10-11} \cmidrule(lr){12-13} \cmidrule(lr){14-15} \cmidrule(lr){16-17}

 & Obj. & Gap & Obj. & Gap & Obj. & Gap & Obj. & Gap & Obj. & Gap & Obj. & Gap & Obj. & Gap & Obj. & Gap \\

\midrule
      
POMO\_CVRP     & \textbf{22.95}   &  \textbf{*} & \textbf{15.72}   &  \textbf{*} & \textbf{12.91}   &  \textbf{*} & 11.48   &  * & 10.64   &  * & 10.04   &  * &  9.75   &  * &  9.24   &  * \\
MTPOMO   & 23.29   &  1.50\% & 15.87   &  0.94\% & 13.07   &  1.24\% & 11.69   &  1.77\% & 10.88   &  2.30\% & 10.34   &  2.90\% & 10.04   &  2.97\% &  9.59   &  3.77\% \\
MVMoE    & 23.04   &  0.43\% & 15.83   &  0.67\% & 12.99   &  0.61\% & 11.54   &  0.49\% & 10.67   &  0.33\% & 10.06   &  0.12\% &  9.74   & -0.09\% &  9.21   & -0.28\% \\
RF-POMO  & 23.10   &  0.69\% & 15.84   &  0.77\% & 13.03   &  0.90\% & 11.61   &  1.07\% & 10.76   &  1.17\% & 10.17   &  1.26\% &  9.86   &  1.12\% &  9.38   &  1.51\% \\
RF-MoE   & 23.13   &  0.80\% & 15.81   &  0.58\% & 13.00   &  0.74\% & 11.59   &  0.89\% & 10.74   &  0.92\% & 10.14   &  0.95\% &  9.82   &  0.69\% &  9.31   &  0.75\% \\
RF-TE    & 22.96   &  0.06\% & 15.79   &  0.44\% & 12.95   &  0.29\% & \textbf{11.47}   & \textbf{-0.07\%} & \textbf{10.56}   & \textbf{-0.71\%} &  \textbf{9.92}   & \textbf{-1.22\%} &  \textbf{9.59}   & \textbf{-1.67\%} &  \textbf{9.02}   & \textbf{-2.36\%} \\

\bottomrule

\end{tabular}}
}
\end{center}
\end{table*}

\begin{wrapfigure}[11]{r}{0.55\textwidth}
  \centering
  \vspace{-5mm}
  \includegraphics[width=\linewidth]{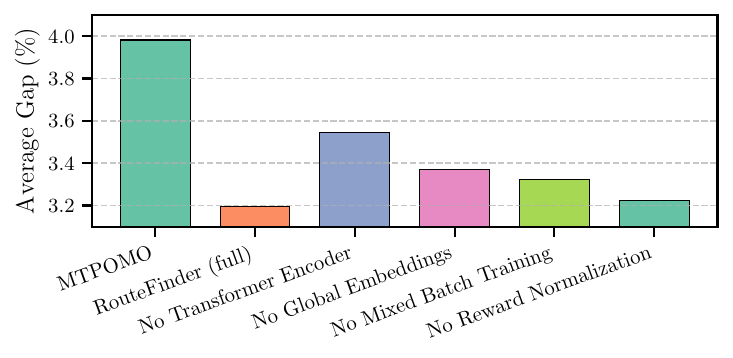}
\caption{Ablation study on \our{} components.}
    \label{fig:ablation-study-main}
\end{wrapfigure}

We further evaluate out-of-distribution values for additional attributes (\cref{append:out-of-distribution-attribute-generalization}) and large-scale CVRPLIB instances (\cref{sec:cvrplib}), and find that \our{} can consistently generalize more robustly in real-world settings than neural baselines, including single-variant POMO \citep{kwon2020pomo}.


\subsection{(RQ2) Ablation Studies}
\label{subsec:ablation-studies-main}

\paragraph{Contribution of components}
We conduct ablation studies to evaluate the impact of our individual contributions.

\cref{fig:ablation-study-main} compares the performance of \our{} (\textsc{RF}-TE) against its variants with ablated components, using the results for MTPOMO as a baseline. 
%
All components contribute to the performance of \our{}. 

%

%


\paragraph{Effect of MBT on training loss and convergence speed}
We compare two \our{} models trained with identical hyperparameters on $n=50$, one with and one without MBT. By keeping the overall sampling distribution the same but mixing variants in the same batch, MBT allows for a more stable gradient across the different tasks, resulting in a substantially more stable loss compared to training without it.

\begin{figure}[H]
    \centering
    \includegraphics[width=.9\linewidth]{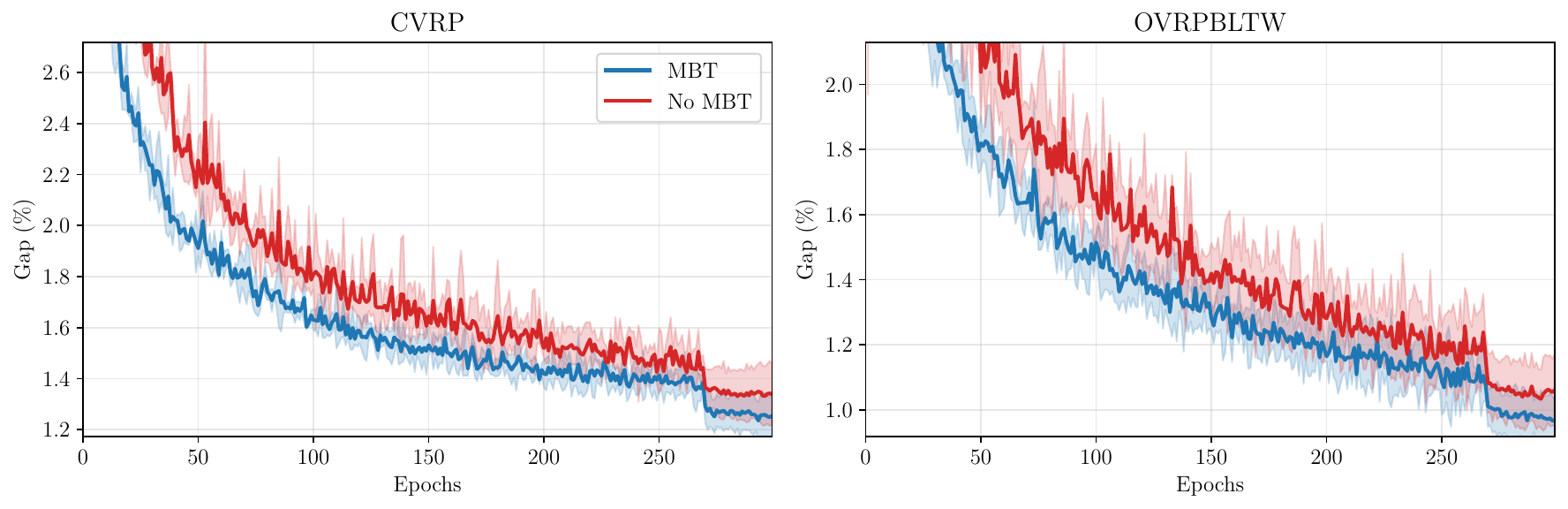}
    \vspace{-3mm}
    \caption{Mixed Batch Training (MBT) improves convergence during training, shown here for the CVRP and OVRPBLTW.}
    \label{fig:mbt-convergence-speed}
\end{figure}

In \cref{fig:mbt-convergence-speed} we show the validation gaps on held-out instances exemplarily for two variants, CVRP and OVRPBLTW. In addition to stabilizing the training loss, MBT also speeds up convergence. \cref{fig:mbt-epochs} in
\cref{subsec:append-mixed-batch-training-study-stability} shows that this holds for all 16 variants that can be generated from OVRPBLTW.


\paragraph{Further ablation studies}
We study the ablation of single Transformer Encoder components and find that the combination of all components provides the best performance (\cref{append:effect-transformer-components}). 
We further study the effect of different Reward Normalization techniques (\cref{append:effect-normalization-scheme}), and the importance of MBT for 
convergence speed (\cref{subsec:append-mixed-batch-training-study-stability}) and its effect on imbalanced variant distributions (\cref{subsec:append-mixed-batch-training-study-stability-imbalanced-distributions}).
Finally, we visualize the rich latent space of \our{} via t-SNE \citep{van2008visualizing} and compare it to those from MTPOMO and MVMoE, where we find that \our{} generates more and better defined clusters, indicating a better-learned representation due to the Global Attribute Embeddings (\cref{append:t-sne-visualization-interpretability}).

\subsection{(RQ3) Finetuning with EAL}
\label{sec:main-paper-eal}

We finally evaluate \our{} (\textsc{RF}-TE) in few-shot learning, finetuning with our proposed EAL to 32 unseen variants, introducing the multi-depot (MD) and mixed backhauls (MB) attributes. Most of these variants are modeled with a learning approach for the first time in this work.  We compare 1) zero-shot performance of \our{}, 2) training a new model from scratch, 3) AL from \citet{lin2024cross}, which adds new adapter layers while keeping the pre-trained backbone, and 4) our proposed EAL. We train a model from scratch and finetune \our{} with AL and EAL similarly to the main experiments, but for $10$ epochs and $10{,}000$ instances sampled for each. 
\cref{tab:eal-finetuning} shows that EAL consistently outperforms baselines in few-shot learning, including a $20\%$ relative improvement over AL. We also compare AL and EAL at ``step 0'', i.e., after replacing the new adapter layers. Notably, while AL with the untrained new layers greatly degrades the performance unless further training is performed, EAL maintains the zero-shot performance without additional training, providing a much better starting point.

\begin{table*}[h!]
\centering
\caption{Finetuning performance on $1{,}000$ with new multi-depot (MD) and mixed backhaul (MB) variants. \our{}'s EAL maintains the zero-shot performance and performs significantly better than AL \citep{lin2024cross}.
}
\vspace{-2mm}
\label{tab:eal-finetuning}
\resizebox{\textwidth}{!}{%
\large %
\renewcommand{\arraystretch}{1.05} %
\begin{tabular}{l | cccccccccccccccc}
\toprule
 & \multicolumn{2}{c}{MDVRPMB} & \multicolumn{2}{c}{MDOVRPMB} & \multicolumn{2}{c}{MDVRPMBL} & \multicolumn{2}{c}{MDVRPMBTW} & \multicolumn{2}{c}{MDOVRPMBL} & \multicolumn{2}{c}{MDOVRPMBTW} & \multicolumn{2}{c}{MDVRPMBLTW} & \multicolumn{2}{c}{MDOVRPMBLTW} \\
\cmidrule(lr){2-3} \cmidrule(lr){4-5} \cmidrule(lr){6-7} \cmidrule(lr){8-9} \cmidrule(lr){10-11} \cmidrule(lr){12-13} \cmidrule(lr){14-15} \cmidrule(lr){16-17} 
Method & Obj. & Gap & Obj. & Gap & Obj. & Gap & Obj. & Gap & Obj. & Gap & Obj. & Gap & Obj. & Gap & Obj. & Gap \\
\midrule
HGS-PyVRP & 10.68 & *      & 7.66 & *            & 10.71 & *         & 19.29 & *         & 7.66 & *          & 12.96 & *         & 19.31 & *         & 12.96 & *         \\
OR-Tools & 12.22 & $14.37\%$ & 8.88 & $15.83\%$  & 12.23 & $14.23\%$ & 22.39 & $16.12\%$ & 8.87 & $15.73\%$  & 14.49 & $11.79\%$ & 22.43 & $16.16\%$ & 14.49 & $11.79\%$ \\
Zero-shot  & 14.99 & $40.80\%$ & 10.77 & $40.67\%$ & 15.28 & $43.27\%$ & 28.43 & $47.93\%$ & 10.76 & $40.62\%$ & 18.49 & $43.14\%$ & 28.80 & $49.69\%$ & 18.50 & $43.17\%$ \\
Train (scratch) &  13.12 & 22.88$\%$ &  9.37 & 22.32$\%$ &  13.24 & 23.72$\%$ &  22.85 & 18.56$\%$ &  9.38 & 22.44$\%$ &  15.13 & 16.75$\%$ &  22.90 & 18.65$\%$ &  15.11 & 16.60$\%$  \\
AL (step 0) & 34.12 & $223.14\%$ & 26.36 & $245.53\%$ & 27.41 & $158.88\%$ & 48.94 & $155.28\%$ & 24.11 & $216.01\%$ & 31.53 & $144.89\%$ & 46.80 & $143.89\%$ & 30.08 & $133.48\%$ \\
AL  &  13.10 & 22.70$\%$ &  9.36 & 22.14$\%$ &  13.20 & 23.36$\%$ &  22.90 & 18.76$\%$ &  9.38 & 22.46$\%$ &  15.28 & 17.91$\%$ &  23.02 & 19.26$\%$ &  15.39 & 18.77$\%$ \\
EAL (step 0) & 14.99 & $40.80\%$ & 10.77 & $40.67\%$ & 15.28 & $43.27\%$ & 28.43 & $47.93\%$ & 10.76 & $40.62\%$ & 18.49 & $43.14\%$ & 28.80 & $49.69\%$ & 18.50 & $43.17\%$ \\
EAL  & \textbf{12.70} & \textbf{18.98}$\%$ & \textbf{8.53} & \textbf{11.35}$\%$ & \textbf{12.68} & \textbf{18.56}$\%$ & \textbf{21.41} & \textbf{11.05}$\%$ & \textbf{8.54} & \textbf{11.43}$\%$ & \textbf{13.93} & \textbf{7.41}$\%$ & \textbf{21.44} & \textbf{11.09}$\%$ & \textbf{13.91} & \textbf{7.32}$\%$ \\\bottomrule
\end{tabular}
}
\end{table*}

\begin{figure}[H]
    \centering
    \begin{subfigure}[t]{0.3\textwidth}
        \centering
        \includegraphics[width=\textwidth]{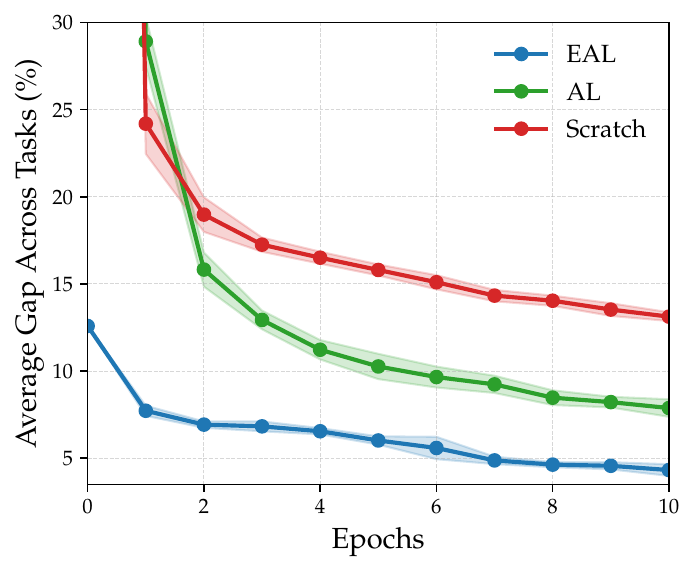}
        \caption{Mixed backhaul}
        \label{fig:finetuning-mb}
    \end{subfigure}
    \hfill
    \begin{subfigure}[t]{0.3\textwidth}
        \centering
        \includegraphics[width=\textwidth]{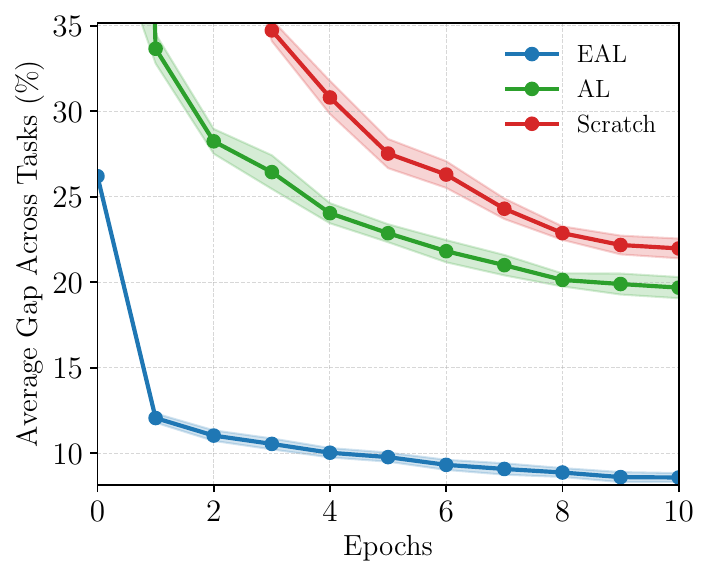}
        \caption{Multi-depot}
        \label{fig:finetuning-md}
    \end{subfigure}
    \hfill
    \begin{subfigure}[t]{0.3\textwidth}
        \centering
        \includegraphics[width=\textwidth]{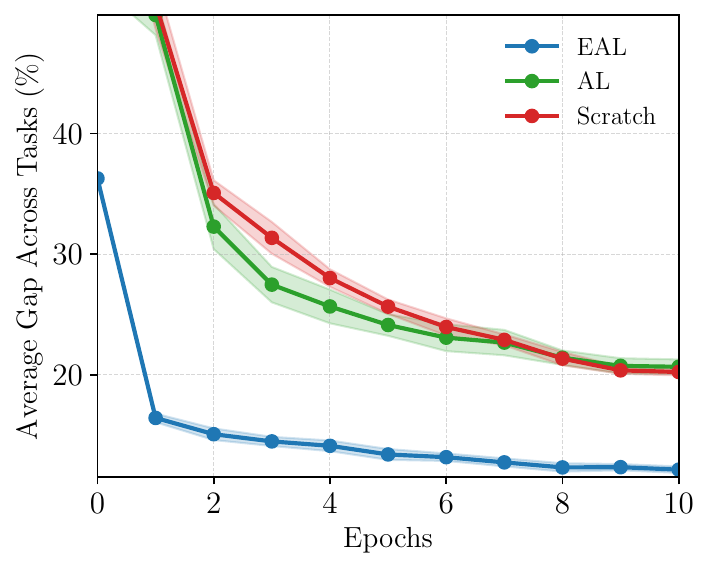}
        \caption{Mixed backhaul \& multi-depot}
        \label{fig:finetuning-both}
    \end{subfigure}
    \caption{Validation gaps averaged across new tasks including unseen features (a) mixed backhaul (MB), (b) multi-depot (MD), and (c) their combination (MB\&MD) for retraining from scratch, AL and EAL finetuning.}
    \label{fig:finetuning-trends}
\end{figure}

We conduct additional experiments on zero-shot generalization 
and finetuning across three different settings of unseen variants in order of difficulty: (a) mixed backhaul, (b) multi-depot, and (c) mixed backhaul \& multi-depot.
We again train for 10 epochs with $10{,}000$ instances sampled for each epoch, and use \our{} models with Transformer Encoder (\textsc{RF}-TE). 
We show the validation gap trends for these three settings in \cref{fig:finetuning-trends}, comparing training from scratch, finetuning with AL, and with EAL.
For AL and EAL we continue training from the checkpoints that result from the main experiments in \cref{subsec:experiments-main-results}.
We can see that EAL, having a much better starting point than AL and training from scratch, clearly dominates the other two methods.
This dominance becomes more pronounced with increasing difficulty of the finetuning task from MB to MB\&MD, indicating it is a suitable method for efficient finetuning to new tasks.

In \cref{subsec:finetuning-appendix} we provide more detailed results on finetuning for unseen variants as well as additional results on finetuning with EAL for single-variant models. These results show that finetuning \our{} to unseen attributes achieves better results than finetuning single-variant models based on POMO. This is a strong argument for foundation models in routing, as the ability to quickly adapt to new tasks is critical in real-world routing problems.

\section{Conclusion} 
\label{sec:conclusion}

In this work, we presented \our{}, a comprehensive framework to develop foundation models for VRPs. 
Extensive evaluations on 48 VRP variants showed that \our{} outperforms state-of-the-art neural baselines. 
\our{} represents an early attempt to learn a foundation model across VRP variants. While demonstrating strong generalization, it does so at a slight expense in solution quality for in-distribution results compared to models trained on single variants. For future work, we plan to extend \our{} to support further variants in the vast VRP literature. We also intend to improve the model with exciting research directions, including decomposition methods \citep{ye2023glop, zheng2024udc} and end-to-end construction and improvement \citep{kong2024efficient}.


\section*{Acknowledgements}

We are deeply grateful to the members of the AI4CO open research community for their invaluable contributions to \our{} and related projects, including RL4CO. 
Our thanks also extend to OMELET for providing additional computing resources.
This work was supported by the Institute of Information \& Communications Technology Planning \& Evaluation (IITP) grant, funded by the Korean government (MSIT) [Grant No. 2022-0-01032, Development of Collective Collaboration Intelligence Framework for Internet of Autonomous Things].
Nayeli Gast Zepeda and André Hottung received support from the Deutsche Forschungsgemeinschaft (DFG, German Research Foundation) under Grant No. 521243122. 
We also gratefully acknowledge the Paderborn Center for Parallel Computing (PC²) for providing valuable computing time for this project.




\bibliography{main}
\bibliographystyle{tmlr}

\newpage
\appendix

\section{Unified VRP Environment Details}
\label{append:unified-vrp-env-details}

 We consider the seven attributes from \cref{sec:vrp_features} for instance generation through our environment definition explained in \cref{sec:vrp_env}. Leveraging our environment's modular structure, we build the 16 VRP variants as used in MVMoE \citep{zhou2024mvmoe}, but by differentiating between \textit{traditional} (B) and \textit{mixed} (MB) backhauls, as defined in 
 \cite{avci2015adaptive}, we extend that number to 24.
 By considering multi-depot problems, we further increase that number to 48 variants that can be solved with \our{} (see \cref{table:vrp_variants_asymmetric}).

 \begin{table}[h!]
\centering
\caption{The 48 VRP variants we consider. All variants include the base Capacity (C). The $k=5$ features O, B, L, TW, and MD can be combined into any subset, including the empty set and itself (i.e., a \textit{power set}) with $2^k = 32$ possible combinations. The Mixed (M) global feature creates new Mixed Backhaul (MB) variants in generalization studies, adding 16 more variants.}
\label{table:vrp_variants_asymmetric}
\scriptsize
\resizebox{\textwidth}{!}{%
\begin{tabular}{l | cccccccc}
\toprule
\textit{VRP Variant} & {\thead{Capacity\\(C)}} & {\thead{Open Route\\(O)}} & {\thead{Backhaul\\(B)}} & {\thead{Mixed\\(M)}} & {\thead{Duration Limit\\(L)}} & {\thead{Time Windows\\(TW)}} & {\thead{Multi-depot\\(MD)}} \\
\hline \\[-1.5ex] %
CVRP & \cmark & & & & & & \\
OVRP & \cmark & \cmark & & & & & \\
VRPB & \cmark & & \cmark & & & & \\
VRPL & \cmark & & & & \cmark & & \\
VRPTW & \cmark & & & & & \cmark & \\
OVRPTW & \cmark & \cmark & & & & \cmark & \\
OVRPB & \cmark & \cmark & \cmark & & & & \\
OVRPL & \cmark & \cmark & & & \cmark & & \\
VRPBL & \cmark & & \cmark & & \cmark & & \\
VRPBTW & \cmark & & \cmark & & & \cmark & \\
VRPLTW & \cmark & & & & \cmark & \cmark & \\
OVRPBL & \cmark & \cmark & \cmark & & \cmark & & \\
OVRPBTW & \cmark & \cmark & \cmark & & & \cmark & \\
OVRPLTW & \cmark & \cmark & & & \cmark & \cmark & \\
VRPBLTW & \cmark & & \cmark & & \cmark & \cmark & \\
OVRPBLTW & \cmark & \cmark & \cmark & & \cmark & \cmark & \\
VRPMB & \cmark & & \cmark & \cmark & & & \\
OVRPMB & \cmark & \cmark & \cmark & \cmark & & & \\
VRPMBL & \cmark & & \cmark & \cmark & \cmark & & \\
VRPMBTW & \cmark & & \cmark & \cmark & & \cmark & \\
OVRPMBL & \cmark & \cmark & \cmark & \cmark & \cmark & & \\
OVRPMBTW & \cmark & \cmark & \cmark & \cmark & & \cmark & \\
VRPMBLTW & \cmark & & \cmark & \cmark & \cmark & \cmark & \\
OVRPMBLTW & \cmark & \cmark & \cmark & \cmark & \cmark & \cmark & \\
MDCVRP & \cmark & & & & & & \cmark \\
MDOVRP & \cmark & \cmark & & & & & \cmark \\
MDVRPB & \cmark & & \cmark & & & & \cmark \\
MDVRPL & \cmark & & & & \cmark & & \cmark \\
MDVRPTW & \cmark & & & & & \cmark & \cmark \\
MDOVRPTW & \cmark & \cmark & & & & \cmark & \cmark \\
MDOVRPB & \cmark & \cmark & \cmark & & & & \cmark \\
MDOVRPL & \cmark & \cmark & & & \cmark & & \cmark \\
MDVRPBL & \cmark & & \cmark & & \cmark & & \cmark \\
MDVRPBTW & \cmark & & \cmark & & & \cmark & \cmark \\
MDVRPLTW & \cmark & & & & \cmark & \cmark & \cmark \\
MDOVRPBL & \cmark & \cmark & \cmark & & \cmark & & \cmark \\
MDOVRPBTW & \cmark & \cmark & \cmark & & & \cmark & \cmark \\
MDOVRPLTW & \cmark & \cmark & & & \cmark & \cmark & \cmark \\
MDVRPBLTW & \cmark & & \cmark & & \cmark & \cmark & \cmark \\
MDOVRPBLTW & \cmark & \cmark & \cmark & & \cmark & \cmark & \cmark \\
MDVRPMB & \cmark & & \cmark & \cmark & & & \cmark \\
MDOVRPMB & \cmark & \cmark & \cmark & \cmark & & & \cmark \\
MDVRPMBL & \cmark & & \cmark & \cmark & \cmark & & \cmark \\
MDVRPMBTW & \cmark & & \cmark & \cmark & & \cmark & \cmark \\
MDOVRPMBL & \cmark & \cmark & \cmark & \cmark & \cmark & & \cmark \\
MDOVRPMBTW & \cmark & \cmark & \cmark & \cmark & & \cmark & \cmark \\
MDVRPMBLTW & \cmark & & \cmark & \cmark & \cmark & \cmark & \cmark \\
MDOVRPMBLTW & \cmark & \cmark & \cmark & \cmark & \cmark & \cmark & \cmark \\
\bottomrule
\end{tabular}
}
\vspace{2mm}
\end{table}

 We describe additional details of the Unified VRP environment, including data generation in \cref{sec:mtvrp-data-generation} and environment logic in \cref{sec:mtvrp-environment-logic}. For a better understanding, we invite the reader to look at the source code, which we tried our best to comment on for clarity, at \url{https://github.com/ai4co/routefinder}.

\subsection{Data generation}
\label{sec:mtvrp-data-generation}

We now explain the individual steps in the data generation process we use for our modular VRP environment, including the node attributes and global attributes.

    While throughout the main part of this paper, we have focused on routing problems with a single depot, our unified environment can actually handle problems with multiple depots, where we define $m$ as the number of depots. For comparability to the neural baselines, the main experiments were run on single-depot problems, but we report results for multi-depot problems (\cref{subsec:finetuning-appendix}).

\paragraph{Locations}
We generate $m + n$ locations randomly with $x_i$ and $y_i \sim U(0,1), \forall i \in \{0,...,m + n - 1\}$, where $[x_i, y_i], i \in \{ 0, ..., m-1\}$ denote the $m$ depots and $[x_i, y_i], i\in \{m,..., m + n - 1\}$, the $n$ customer nodes. Note that this setting can be expanded to consider more realistic distributions as in \citep{bi2022learning, zhou2023towards, gao2023towards}, and our implementation is already set up in such a way to allow for different distributions in the future via the \texttt{get\_sampler} method.

\paragraph{Multiple depots (MD)}
    Depot nodes, in principle, have the same node attributes as customer nodes. The location, however, is the only attribute that is generated in the same way, which is explained below. For all other attributes, the values are fixed and identical for all depots. Linehaul and backhaul demands, as well as service durations, are set to zero, while the time windows of all depots in an instance are set to $[e_i, l_i] = [0, t_{\mathit{max}}], i \in \{0, ..., m-1\}$, where $t_{\mathit{max}}$ denotes the system end time and $M$ the number of depots. For problems without time windows, $t_{\mathit{max}}$ is set to $\infty$. In the unseen variants experiments of \cref{subsec:finetuning-appendix}, we employ $m=3$ depots for the MD finetuning variants.

\paragraph{Vehicle capacity (C)}
The vehicle capacity $C$ is a fixed value applied to all vehicles and calculated according to: 
\[
C =
\begin{cases} 
      30 + \left\lfloor \frac{1000}{5} + \frac{n - 1000}{33.3} \right\rfloor & \text{if } 1000 < n \\
      30 + \left\lfloor \frac{n}{5} \right\rfloor & \text{if } 20 < n \leq 1000 \\
      30 & \text{otherwise}
\end{cases}
\]

which is commonly used in NCO for VRP approaches \citep{kool2018attention, kwon2020pomo}.

\paragraph{Linehaul and backhaul demands (C) / (B) / (MB)}
We generate demands according to the following scheme:
\begin{enumerate}
    \item Generate linehaul demands $q_i$ for all customers $i \in N_c$ by sampling uniformly from the set of integers $\{1, 2, ..., 9\}$.
    \item Generate backhaul demands $p_i$ for all customers $i \in N_c$ by sampling uniformly from the set of integers $\{1, 2, ..., 9\}$. 
    \item For each customer $i \in N_c$, generate a temporary decision variable $z_i \in \{0,1\}$ with probabilities $\mathbb{P}(z_i = 0) = 0.8$ and $\mathbb{P}(z_i = 1) = 0.2$.
    \begin{itemize}
        \item If $z_i = 0$, keep the linehaul demand $q_i$ and set the backhaul demand $p_i = 0$.
        \item If $z_i = 1$, set the linehaul demand $q_i = 0$ and keep the backhaul demand $p_i$.
    \end{itemize}
\end{enumerate}
This demand generation scheme ensures that each customer has either a linehaul demand or a backhaul demand, but not both. With a probability of 0.8, a customer will have only a linehaul demand, and their backhaul demand will be set to 0. Conversely, with a probability of 0.2, a customer will have only a backhaul demand, and their linehaul demand will be set to 0. It is important to note that not all customers are typically backhaul customers, even in a backhaul setting. Therefore, this scheme allows for the consideration of both linehaul and backhaul demands in backhaul problem settings while ensuring that each customer has only one type of demand.

We note that this can be easily extended to the case of VRP with simultaneous pickup and delivery (VRPSPD), in which a customer can have both linehaul and backhaul demand \citep{ai2009particle, kocc2020review}. In such a case, we could duplicate the customer node into two nodes with the same attributes, such as locations, but different values for linehaul (pickup) and backhaul (delivery) in the current VRP environment or allow for both linehaul and backhaul to be present at the same time in a single node with small modifications in the action masking.

\paragraph{Backhaul class (B) / (MB)}
For testing the few-shot setting described in \cref{sec:main-paper-eal}, we generate instances with \textit{mixed} backhauls. The instances themselves are actually identical to instances with the \textit{traditional} backhaul, and we use a global attribute in the instance to differentiate between them. For this purpose, we allow either setting a fixed value $\in \{1,2\}$ or sampling from $\{1,2\}$ for every customer with equal probabilities $p(1)=p(2)=0.5$, allowing for different backhaul settings within one batch, if needed (see the batching procedure described in \cref{subsec:mixed-batch-training}). Note that we sample from $\{1,2\}$ instead of boolean sampling because we plan to extend the number of backhaul settings in the future.

\paragraph{Open routes (O)}
For open routes, we generate a boolean vector with all \verb|True| values. During sampling (see \cref{subsec:mixed-batch-training}), the actual ratio of open route instances is defined, not at the initial instance generation (i.e., we temporarily change the \verb|True| value to \verb|False| for every batch element with a certain probability).

\paragraph{Time Windows (TW)}
We generate the time windows $[e_i, l_i]$ and service times $s_i$ in several steps for all customers $i \in N_c$: 
\begin{enumerate}
    \item Generate service times $s_i \in [0.15, 0.18]$.
    \item Generate time window lengths $t_i \in [0.18, 0.2]$.
    \item Calculate \rebuttal{the maximum distance from any of the depots $j \in \{0,...,m-1\}$ to customer $i$: $d_{\mathit{max}} = \max_j(d_{ij})$}.
    \item Calculate upper bounds for time window start times $h_i = \frac{t_{\mathit{max}} - s_i - t_i}{d_{\mathit{max}}} - 1$.
    \item Calculate time window start times as $e_i = (1 + (h_i - 1) \cdot u_i) \cdot d_{\mathit{max}}$ with $u_i \sim U(0,1)$.
    \item Calculate time window end times as $l_i = e_i + t_i$.
\end{enumerate}

When calculating the action mask, we have the constraint that the expected arrival time should be earlier than the end time of nodes; if the problem is a closed problem, we should also consider the time back to the depot,  i.e., \(\max(t_{\mathrm{curr}} + d_{ij}, e_j) + s_j + d_{\mathit{max}} < l_0\). We note that for simplicity, we set the vehicle speed to $1.0$ in equations and normalize time windows accordingly so that travel time from two nodes is the same numerically as the distance between them. This can be easily modified in the code.

We mention as an alternative TW generation procedure the one from the Solomon benchmark \citep{solomon1987algorithms, li2021learning}, which may perform better in that benchmark, as done in \citet{zhou2024mvmoe}.

\paragraph{Distance limit (L)}
The distance limit is sampled from a uniform distribution to ensure meaningful and feasible constraints. Specifically, we sample $L$ from $U(2 \cdot \max(d_{0i}), l_\text{max}))$, where $d_{0i}$ is the distance from the depot to customer $i$, and $l_\text{max} = 3.0$ is a predefined upper bound. 
This approach ensures that $L$ is always greater than the round trip to the farthest customer $(2 \cdot \max(d_{0i}))$, making all customers reachable, while also allowing for variation in the constraint tightness. 
For the multi-depot case we replace $\max(d_{0i})$ with $\min_j(\max_i(d_{ij})), i \in \{m, ..., m+n\}, j \in \{0,...,m\}$, i.e., we first get the maximum distance from any customer node to each of the depots and then take the minimum out of those distances. By taking the maximum in the first step we ensure that all customers are reachable, and by taking the minimum across depots, we make the problem more challenging, because even though all nodes can in principle be serviced, some may only be serviced by one (or a subset) of the available depots.
This sampling method produces more variation than previous works \citet{liu2024multi, zhou2024mvmoe} (where there was virtually no difference in solutions of (L) and non-(L) variants), as it guarantees feasible instances while still providing a range of challenging scenarios.

\paragraph{Attribute Normalization and Scaling}
All demands, both linehauls and backhauls, are scaled to lie in $[0,1]$ through division by the vehicle capacity. $q'_i = q_i / C, p'_i = p_i / C$. All other features are already sampled from a normalized range.  Note that during loading instances from e.g. CVRPLib, we normalize features before passing them to the policy - for instance, locations are normalized between 0 and 1.

\subsection{Environment Logic}
\label{sec:mtvrp-environment-logic}

To determine available actions for the Unified VRP environment formulation, the constraints for the individual problems have to be combined in the action mask (\verb|action_mask| in the code following RL4CO, where \verb|True| means that the action is feasible \citep{berto2024rl4co}). We build a logical test structure, essentially separating the checks in the action mask according to the individual VRP problem types and then bringing them all together again. The individual \verb|action_mask| checks are the following:

\begin{enumerate}[leftmargin=*]
\renewcommand{\labelenumi}{\alph{enumi})}
    \item \textit{Can reach in time}: depending on the current time and the travel distance to every node not yet visited, can we reach that node before its service time window ends? $t_{\mathrm{curr}} + d_{ij} < l_j$, where $t_{\mathrm{curr}}$ is the current time.
    \item \textit{Does not exceed distance limit}: depending on the current length of the route, if we travel to any available node, will we exceed the total distance limit for the route? $l_{\mathrm{curr}} + d_{ij} < L$, where $l_{\mathrm{curr}}$ is the current length.
    \item \textit{Can reach depot}: there are two types of constraints from time windows (TW) and distance limit (L):
    \begin{itemize}
        \item If we need to ensure we can reach the depot in time, i.e., the current time plus traveling time to the depot must be smaller than the system end time: $\max(t_{\mathrm{curr}} + d_{ij}, e_j) + s_j + d_{j0} < \rebuttal{t_{\mathit{max}}}$. 
        \item If we need to ensure we can reach the depot without exceeding the distance limit, i.e., the current distance plus the traveling distance to the depot must be smaller than the distance limit: $l_{\mathrm{curr}} + d_{ij} + d_{j0} < L$. 
    \end{itemize}
    For the multi-depot case we replace $d_{j0}$ in both these constraints with $d_{jk}$, where $k \in \{0,..., m-1\}$ indexes the depot the current route \textit{started} from.
    For open routes, this will always be set to \verb|True|, i.e., this constraint does not apply. 
    \item \textit{Demand constraints for backhaul problems}:
    \begin{itemize}
        \item Checks for \textit{all} backhauls problems:
        \begin{itemize}
            \item Does the linehaul demand exceed vehicle capacity if we add a node's demand to the current vehicle? $c_{\mathrm{curr}} + q_{j} < C$, where $c_{\mathrm{curr}}$ is the used capacity.
            \item Does the backhaul demand exceed vehicle capacity if we add a node's demand to the current vehicle? $c_{\mathrm{curr}} + p_{j} < C$, where $c_{\mathrm{curr}}$ is the used capacity.
        \end{itemize}
        \item Checks for traditional backhaul settings:
        \begin{itemize}
            \item Carrying backhaul: if we are already picking up backhaul demands, we cannot service any linehaul demands on this route anymore.
            \item If we are not carrying backhaul demands yet, are there any unserved linehaul demands left?
            \item If there are no linehaul demands left or we are already carrying backhauls, are there still unserved backhaul demands?
        \end{itemize}
        \item Checks for \textit{mixed} backhaul settings:
        \begin{itemize}
            \item Cannot service linehaul demands: depending on the backhaul demands currently loaded in the vehicle, do we have space left for further linehaul demands?
        \end{itemize}
        We additionally remark that our definition of backhauls follows the generally accepted definition in the OR community, originally due to~\cite{goetschalckx1989vrpb}. This definition differs from the routing problems with backhaul considered in several recent papers in the machine learning (e.g.,~\cite{liu2024multi, zhou2024mvmoe}), who define backhaul customers as having a negative demand of the same commodity used for linehaul, and do not consider the precedence constraint that all linehaul must be completed before backhaul may start on the route. The problem setting with a single commodity is not commonly studied in the OR literature since it implies pickups may be used for deliveries at later customers, while the relaxation of the precedence constraint is more properly referred to as a \textit{mixed} backhaul problem~\citep{koc2018vrpb}.
    \end{itemize}
    \item \textit{Already visited}: every customer node needs to be visited exactly once.
\end{enumerate}
We bring together checks a) to e) and introduce an additional check for the depot: if we are currently in the depot and there are still unserved customers, we cannot select the depot as the next action to ensure the model cannot get stuck during decoding. 

    For the multi-depot case we further extend this check. If we are currently in a depot and there are unserved customers, we cannot visit \textit{any} depot. If no further customers can be serviced, all depots are available actions again. However, if we are currently in a depot and no customers can be served from this depot, we mask it out so as to service the remaining customers from the remaining depots that can actually service them.

Combining these checks in this way allows us to meticulously check for individual VRP settings while at the same time maintaining the necessary flexibility the unified environment formulation requires.

\section{\our{} Model Details}
\label{append:model}

\our{} follows the encoder-decoder architecture from the Attention Model \citep{kool2018attention}, a transformer-like architecture based on the attention mechanism \citep{vaswani2017attention}. We additionally improve the encoder architecture in \textsc{RF}-TE as explained in \cref{sec:routefinder_model}. We focus the explanation on modeling \textit{all} attributes possible with the MDOVRPMBLTW, noting that in the main training runs, we do so without considering attributes from multi-depots and mixed backhaul, whose additional parameters are added upon EAL finetuning.

\subsection{Multi-Head Attention}

At the core of \our{} lies the Multi-Head Attention (MHA) mechanism, proposed by \citet{vaswani2017attention}. MHA concurrently attends to information from various representation subspaces, facilitating the capture of diverse relationships between input elements. Notably, MHA is capable of handling a variable number of elements.

The MHA operation starts by linearly projecting the input sequences of queries $Q$, keys $K$, and values $V$ to $H$ distinct subspaces using learned projection matrices $W_i^Q$, $W_i^K$, and $W_i^V$, respectively, where $H$ denotes the number of attention heads: $Q_i = Q W_i^Q$, $K_i = K W_i^K$, $V_i = V W_i^V$ for $i = 1, \dots, H$. Subsequently, the attention weights for each head are computed by performing a scaled dot product between the projected queries and keys, followed by a softmax operation:
\begin{equation}
A_i = \text{Softmax}\left(\frac{Q_i K_i^T}{\sqrt{d_k}} + M\right)
\end{equation}
where $d_k$ represents the dimension of the keys,  acting as a scaling factor to prevent the dot products from growing too large, $\text{Softmax}(x_i) = \frac{\exp(x_i)}{\sum_{j=1}^N \exp(x_j)}$ and $M$ is an optional attention mask that can be used to prevent attending to certain positions (e.g., infeasible actions), which can be done by setting elements to $-\infty$. The output of each attention head is then calculated as a weighted sum of the projected values, using the attention weights: $Z_i = A_i V_i$.

Lastly, the outputs from all attention heads are concatenated and linearly projected using a learned matrix $W^O$ to yield the final output of the MHA operation:
\begin{equation}
\text{MHA}(Q, K, V) = \text{Concat}(Z_1, \dots, Z_H) W^O
\end{equation}
While the MHA grows quadratically, i.e., with sequence length (i.e., number of nodes) $N$, it grows as $O(N^2)$, several efficient implementations have been proposed over the years, and we use FlashAttention \citep{dao2022flashattention, dao2023flashattention} to speed up the model.

\subsection{Encoder}
\label{append:encoder}
The Encoder transforms an input instance $\bm{x}$ into a hidden embedding $\bm{h}$. The Encoder architecture consists of the following main components: 1) Global Embedding, 2) Node Embedding, and 3) a series of Encoder Layers. We consider a VRP instance of $n$ locations as having $n+1$ nodes, where node $0$ is the depot and nodes $\{1, \dots, n\}$ are $n$ customers. For problems with multiple depots, we define $m$ as the number of depots, i.e., nodes $\{ 0, \dots, m-1\}$ are the depot nodes, and $m, \dots, m+n-1$ are the $n$ customer nodes.

\paragraph{Global Embedding} 
Since Global Attributes contain a single value for all the $m+n$ problem nodes, we embed them in depot nodes, in a similar fashion to how traditional solvers as PyVRP encode information about the global problem structure on depot nodes.. Global Embeddings include global attributes Open Routes $o \in \{0, 1\}$, Duration Limits $l \in [0, L]$, and Mixed Backhauls flag $\mu \in \{0, 1\}$, as well as
\rebuttal{
    the locations of the depot node(s) $[x_i, y_i] \in \mathbb{R}^2, i \in \{0, \dots, m-1\}$ and the system end time $l_{\text{max}}$ (i.e., the depot(s) time window). In practice, for the multi-depot case with $m>1$, the global attributes are projected on the depot nodes.
}
In \our{}, the global embedding $f$ is a linear projection layer $\mathbf{W}_g \in \mathbb{R}^{k \times d}$ where $k=6$ features and $d=128$ is the hidden dimension. The initial projected global hidden embedding per depot $g_i$ can be written as \rebuttal{$\bm{h}^{(0)}_{g_i} = \mathbf{W}_g [x_i, y_i, l_{\text{max}}, o, l, \mu]^\top$}. 

\paragraph{Node Embedding} The node embeddings, on the other hand, capture customer-specific attributes and are projected onto the remaining $n$ nodes. These attributes include for nodes 
$i \in \{m, \dots m + n-1\}$: Linehaul demands $q_i \in [0, Q]$, Time Windows parameters $e_i, s_i, l_i \in [0, T]^3$ where $e$ and $l$ denote the time window's start and end and $s$ is the service time, the Backhaul demands $p_i \in [0, Q]$, and finally the node locations $[x_i, y_i] \in \mathbb{R}^2$. In \our{} this a linear projection layer $\mathbf{W}_n \in \mathbb{R}^{k \times d}$ where $k=7$ features and $d=128$ is the hidden dimension. The initial projected node hidden embedding can be written for each node $n_i$ as $\bm{h}^{(0)}_{n_i} = \mathbf{W}_n [x_i, y_i, q_i, e_i, s_i, l_i, p_i]^\top$.

\paragraph{Raw Features to Hidden States} The projected global embedding and node embeddings are concatenated to obtain the initial hidden representation $\bm{h}^{(0)} \in \mathbb{R}^{(m + n) \times d}$, where $m + n$ is the total number of nodes ($m$ depots + $n$ customers) and $d$ is the hidden dimension:
\begin{align}
\\bm{h}^{(0)} = \text{Concat}( \bm{h}^{(0)}_{g_1}, \dots, \bm{h}^{(0)}_{g_m}, \bm{h}_{n_1}^{(0)}, \dots, \bm{h}_{n_{n}}^{(0)})
\end{align}
The initial hidden representation $\bm{h}^{(0)}$ is then passed through a series of Encoder Layers to refine and enrich the representation. Each Encoder Layer consists of a Multi-Head Attention (MHA) layer and a Multi-Layer Perceptron (MLP) layer, as described in \cref{eq:am-mha} and \cref{eq:am-mlp}, respectively.

The Encoder can be represented as:
\begin{align}
\label{eq:autoregressive_encoder}
\bm{h} = \text{EncoderBlocks}(\bm{h}^{(0)})
\end{align}
Each EncoderBlock consists of two sub-layers: a Multi-Head Attention (MHA) layer and a Multi-Layer Perceptron (MLP) layer (or SwiGLU as we propose). The MHA layer allows the model to capture dependencies between different positions in the input sequence, while the MLP layer applies non-linear transformations to the features at each position. The input to each EncoderBlock is first passed through the MHA layer, which computes the self-attention using the input as queries, keys, and values:
\begin{align}
\hat{\bm{h}} &= \text{Norm} \left(\bm{h}^{(\ell-1)} + \text{MHA}(\bm{h}^{(\ell-1)}, \bm{h}^{(\ell-1)}, \bm{h}^{(\ell-1)})\right) \label{eq:am-mha}
\end{align}
where $\bm{h}^{(\ell-1)}$ represents the input to the $\ell$-th EncoderBlock, and $\text{Norm}$ denotes a normalization operation, in \our{} we employ Instance Normalization (IN). The output of the MHA layer, $\hat{\bm{h}}$, is then passed through the MLP layer, which applies a series of linear transformations with non-linear activations:
\begin{align}
\bm{h}^{(\ell)} &= \text{Norm} \left(\hat{\bm{h}} + \text{MLP}(\hat{\bm{h}})\right) \label{eq:am-mlp}
\end{align}
The pointwise MLP layer consists of two linear layers with a non-linear activation function as ReLU, between them. 

\paragraph{Transformer-based Encoder}

We further explicit our proposed Transformer-based encoder. Each EncoderBlock consists of two sub-layers: a Multi-Head Attention (MHA) layer and a Feed Forward SwiGLU layer \citep{shazeer2020glu}. The MHA layer captures dependencies between different positions in the input sequence, while the SwiGLU layer applies non-linear transformations to the features. We employ RMS normalization \citep{zhang2019root} and pre-norm architecture for improved stability and faster convergence:
\begin{align}
\hat{\bm{h}} &= \bm{h}^{(\ell-1)} + \text{MHA}(\text{RMSNorm}(\bm{h}^{(\ell-1)}), \text{RMSNorm}(\bm{h}^{(\ell-1)}), \text{RMSNorm}(\bm{h}^{(\ell-1)})) \label{eq:rf-mha} \\
\bm{h}^{(\ell)} &= \hat{\bm{h}} + \text{SwiGLU}(\text{RMSNorm}(\hat{\bm{h}})) \label{eq:rf-swiglu}
\end{align}
where $\bm{h}^{(\ell-1)}$ represents the input to the $\ell$-th EncoderBlock. The SwiGLU MLP \citep{shazeer2020glu} is defined as:
%
\begin{align}
    \text{SwiGLU}(x) = \bf{W}_3 ( \text{SiLU}(\bf{W_1} x ) \odot (\bf{W_2}x)
\end{align}
where $\odot$ denotes the Hadamard product, SiLU is the activation function \citep{elfwing2018sigmoid}, and $\bf{W}_1$, $\bf{W}_2$ and $\bf{W}_3$ are learnable weight matrices\footnote{In our code this is the \texttt{ParallelGatedMLP} inspired by the StripedHyena architectures \citep{ku2025systems}.}.  We use FlashAttention \citep{dao2022flashattention, dao2023flashattention} in the MHA layer for enhanced performance.

\subsection{Decoder} 
\label{append:decoder}

The Decoder autoregressively constructs the solution based on the Encoder output $\bm{h}$ and the state $s_t$ at the current step $t$.

\paragraph{Context Embedding} The context embedding is used to modify the query embedding of the problem node of the current partial solution. It consists of a linear layer that projects the concatenated current node embedding and state embedding to the embedding space.  The state embedding is computed by projecting the following: the current node embedding $\bm{h}_t$ and a set of dynamic features from state $s_t$, i.e. the available load $c_t$, current time $t_t$, current distance traveled $d_t$, the available backhaul load $b_t$ -- i.e. the difference between the vehicle capacity $Q$ and the \textit{used backhaul capacity}, which is necessary because if we pick up items, the deliverable quantity must exceed the remaining capacity after pick up for mixed backhauls (MB) -- as well as the location of the origin depot $o$ we have to return to at step $t$: $[x^o_t, y^o_t]$ for the multi-depot variants (MD). In \our{} the context embedding  $\mathbf{W}_{c} \in \mathbb{R}^{d \times (d + k)}$ is a linear projection matrix, $d=128$ is the hidden dimension, and $k=6$ is the number of state features.  The context embedding at step $t$ is thus computed as $\mathbf{h}_{c}^{(t)} =\mathbf{W}_{c}  \text{Concat}([\bm{h}_{t}; [c_t, t_t, d_t, b_t, x^o_t, y^o_t]])^\top$.

\paragraph{Attention and Pointer Mechanism}
The query $q_t$ is obtained directly from the context embedding $q_t = \mathbf{h}_{c}^{(t)}$ and then passed into a masked MHA layer and final single-head attention to obtain logits $\bm{z}$:
\begin{align}
h^c_t &= \text{MHA}(q_t, {K}^g_t, {V}^g_t, M_t), \label{eq:am-pointer-mechanism-init} \\
\bm{z} &= \frac{{V}_t^p h^c_t}{\sqrt{d_k}} \label{eq:am-pointer-mechanism}
\end{align}
where $M_t$ is the set of feasible actions (i.e., the \verb|action_mask|), and projections ${K}^g_t, {V}^g_t, {V}_t^p = W_k^g \bm{h}, W_v^g \bm{h}, W_v^p \bm{h}$ are precomputed once as cache. We note that \cref{eq:am-pointer-mechanism} is usually referred to as the pointer mechanism \citep{vinyals2015pointer}. 

\paragraph{Logits processing}

Finally, logits $\bm{z}$  are transformed into a probability distribution:
 \begin{equation}
     p = \text{Softmax} \left(    C \cdot \text{tanh} (\bm{z})  \right)
     \label{eq:am-logits-processing}
 \end{equation}
 where logits for infeasible actions can be masked, and $C$ is the \textit{tanh clipping} that serves in improving the exploration, which we set to 10 according to \citet{bello2016neural}.

 \paragraph{Action selection} During training, we use the POMO \texttt{multistart} sampling. 
 
    For the multi-depot case we force the first action to start from all depots in the instance. For the single-depot case we force the first action to start with every customer node to maximize diversity. Note that if \texttt{num\_starts} is not divisible by the number of depots $m$, the resulting tensor will not have an equal number of indices for each depot, i.e., the number of starts will not be distributed evenly across the depots, as we use the modulo operator for the assignment.

 During testing, we also employ \texttt{multistart} but with greedy selection (i.e., selecting the maximum probability). Prior to the selection, a dihedral augmentation is also performed prior to encoding instance $\bm{x}$ in the encoder, which enables exploring $8\times$ as many solutions with 4 rotations $\times$ 2 flips. We note that additional augmentations and techniques can be performed during inference, which can further boost evaluation performance \citep{kim2022sym, ma2022efficient, choo2022simulation, luo2024neural}. For fairness of comparison, we do not employ additional augmentations but assume that this could further boost the performance of \our{}.

\subsection{EAL Modeling}
\label{append:eal-modeling-details}

We describe in more detail the procedure for Efficient Adapter Layers (EAL) modeling. Our initial model trained from \cref{subsec:experiments-main-results} has linear projections layers as referenced in full detail in \cref{append:encoder} and \cref{append:decoder} without additional parameters for mixed backhaul and multi-depots.

\subsubsection{EAL for mixed backhauls} 
\label{append:eal-modeling-details-mb}

This adds, as explained in \cref{subsec:efficient-adapter-layers-eal}, a single ($l=1$) parameter row $\mathbf{W}'_0$ for the mixed backhaul flag $\mu$ to the global embedding. Moreover, we add $l=1$ rows for the context embedding resulting $\mathbf{W}_c'$ for the available backhaul load $b_t$ at step $t$, i.e. the difference between the vehicle capacity $Q$ and the \textit{used backhaul capacity}.

\subsubsection{EAL for multi-depots}
\label{append:eal-modeling-details-md}

In this case, we do not modify the global embedding but directly project multiple times global attributes and depot locations at each depot node as explained in \cref{append:encoder}. However, we modify the context embedding $\mathbf{W}_c'$ by adding $l=2$ rows to keep track of the location of the origin depot $o$ we have to return to at step $t$: $[x^o_t, y^o_t]$.

\subsubsection{EAL for multi-depots \& mixed backhauls}
\label{append:eal-modeling-details-md-mb}

Here we combine the EAL implementations of the previous two paragraphs. We add the $l=1$ parameter row $\mathbf{W}'_0$ for the mixed backhaul flag $\mu$ to the global embedding and project the global embedding $m$ according to the number of depots and modify the context embedding $\mathbf{W}_c'$ by adding $l=3$ rows to keep track of the available backhaul load $b_t$ and the location of the origin depot $[b_t, x^o_t, y^o_t]$.

\paragraph{}

\section{Additional Material}

\subsection{Details for Average Batch Reward for Multi-task Reward Normalization}
\label{app:avg_batch_reward}

At each training step $t = 1, \dots, T$ we train on a batch of $b = 1, \dots , B$ problem instances, each of which belongs to one of the $k \in K$ problem variants covered by \our{}.
Let $\mathbbm{1}_{b,k} \in \{0,1\}$ be an indicator function such that:
\begin{align*}
\mathbbm{1}_{b,k} = 
    \begin{cases} 
        1 & \text{if instance } b \text{ is of type } k \\
        0 & \text{otherwise}
    \end{cases}
\end{align*}
which is efficiently calculated in our unified VRP environment based on vectorized checks.
The reward $r_{bt}^{(k)}$ for instance $b$ of variant $k$ at training step $t$ can then be expressed as $r_{bt}^{(k)} = r_{bt} \cdot \mathbbm{1}_{b,k}$.
The average batch reward $\bar{r}_t^{(k)}$ for variant $k$ at training step $t$ over all instances of type $k$ in a batch can then be expressed as:
\begin{align*}
    \bar{r}_t^{(k)} 
        = \frac{\sum_{b=1}^{B} r_{bt}^{(k)}}{\sum_{b=1}^{B} \mathbbm{1}_{b,k}} 
        = \frac{\sum_{b=1}^{B} r_{bt} \cdot \mathbbm{1}_{b,k}}{\sum_{b=1}^{B} \mathbbm{1}_{b,k}} ,
        \qquad \forall k \in K.
\end{align*}
This average batch reward $\bar{r}_t^{(k)}$ is the basis for the reward normalization explained in \cref{sec:reward-normalization}.

\subsection{Hyperparameter Details}

We report in \cref{tab:hyperparameters} the hyperparameter details common across the main experiments. \our{} variants additionally employ the proposed contributions as outlined in the main experiments of \cref{subsec:experiments-main-results}. 

\begin{table}[htbp]
\centering
\caption{Experiment hyperparameters. Values with ``/'' indicate different choices depending on the model, i.e., on the right are values for the Transformer-Based encoder.}
\label{tab:hyperparameters}
\vspace{2mm}
\begin{tabular}{ll}
\toprule
\textbf{Hyperparameter} & \textbf{Value} \\
\midrule
\multicolumn{2}{l}{\textit{Model}} \\
Embedding dimension & 128 \\
Number of attention heads & 8 \\
Number of encoder layers & 6 \\
Use Pre-norm & False / True \\
Normalization & Instance / RMSNorm \\
Feedforward hidden dimension & 512 \\
Feedforward structure & MLP / Gated MLP \\
Feedforward activation & ReLU / SwiGLU \\
Tanh clipping & 10.0 \\
Mask logits & True \\
\midrule
\multicolumn{2}{l}{\textit{Training}} \\
Train decode type & multistart sampling \\
Val \& Test decode type & multistart greedy \\
Augmentation function & dihedral \\
Batch size & 256 \\
Train data per epoch & 100,000 \\
Reward normalization & Exponentially smoothed mean \\
Normalization $\alpha$ & 0.25 \\
\midrule
\multicolumn{2}{l}{\textit{Optimization}} \\
Optimizer & Adam \\
Learning rate & 3e-4 \\
Weight decay & 1e-6 \\
LR scheduler & MultiStepLR \\
LR milestones & [270, 295] \\
LR gamma & 0.1 \\
Gradient clip value & 1.0 \\
Max epochs & 300 \\
\bottomrule
\end{tabular}
\end{table}

\subsection{Additional Discussion}

\paragraph{Motivation} Foundation models have been successful in several areas in recent years, including large language models \citep{achiam2023gpt}, computer vision \citep{kirillov2023segment} as well as other domains such as biology \citep{abramson2024accurate, nguyen2024sequence}. However, foundation models for discrete decision-making, such as CO and our target VRPs, are still under-explored as an area - one reason being the lack of large, high-quality open datasets that can effectively be employed to train such models - which motivates our use of RL. Such foundation models may not only obtain solutions faster than traditional OR counterparts but also avoid the requirement of possibly decades of research and resources to tackle a single task, while a foundation model may automatically learn heuristics without supervision.

\paragraph{Generalist, or specialized?} Another open question is the idea of generality behind the model. In \our{}, we argue that a model might not need to be extremely complex and be specialized for a specific application (such as routing). One such reason is that with larger model capabilities comes larger size and inference time, which is crucial for real-world deployment. An interesting future direction would be to attempt to generalize a model as a "foundation model for CO", for instance, based on a general formulation \citep{boisvert2024towards}, and see whether the additional training and inference costs are worth a (possible) boost in optimality gaps and generalization ability. Such a model may be able to attain a better few-shot generalization to totally unseen attributes, either with adapter layers \citep{lin2024cross} or with our proposed EAL. However, we believe that tailored, specialized foundation models as \our{} for VRPs may be more practical and efficient. We note that an orthogonal direction to ours is the use of LLMs as hyper-heuristics \citep{romera2024mathematical, fei2024eoh, ye2024reevo}, which starts from a generalist LLM agent to generate algorithms that can be used to improve the optimization of CO problems as VRPs. However, such models are not used at inference time due to the inefficiency of using billions of parameters that are not tailored for the problem at hand.

\paragraph{Going forward} in specialized foundation models for VRPs, there are several challenges yet to be addressed. One such challenge is the still sub-par performance compared to state-of-the-art solvers \citep{wouda2023alns, wouda2024pyvrp}, which may be offset on a larger scale by several means, including decompositions. Another way to attain better performance would be to integrate with local search \citep{ye2024deepaco, kim2024gfacs_ant_colony_optimization} and hybridize constructive (the current policy paradigm) with improvement methods \citep{ma2021learning, ma2024learning} to guarantee monotonic improvements given larger time budgets. Finally, given the robust cross-task performance even compared to single-task models, we believe expanding to more VRP variants (and their attribute distributions) may further improve overall performance.

\subsection{Licenses for used assets}
\label{app: licenses for used assets}

\cref{tab:asset} lists the used assets and their licenses. Our code is licensed under the MIT License.

\begin{table}[htbp]
  \centering
  \setlength\tabcolsep{9pt}
  \caption{Used assets and their licenses.}
  \vspace{4pt}
    \resizebox{0.8\textwidth}{!}{
    \begin{tabular}{cccc}
    \toprule
    Type  & Asset & License & Usage \\
    \midrule
    \multirow{6}[2]{*}{Code}
          & POMO \citep{kwon2020pomo}   & MIT License & Evaluation \\
          & MTPOMO \citep{liu2024multi}   & MIT License & Evaluation \\
          & MVMoE \citep{zhou2024mvmoe}   & MIT License & Evaluation \\
          & RL4CO \citep{berto2024rl4co}   & MIT License & Evaluation \\
          & AL \citep{lin2024cross} & MIT License & Evaluation \\
          & ORTools \citep{cpsatlp}   & Apache-2.0 & Evaluation \\
          & PyVRP \citep{wouda2024pyvrp}   & MIT License & Evaluation \\
    \midrule
    Dataset
     & CVRPLIB \citep{lima2014cvrplib} & Available for any non-commercial use & Testing \\
    \bottomrule
    \end{tabular}%
    }
  \label{tab:asset}%
\end{table}%

\section{Additional Empirical Results}
\label{append:experiments-results}

This Section supplements the main paper with several experiments evaluating various aspects of \our{}:

\begin{itemize}
    \item \cref{append:out-of-distribution-attribute-generalization}: here we motivate our \our{} foundation model for VRPs when compared to single-variant models in 1) finetuning performance and 2) out-of-distribution generalization.
    \item \cref{sec:cvrplib}: we evaluation large-scale and real-world distributions in CVRPLIB.
    \item \cref{append:effect-transformer-components}: we study the effect and interactions of Transformer Encoder components.
    \item \cref{append:effect-normalization-scheme}: here, we show the effect of different reward normalization techniques.
    \item \cref{subsec:append-mixed-batch-training-study}: here we study Mixed Batch Training and its effect on 1) training stability and 2) imbalanced variant distributions.
    \item \cref{append:t-sne-visualization-interpretability}: we study the latent learning representation ability of different models via t-SNE across 1) encoding layers 2) effect of different attributes on the latent embeddings.
    \item \cref{subsec:finetuning-appendix}: this section adds additional experiments for zero-shot and finetuning performances with EAL on three unseen new attribute setups: 1) with mixed backhauls 2) with multi-depots and 3) with both mixed backhauls and multi-depots as well as comparison to finetuning with single-variant models.
\end{itemize}


\subsection{Out-of-Distribution Attribute Generalization} 
\label{append:out-of-distribution-attribute-generalization}

In this section, we study our foundation model and ask the following question: how does \our{} perform when compared to models trained specifically on a single variant?  To answer this question, we compare \our{} and other multi-task learning methods with POMO trained on single variants, including CVRP, VRPL, VRPTW, OVRP, and VRPB. For fairness of comparison, we train the POMO models with the same hyperparameters as the other models (from \cref{tab:hyperparameters}), including the same batch size, learning rate, and training epochs on $n=100$ nodes.

We also study out-of-distribution generalization for unseen attribute values of 
time windows (C) and duration limits (L), for multi-task learning models and single-variant POMO ones. We compare cost values and gaps (the lower, the better) to the results of POMO training specifically for that single variant, similarly to \citet[Appendix D]{liu2024multi}. All experiments are performed on $1000$ variants for each setting with $n=100$.



In VRPTW, we consider different values of the time interval, i.e., the minimum and maximum values from which service times $s_i$ and time window lengths $t_i$ are sampled (points 1 and 2 for time window generation of \cref{sec:mtvrp-data-generation}). In distribution, these values are sampled from $[0.15, 0.20]$. In the out-of-distribution settings, we consider them as $\{ [0.05, 0.1], [0.15, 0.20], \dots, [0.85, 0.9], [0.85, 1.0] \}$. The results in \cref{tab:ood-vrptw} demonstrate again that for values differing from the in-training distribution, our model obtains better results than POMO trained solely on VRPTW.

\begin{table*}[h!]
\caption{\rebuttal{Comparison of our model with single-task POMO on out-of-distribution VRPTW instances.}}
\label{tab:ood-vrptw}
\begin{center}
\renewcommand\arraystretch{1.05}
\rebuttal{
\large %
\renewcommand{\arraystretch}{1.05} %
\resizebox{\textwidth}{!}{ 
\begin{tabular}{l|cccccccccccccccccccc}

\toprule

Time Interval & \multicolumn{2}{c}{[0.05, 0.10]} & \multicolumn{2}{c}{[0.15, 0.20]} & \multicolumn{2}{c}{[0.25, 0.30]} & \multicolumn{2}{c}{[0.35, 0.40]} & \multicolumn{2}{c}{[0.45, 0.50]} & \multicolumn{2}{c}{[0.55, 0.60]} & \multicolumn{2}{c}{[0.65, 0.70]} & \multicolumn{2}{c}{[0.75, 0.80]} & \multicolumn{2}{c}{[0.80, 0.85]} & \multicolumn{2}{c}{[0.95, 1.00]} \\
\cmidrule(lr){2-3} \cmidrule(lr){4-5} \cmidrule(lr){6-7} \cmidrule(lr){8-9} \cmidrule(lr){10-11} \cmidrule(lr){12-13} \cmidrule(lr){14-15} \cmidrule(lr){16-17} \cmidrule(lr){18-19} \cmidrule(lr){20-21}

 & Obj. & Gap & Obj. & Gap & Obj. & Gap & Obj. & Gap & Obj. & Gap & Obj. & Gap & Obj. & Gap & Obj. & Gap & Obj. & Gap & Obj. & Gap \\

\midrule
      
POMO\_VRPTW     & \textbf{25.30}   &  \textbf{*} & \textbf{26.27}   &  \textbf{*} & \textbf{28.11}   &  \textbf{*} & 31.36   &  * & 35.25   &  * & 39.66   &  * & 44.43   &  * & 48.17   &  * & 52.60   &  * & 55.24   &  * \\
MTPOMO   & 25.51   &  0.84\% & 26.59   &  1.20\% & 28.27   &  0.57\% & 31.28   & -0.26\% & 35.05   & -0.56\% & 39.51   & -0.39\% & 44.34   & -0.21\% & 48.25   &  0.17\% & 52.85   &  0.47\% & 55.67   &  0.78\% \\
MVMoE    & 25.47   &  0.66\% & 26.57   &  1.15\% & 28.25   &  0.50\% & 31.19   & -0.54\% & 34.97   & -0.79\% & 39.34   & -0.82\% & 44.15   & -0.63\% & 48.05   & -0.26\% & 52.68   &  0.14\% & 55.61   &  0.68\% \\
RF-POMO  & 25.45   &  0.58\% & 26.49   &  0.85\% & 28.23   &  0.44\% & 31.32   & -0.11\% & 35.19   & -0.18\% & 39.58   & -0.22\% & 44.41   & -0.06\% & 48.20   &  0.06\% & 52.61   &  0.02\% & 55.22   & -0.03\% \\
RF-MoE   & 25.43   &  0.51\% & 26.49   &  0.85\% & 28.21   &  0.35\% & 31.25   & -0.35\% & 35.10   & -0.43\% & 39.54   & -0.32\% & 44.35   & -0.19\% & 48.13   & -0.09\% & 52.53   & -0.14\% & 55.18   & -0.10\% \\
RF-TE    & 25.33   &  0.10\% & 26.40   &  0.50\% & 28.14   &  0.11\% & \textbf{31.17}   & \textbf{-0.61\%} & \textbf{34.91}   & \textbf{-0.95\%} & \textbf{39.30}   & \textbf{-0.93\%} & \textbf{44.08}   & \textbf{-0.80\%} & \textbf{47.86}   & \textbf{-0.65\%} & \textbf{52.40}   & \textbf{-0.38\%} & \textbf{55.16}   & \textbf{-0.14\%} \\

\bottomrule

\end{tabular}}
}
\end{center}
\end{table*}

For VRPL, we consider different distance limit values $l$. During training, we sample feasible instances with $l_\text{max} = 3.0$ as described in \cref{sec:mtvrp-data-generation}. For out-of-distribution settings, we test distances for values of $l = \{ 2.9, 3.0, 3.1, 3.2, 3.3, 3.4, 3.5 \} $. Interestingly, as shown in \cref{tab:ood-vrpl}, our model already outperforms POMO\_VRPL in distribution, and the trend is maintained for larger values of $l$.

\begin{table*}[h!]
\caption{\rebuttal{Comparison of our model with single-task POMO on out-of-distribution VRPL instances.}}
\label{tab:ood-vrpl}
\begin{center}
\large %
\renewcommand{\arraystretch}{1.05} %
\rebuttal{
\resizebox{\textwidth}{!}{ 
\begin{tabular}{l|cccccccccccccc}

\toprule

Distance Limit & \multicolumn{2}{c}{2.9} & \multicolumn{2}{c}{3.0} & \multicolumn{2}{c}{3.1} & \multicolumn{2}{c}{3.2} & \multicolumn{2}{c}{3.3} & \multicolumn{2}{c}{3.4} & \multicolumn{2}{c}{3.5} \\
\cmidrule(lr){2-3} \cmidrule(lr){4-5} \cmidrule(lr){6-7} \cmidrule(lr){8-9} \cmidrule(lr){10-11} \cmidrule(lr){12-13} \cmidrule(lr){14-15} 

 & Obj. & Gap & Obj. & Gap & Obj. & Gap & Obj. & Gap & Obj. & Gap & Obj. & Gap & Obj. & Gap \\

\midrule
      
POMO\_VRPL     & 15.84   &  * & 16.00   &  * & 16.04   &  * & 15.52   &  * & 16.02   &  * & 15.74   &  * & 15.85   &  * \\
MTPOMO   & 15.92   &  0.49\% & 16.08   &  0.53\% & 16.12   &  0.54\% & 15.59   &  0.47\% & 16.11   &  0.60\% & 15.81   &  0.48\% & 15.92   &  0.43\% \\
MVMoE    & 15.88   &  0.22\% & 16.03   &  0.22\% & 16.08   &  0.27\% & 15.54   &  0.11\% & 16.04   &  0.15\% & 15.78   &  0.25\% & 15.88   &  0.20\% \\
RF-POMO  & 15.91   &  0.41\% & 16.04   &  0.27\% & 16.09   &  0.33\% & 15.56   &  0.28\% & 16.06   &  0.29\% & 15.78   &  0.29\% & 15.87   &  0.13\% \\
RF-MoE   & 15.86   &  0.12\% & 16.03   &  0.21\% & 16.05   &  0.08\% & 15.53   &  0.09\% & 16.04   &  0.15\% & 15.77   &  0.21\% & 15.87   &  0.12\% \\
RF-TE    & \textbf{15.82}   & \textbf{-0.17\%} & \textbf{15.96}   & \textbf{-0.21\%} & \textbf{16.02}   & \textbf{-0.10\%} & \textbf{15.50}   & \textbf{-0.10\%} & \textbf{16.00}   & \textbf{-0.11\%} & \textbf{15.72}   & \textbf{-0.11\%} & \textbf{15.82}   & \textbf{-0.16\%} \\

\bottomrule

\end{tabular}}
}
\end{center}
\end{table*}

Finally, \cref{sec:cvrplib} reports the results for large-scale CVRPLIB, which demonstrate \our{} better generalize across sizes and real-world distributions than other multi-task models and single-variant ones. Overall, we can see that \our{} is robust, and its advantage is more pronounced the further away from the training distribution we go. This motivates future work in foundation models for VRPs, where we believe that exploring diverse solutions and variants will significantly advance the field.

\subsection{CVRPLIB Evaluation}
\label{sec:cvrplib}

\begin{table}[h]
  \caption{Results on large-scale CVRPLIB instances from the X set.  All models are only trained on the uniformly distributed data with the size $n=100$ and evaluated via greedy rollouts. Results for methods with $\dagger$ are drawn from \citet{zhou2024mvmoe}, models trained with single features excluding feature compositions (except for OVRPTW). Training on multiple variants enhances generalization across models.}
  \label{tab:vrplib-large}
  \begin{center}
  \begin{small}
  \renewcommand\arraystretch{1.5}
  \resizebox{\textwidth}{!}{ 
  \begin{tabular}{ll|cccccccc|cccccc}
    \toprule
    \midrule
    \multicolumn{2}{c|}{Set-X} & \multicolumn{2}{c}{POMO$^\dagger$} & \multicolumn{2}{c}{MTPOMO $^\dagger$} & \multicolumn{2}{c}{MVMoE$^\dagger$} & \multicolumn{2}{c|}{MVMoE-L$^\dagger$} & \multicolumn{2}{c}{MTPOMO} & \multicolumn{2}{c}{MVMoE} & \multicolumn{2}{c}{RF-TE}\\
     Instance & Opt. & Obj. & Gap & Obj. & Gap & Obj. & Gap & Obj. & Gap & Obj. & Gap & Obj. & Gap & Obj. & Gap \\
    \midrule
     X-n502-k39  & 69226  & 75617  & 9.232\%  & 77284  & 11.640\% & 73533  &  6.222\% & 74429  &  7.516\% & 69226  &  9.410\% & 76338  & 10.274\% & 71791  &  3.705\% \\
     X-n513-k21  & 24201  & 30518  & 26.102\% & 28510  & 17.805\% & 32102  & 32.647\% & 31231  & 29.048\% & 24201  & 42.511\% & 32639  & 34.866\% & 28465  & 17.619\%  \\
     X-n524-k153 & 154593 & 201877 & 30.586\% & 192249 & 24.358\% & 186540 & 20.665\% & 182392 & 17.982\% & 154593 & 14.771\% & 170999 & 10.612\% & 174381 & 12.800\%  \\
     X-n536-k96  & 94846  & 106073 & 11.837\% & 106514 & 12.302\% & 109581 & 15.536\% & 108543 & 14.441\% & 94846  & 16.109\% & 105847 & 11.599\% & 103272 &  8.884\%  \\
     X-n548-k50  & 86700  & 103093 & 18.908\% & 94562  &  9.068\% & 95894  & 10.604\% & 95917  & 10.631\% & 86700  & 27.851\% & 104289 & 20.287\% & 100956 & 16.443\%  \\
     X-n561-k42  & 42717  & 49370  & 15.575\% & 47846  & 12.007\% & 56008  & 31.114\% & 51810  & 21.287\% & 42717  & 30.770\% & 53383  & 24.969\% & 49454  & 15.771\%  \\
     X-n573-k30  & 50673  & 83545  & 64.871\% & 60913  & 20.208\% & 59473  & 17.366\% & 57042  & 12.569\% & 50673  & 20.210\% & 61524  & 21.414\% & 55952  & 10.418\%  \\
     X-n586-k159 & 190316 & 229887 & 20.792\% & 208893 &  9.761\% & 215668 & 13.321\% & 214577 & 12.748\% & 190316 & 19.125\% & 212151 & 11.473\% & 205575 &  8.018\%  \\
     X-n599-k92  & 108451 & 150572 & 38.839\% & 120333 & 10.956\% & 128949 & 18.901\% & 125279 & 15.517\% & 108451 & 21.098\% & 126578 & 16.714\% & 116560 &  7.477\%  \\
     X-n613-k62  & 59535  & 68451  & 14.976\% & 67984  & 14.192\% & 82586  & 38.718\% & 74945  & 25.884\% & 59535  & 30.523\% & 73456  & 23.383\% & 67267  & 12.987\%  \\
     X-n627-k43  & 62164  & 84434  & 35.825\% & 73060  & 17.528\% & 70987  & 14.193\% & 70905  & 14.061\% & 62164  & 23.193\% & 70414  & 13.271\% & 67572  &  8.700\%  \\
     X-n641-k35  & 63682  & 75573  & 18.672\% & 72643  & 14.071\% & 75329  & 18.289\% & 72655  & 14.090\% & 63682  & 30.321\% & 71975  & 13.023\% & 70831  & 11.226\%  \\
     X-n655-k131 & 106780 & 127211 & 19.134\% & 116988 &  9.560\% & 117678 & 10.206\% & 118475 & 10.952\% & 106780 & 12.731\% & 119057 & 11.497\% & 112202 &  5.078\%  \\
     X-n670-k130 & 146332 & 208079 & 42.197\% & 190118 & 29.922\% & 197695 & 35.100\% & 183447 & 25.364\% & 146332 & 24.809\% & 168226 & 14.962\% & 168999 & 15.490\%  \\
     X-n685-k75  & 68205  & 79482  & 16.534\% & 80892  & 18.601\% & 97388  & 42.787\% & 89441  & 31.136\% & 68205  & 36.550\% & 82269  & 20.620\% & 77847  & 14.137\%  \\
     X-n701-k44  & 81923  & 97843  & 19.433\% & 92075  & 12.392\% & 98469  & 20.197\% & 94924  & 15.870\% & 81923  & 13.319\% & 90189  & 10.090\% & 89932  &  9.776\% \\
     X-n716-k35  & 43373  & 51381  & 18.463\% & 52709  & 21.525\% & 56773  & 30.895\% & 52305  & 20.593\% & 43373  & 37.657\% & 52250  & 20.467\% & 49669  & 14.516\%  \\
     X-n733-k159 & 136187 & 159098 & 16.823\% & 161961 & 18.925\% & 178322 & 30.939\% & 167477 & 22.976\% & 136187 & 28.910\% & 156387 & 14.833\% & 148463 &  9.014\%  \\
     X-n749-k98  & 77269  & 87786  & 13.611\% & 90582  & 17.229\% & 100438 & 29.985\% & 94497  & 22.296\% & 77269  & 32.182\% & 92147  & 19.255\% & 85171  & 10.227\%  \\
     X-n766-k71  & 114417 & 135464 & 18.395\% & 144041 & 25.891\% & 152352 & 33.155\% & 136255 & 19.086\% & 114417 & 16.692\% & 130505 & 14.061\% & 129935 & 13.563\%  \\
     X-n783-k48  & 72386  & 90289  & 24.733\% & 83169  & 14.897\% & 100383 & 38.677\% & 92960  & 28.423\% & 72386  & 50.140\% & 96336  & 33.087\% & 83185  & 14.919\%  \\
     X-n801-k40  & 73305  & 124278 & 69.536\% & 85077  & 16.059\% & 91560  & 24.903\% & 87662  & 19.585\% & 73305  & 24.536\% & 87118  & 18.843\% & 86164  & 17.542\%  \\
     X-n819-k171 & 158121 & 193451 & 22.344\% & 177157 & 12.039\% & 183599 & 16.113\% & 185832 & 17.525\% & 158121 & 22.148\% & 179596 & 13.581\% & 174441 & 10.321\%  \\
     X-n837-k142 & 193737 & 237884 & 22.787\% & 214207 & 10.566\% & 229526 & 18.473\% & 221286 & 14.220\% & 193737 & 19.429\% & 230362 & 18.904\% & 208528 &  7.635\%  \\
     X-n856-k95  & 88965  & 152528 & 71.447\% & 101774 & 14.398\% & 99129  & 11.425\% & 106816 & 20.065\% & 88965  & 33.103\% & 105801 & 18.924\% & 98291  & 10.483\%  \\
     X-n876-k59  & 99299  & 119764 & 20.609\% & 116617 & 17.440\% & 119619 & 20.463\% & 114333 & 15.140\% & 99299  & 15.240\% & 114016 & 14.821\% & 107416 &  8.174\%  \\
     X-n895-k37  & 53860  & 70245  & 30.421\% & 65587  & 21.773\% & 79018  & 46.710\% & 64310  & 19.402\% & 53860  & 96.818\% & 69099  & 28.294\% & 64871  & 20.444\%  \\
     X-n916-k207 & 329179 & 399372 & 21.324\% & 361719 &  9.885\% & 383681 & 16.557\% & 374016 & 13.621\% & 329179 & 18.134\% & 373600 & 13.494\% & 352998 &  7.236\% \\
     X-n936-k151 & 132715 & 237625 & 79.049\% & 186262 & 40.347\% & 220926 & 66.466\% & 190407 & 43.471\% & 132715 & 50.654\% & 161343 & 21.571\% & 163162 & 22.942\%  \\
     X-n957-k87  & 85465  & 130850 & 53.104\% & 98198  & 14.898\% & 113882 & 33.250\% & 105629 & 23.593\% & 85465  & 48.127\% & 123633 & 44.659\% & 102689 & 20.153\%  \\
     X-n979-k58  & 118976 & 147687 & 24.132\% & 138092 & 16.067\% & 146347 & 23.005\% & 139682 & 17.404\% & 118976 & 16.711\% & 131754 & 10.740\% & 129952 &  9.225\% \\
     X-n1001-k43 & 72355  & 100399 & 38.759\% & 87660  & 21.153\% & 114448 & 58.176\% & 94734  & 30.929\% & 72355  & 82.677\% & 88969  & 22.962\% & 85929  & 18.760\%  \\
    \midrule
     \multicolumn{2}{c|}{Avg. Gap} & \multicolumn{2}{c}{29.658\%} & \multicolumn{2}{c}{16.796\%} & \multicolumn{2}{c}{26.408\%} & \multicolumn{2}{c}{19.607\%} & \multicolumn{2}{c}{30.202\%} & \multicolumn{2}{c}{18.795\%} & \multicolumn{2}{c}{\textbf{12.303\%}}\\
    \midrule
    \bottomrule
  \end{tabular}}
  \end{small}
  \end{center}
\end{table}

We report in \cref{tab:vrplib-large} the results for large-scale CVRPLIB \citep{lima2014cvrplib} with sizes greater than $500$ as done in MVMoE \citep{zhou2024mvmoe}. We report the original POMO \citep{kwon2020pomo} alongside versions of MTPOMO and MVMoE that were initially trained on mixtures of only CVRP, OVRP, VRPL, VRPB, VRPTW, and OVRPTW for more than $3\times$ longer than our setting with all variants. Interestingly, training on all variants improves the generalization performance of MVMoE compared to the original setting, while it decreases the MTPOMO one (possibly due to the fact several more CVRP instances were sampled in MVMoE's setting). Notably, \our{} vastly outperforms other SOTA single and multi-task RL baselines.

\subsection{Effect of Transformer Encoder Components}
\label{append:effect-transformer-components}

We study the effect of the proposed Transformer Encoder by ablating its components, in particular:
%
%
\begin{enumerate}
    \item \our{}: uses the full proposed Transformer Encoder as described in \cref{subsec:transformer-architecture}.
    \item \our{} (No RMSNorm): removes the RMSNorm in pre-norm, but keeps the SwiGLU MLP.
    \item \our{} (No SwiGLU): removes the SwiGLU MLP, but leaves the RMSNorm
    \item \our{} (No SwiGLU, No RMSNorm): removes all components and is equivalent to the commonly used Attention Model-style encoder \citep{kool2018attention}.
\end{enumerate}

\begin{figure}[H]
    \centering
    \includegraphics[width=0.5\linewidth]{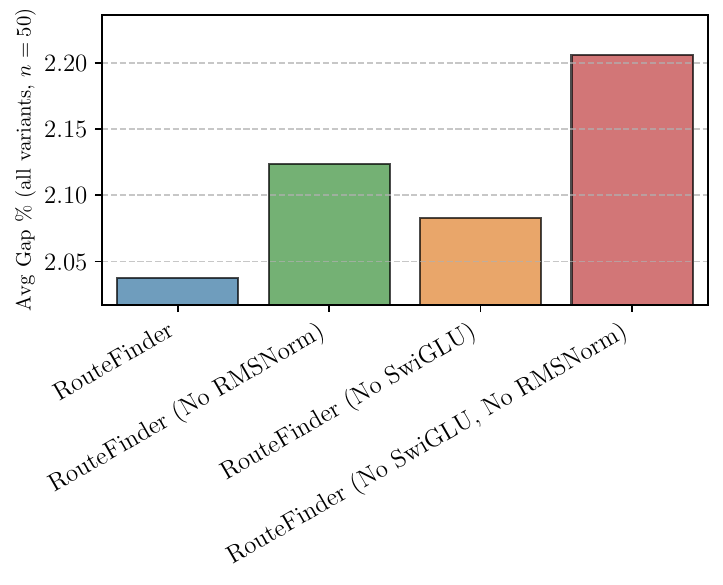}
    \caption{Effect of encoder components.}
    \label{fig:encoder-ablation-test}
\end{figure}

\begin{figure}[H]
    \centering
    \includegraphics[width=.8\linewidth]{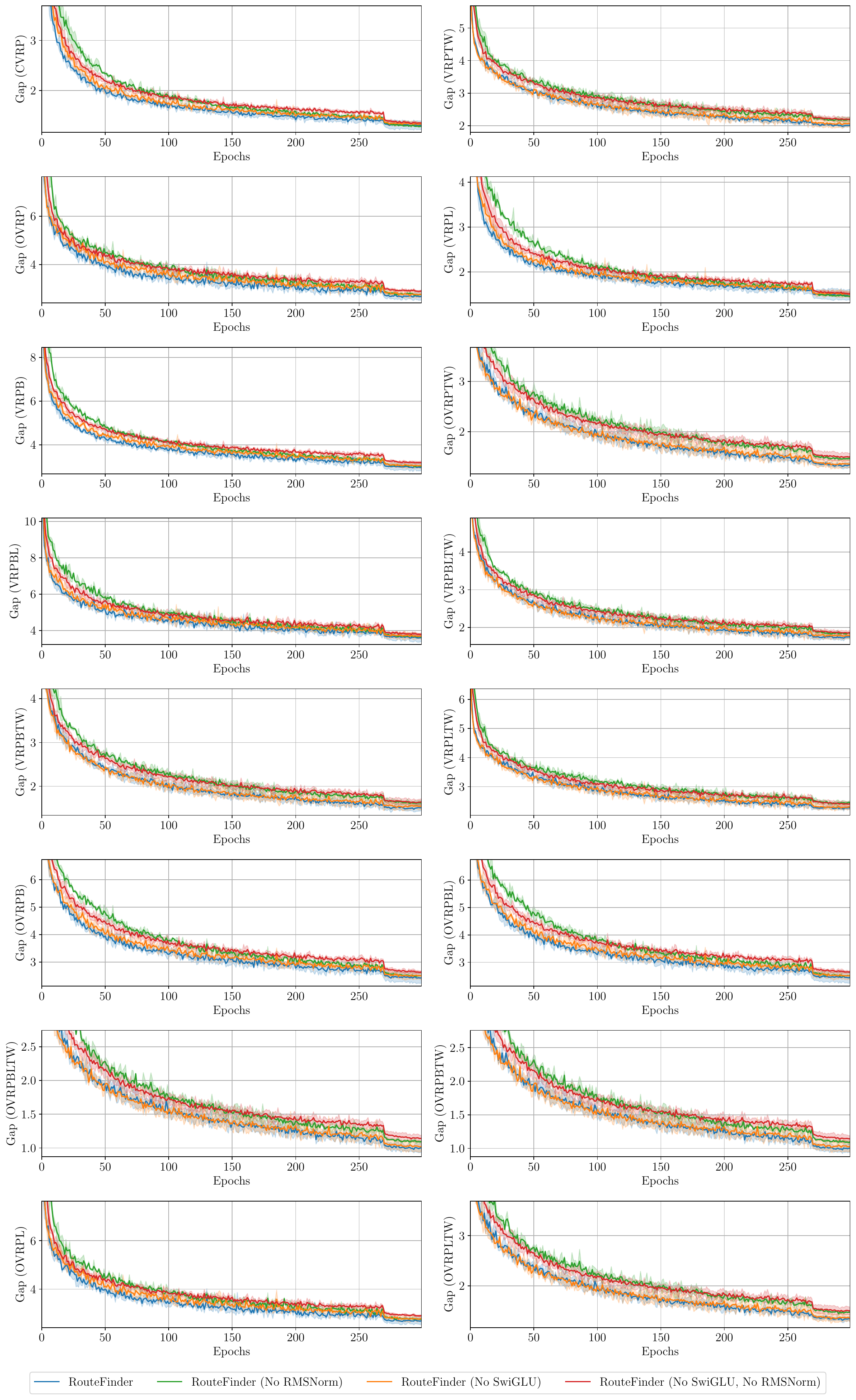}
    \label{fig:encoder-ablation-epochs}
    \caption{Ablation study on proposed encoder components over training.}
\end{figure}

We show in \cref{fig:encoder-ablation-test} the effect of each component on the test gaps for $n=50$ nodes, averaged across the 16 variants of \cref{tab:vrplib-large}. The full \our{} provides the best performance. We additionally study the behavior of each single component on validation data during the training epochs across different variants in \cref{fig:encoder-ablation-epochs}. Interestingly, as shown in \cref{fig:encoder-ablation-epochs}, while the final performance for the variant with no RMSNorm outperforms the baseline due to its enhanced capability in representation learning, its convergence is slower in the beginning. However, the full Transformer Encoder containing both RMSNorm and SwiGLU not only performs the best, but also converges the fastest, indicating the importance of each single component.

\paragraph{FlashAttention speedup}
FlashAttention \citep{dao2022flashattention,dao2023flashattention} is a recent exact attention algorithm that can be used to significantly speed up computations with mixed precision. This can be applied to any model with an attention-based mechanism, so we apply it by default to all neural networks compared in this work. Overall, we can improve training and inference speed by up to $20\%+$ with no performance degradation.

\subsection{Effect of Reward Normalization Techniques}
\label{append:effect-normalization-scheme}

In this section, we study the effect of different reward normalization techniques, i.e., 1)---4) from \cref{sec:reward-normalization}.
\begin{figure}[H]
    \centering    \includegraphics[width=0.6\linewidth]{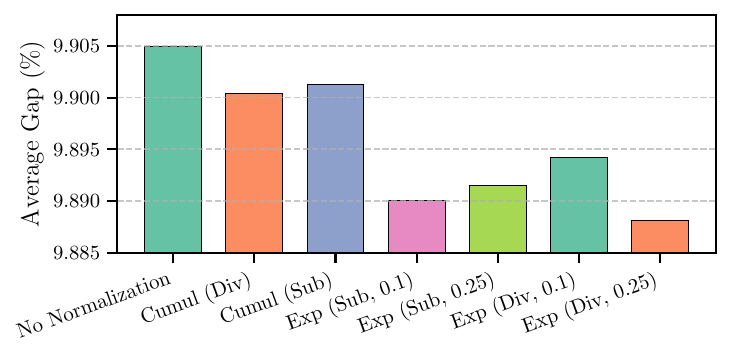}
    \vspace{-3mm}
    \caption{Effect of reward normalization.}
    \label{fig:ablation-study-reward-normalization}
    \vspace{-3mm}
\end{figure}
\cref{fig:ablation-study-reward-normalization} shows that reward normalization successfully improves performance across variants. We also show different values of $\alpha$ for the exponential moving averages and find that the division through the exponentially smoothed mean with $\alpha=0.25$ works best. Future normalization research may further improve performance.

\subsection{Studies on Mixed Batch Training}
\label{subsec:append-mixed-batch-training-study}

\subsubsection{Effect on convergence speed}
\label{subsec:append-mixed-batch-training-study-stability}

\begin{figure}[H]
    \centering
    \includegraphics[width=.85\linewidth]{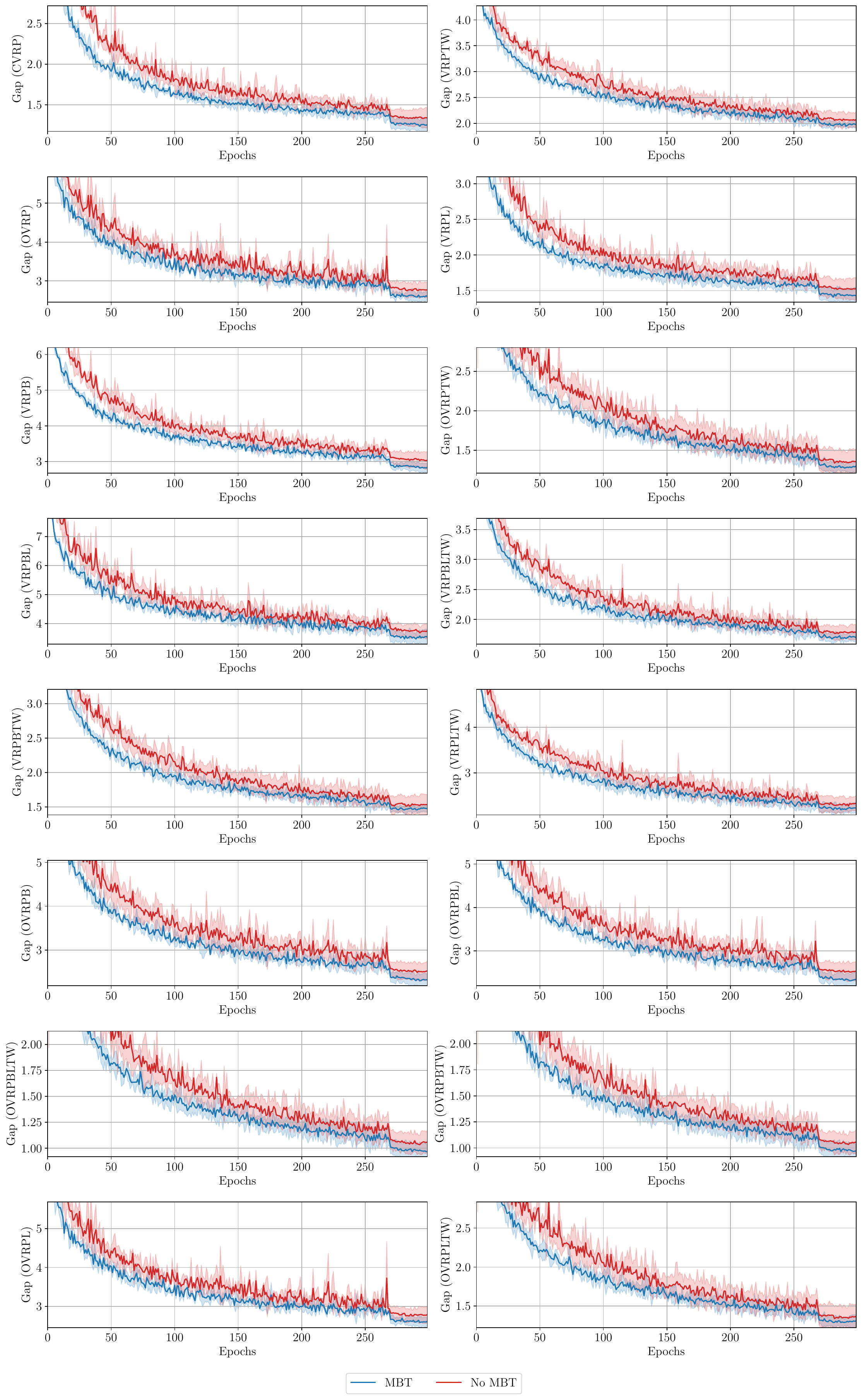}
    \caption{Mixed Batch Training (MBT) allows for better convergence across all variants.}
    \label{fig:mbt-epochs}
\end{figure}

In addition to the effect of MBT on the training loss shown in \cref{subsec:ablation-studies-main}, we also show the validation gaps on held-out instances in \cref{fig:mbt-epochs}, where MBT speeds up convergence across all variants.

\subsubsection{Effect on imbalanced variant distributions} 
\label{subsec:append-mixed-batch-training-study-stability-imbalanced-distributions}

\begin{figure}[H]
    \centering
    \includegraphics[width=.8\linewidth]{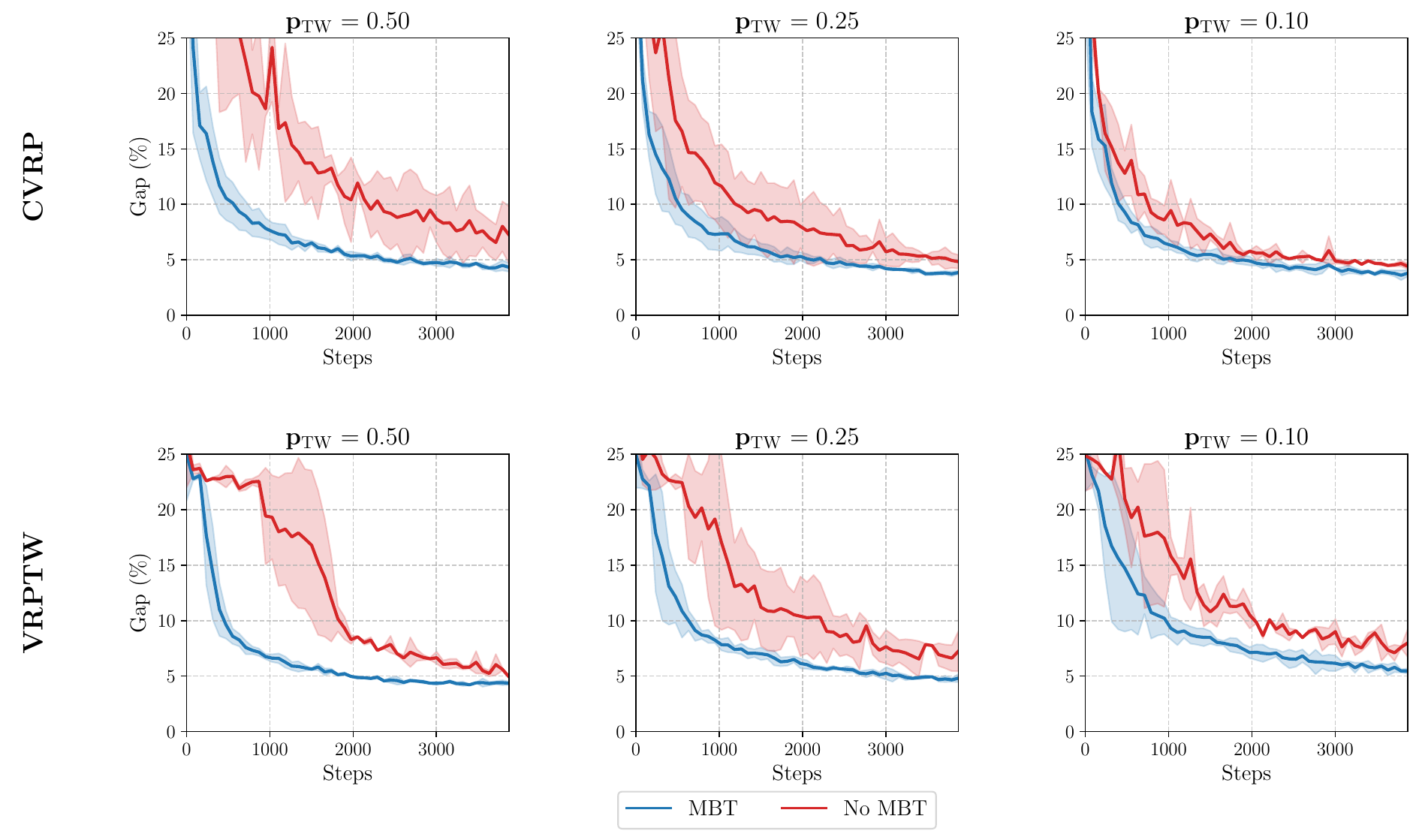}
    \caption{\rebuttal{Effect of Mixed Batch Training (MBT) on imbalanced variant distributions with varying probability $\mathbf{p}_{\text{TW}}$ of sampling time windows (TW). MBT stabilizes the training not only for the downsampled TW variants such as VRPTW but also improves the performance for variants with more samples as CVRP.}}
    \label{fig:mbt-imbalanced}
\end{figure}

As explained in \cref{subsec:mixed-batch-training}, we can sample variants uniformly by setting the probability of sampling base attributes $\nu$ as $\mathbf{p}_{\nu} = 0.5$. We study the behavior of MBT in imbalanced attribute distributions. We train \our{} models from scratch with the same setting as the main experiments for $50$ epochs with $10,000$ instances of size $50$ sampled per epoch, with and without MBT, and at different values of the sampling probability for time window attributes $\mathbf{p}_\text{TW}$ as $0.5$, $0.25$, and $0.10$. \cref{fig:mbt-imbalanced} shows the validation gaps over the training.
Decreasing $\mathbf{p}_\text{TW}$ (towards the right of the plot) results in fewer time window attributes; thus, the convergence is slower for variants such as VRPTW. On the other hand, variants like the CVRP will be sampled with higher probability, which results in slightly faster convergence. MBT plays an important role in stabilizing the training for all cases. Interestingly, while its effect is more moderate for the majority samples (CVRP), this effect is higher on minority samples as VRPTW, where it results in a stable training curve, yielding fast convergence.

\subsection{T-SNE Visualizations}
\label{append:t-sne-visualization-interpretability}

For interpretability, we study the representations learned from the model across different variants. Given their high dimensionality, we employ t-SNE \citep{van2008visualizing} to project them in 2D space. We employ the implementation from \texttt{scikit-learn} with the default perplexity of $30$ and use $100$ instances of size $100$ for each of the 16 variants of the main experiments from \cref{subsec:experiments-main-results}.

\subsubsection{Layer-wise visualization}
\label{append:t-sne-visualization-interpretability-layer-wise-viz}

\begin{figure}[H]
    \centering
    \includegraphics[width=\linewidth]{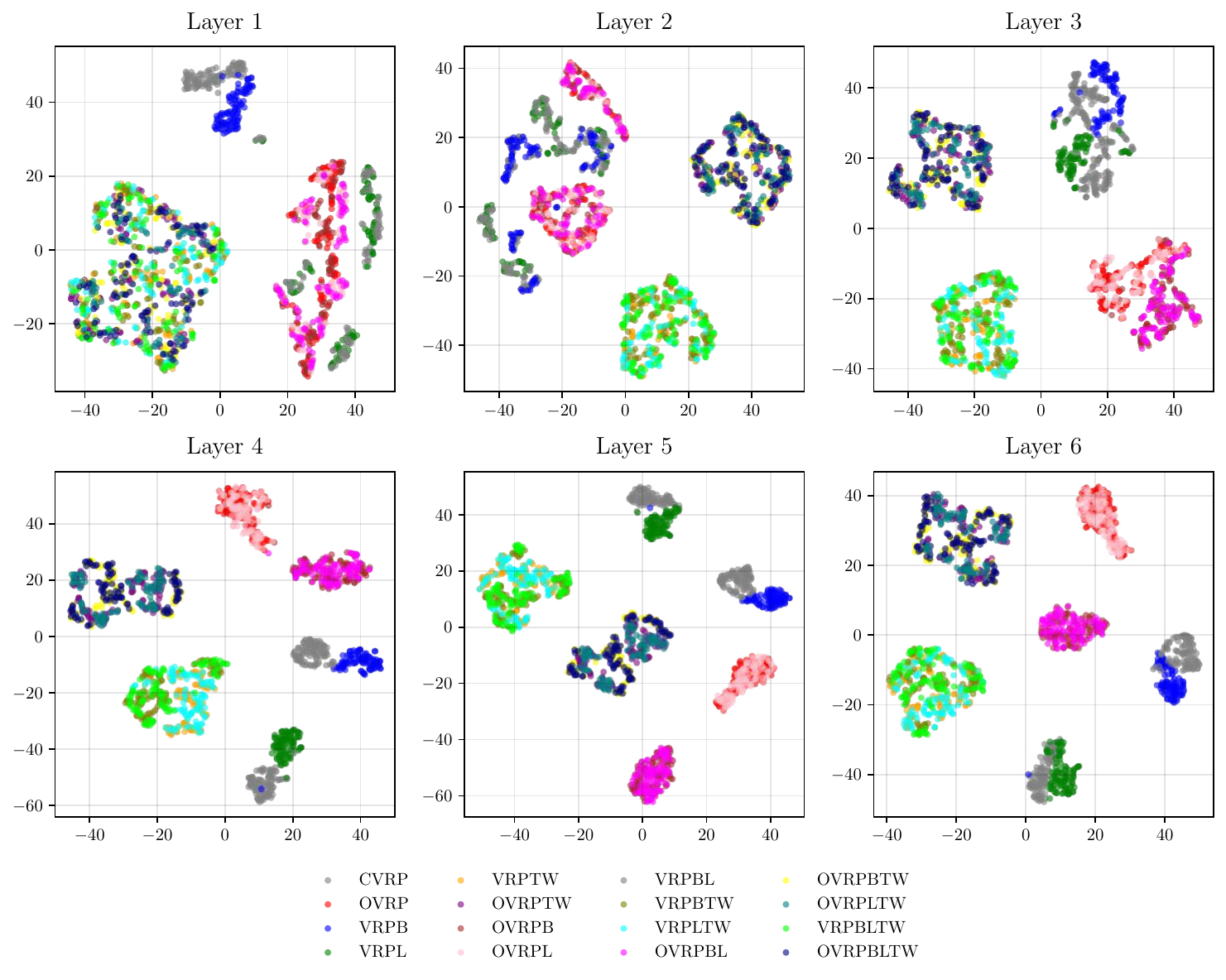}
    \caption{Visualization of \our{}'s Transformer Encoder latent space via t-SNE analysis by layer. Problem patterns become more visible with deeper layers, generating distinct clusters.}
    \label{fig:tsne-routefinder}
\end{figure}

  We study \our{}'s Transformer Encoder layers.  As shown in \cref{fig:tsne-routefinder}, distinct clusters emerge at different model layers, indicating that the model progressively separates the problem variants with increasing depth. Early layers (Layer 1) exhibit high overlap between different variants, suggesting shared feature extraction. However, as we proceed to deeper layers (Layer 6), the clusters become more distinct, particularly for more complex variants such as OVRPB, VRPBLTW, and VRPBTW, signifying the model's capacity to capture and differentiate intricate problem structures.

\subsubsection{Comparison across models and VRP variants}
\label{append:t-sne-visualization-interpretability-comparison-models}

\begin{figure}[h!]
    \centering
    
    \begin{minipage}{\textwidth}
        \centering
        \makebox[0.9cm]{}%
        \makebox[0.92\textwidth][c]{%
            \makebox[0.23\textwidth]{\footnotesize Open Routes (O)}%
            \hfill
            \makebox[0.23\textwidth]{\footnotesize Distance Limits (L)}%
            \hfill
            \makebox[0.23\textwidth]{\footnotesize Backhauls (B)}%
            \hfill
            \makebox[0.23\textwidth]{\footnotesize Time Windows (TW)}%
        }
    \end{minipage}

    \vspace{0.2cm}
    \begin{minipage}{\textwidth}
        \centering
        \makebox[0.9cm][c]{\rotatebox{90}{\footnotesize $~~~~~~~~~~$ MTPOMO}}%
        \makebox[0.92\textwidth][c]{%
            \includegraphics[width=0.92\textwidth]{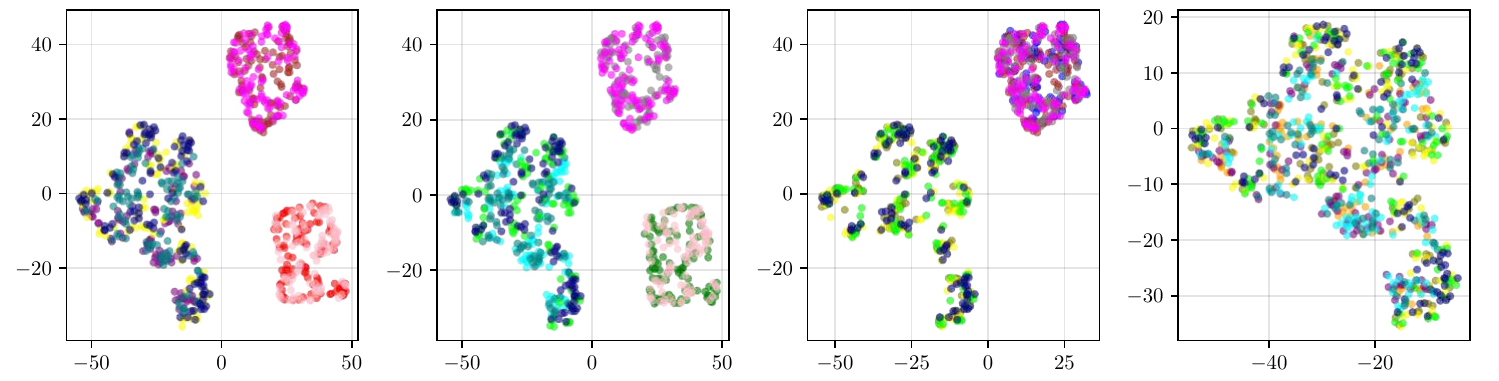}%
        }
    \end{minipage}

    \vspace{0.2cm}
    \begin{minipage}{\textwidth}
        \centering
        \makebox[0.9cm][c]{\rotatebox{90}{\footnotesize $~~~~~~~~~~~~~$ MVMoE}}%
        \makebox[0.92\textwidth][c]{%
            \includegraphics[width=0.92\textwidth]{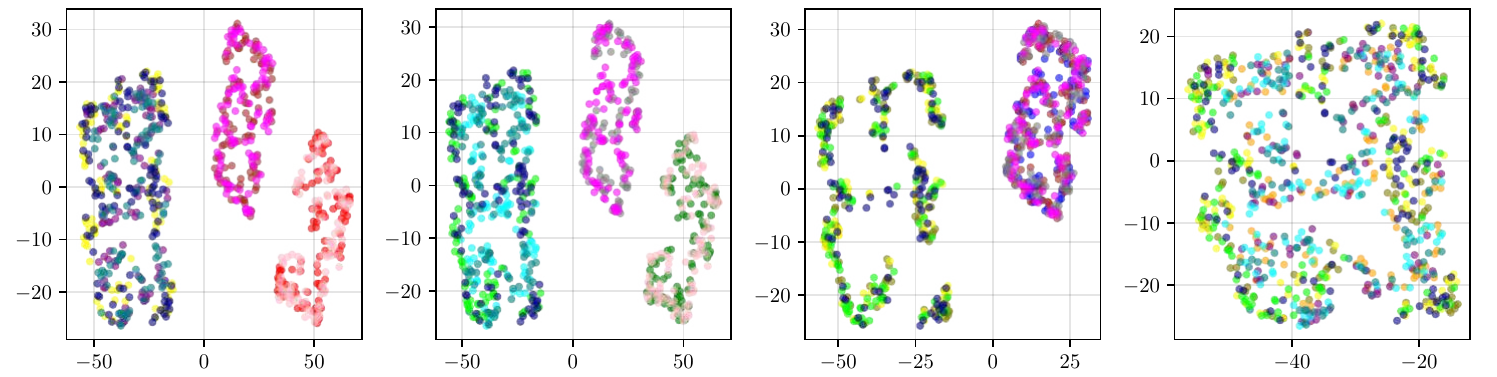}%
        }
    \end{minipage}

    \vspace{0.2cm}
    \begin{minipage}{\textwidth}
        \centering
        \makebox[0.9cm][c]{\rotatebox{90}{\footnotesize $~~~~~~~~$ \our{}}}%
        \makebox[0.92\textwidth][c]{%
            \includegraphics[width=0.92\textwidth]{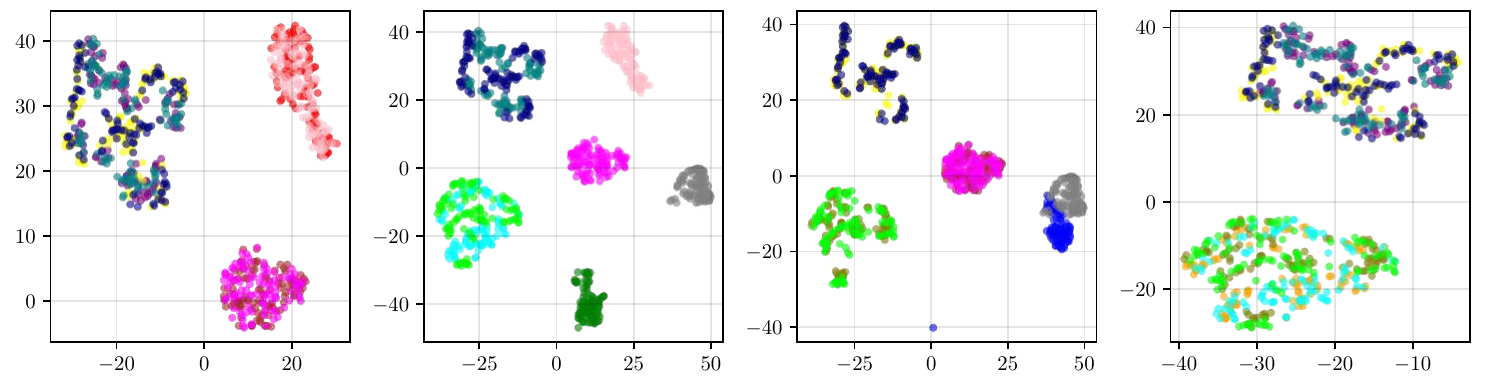}%
        }
    \end{minipage}

    \vspace{0.2cm}
    \begin{minipage}{\textwidth}
        \centering
        \makebox[0.9cm]{}%
        \makebox[0.92\textwidth][c]{%
            \includegraphics[width=0.92\textwidth]{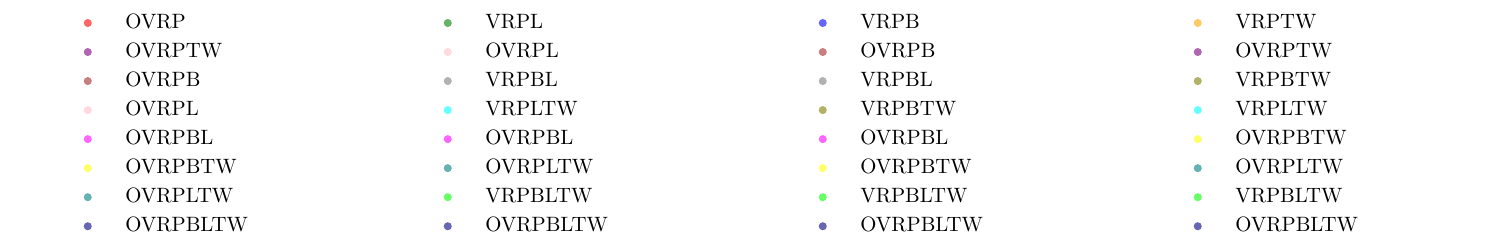}%
        }
    \end{minipage}
    \vspace{-3mm}
    \caption{\rebuttal{Analysis of the t-SNE latent space for the last encoder layer for different attributes. \our{} yields well-defined, tightly grouped, and distinct clusters on all variants -- a strong indicator of its capability to generalize and specialize effectively in solving diverse VRP variants. For example, unlike the baselines, \our{} distinctly separates time window variants into two clusters with and without open routes (bottom-right image) thanks to the Global Attribute Embeddings.}}
    \label{fig:tsne_by_variant}
\end{figure}

We also compare t-SNE analyses across the models, in particular, MTPOMO and MVMoE, compared to our \our{} with Transformer Encoder layers, with embeddings taken in the last encoder layer for all models. In particular, we aim to analyze the differences in latent representation problem variants across the four attributes: open routes (O), distance limits (L), backhauls (B), and time windows (TW). \cref{fig:tsne_by_variant} shows that \our{} generates more and defined clusters, indicating a better-learned representation \citep{arora2018analysis}. For open routes, \our{} has more defined clusters than the baselines. In distance limits, our model generates double the clusters, which indicates different relations between attributes; for instance, the model clearly separates backhaul variants VRPB and VRPBL (green and grey, respectively), while other models do not clearly do this. This also holds in the backhaul attribute clusters, where \our{} more clearly separates different types of time windows as well as distance limits. Finally, for time windows clusters, we notice the most striking difference -- while MTPOMO and MVMoE fail to distinguish between time window variants, resulting in a single and sparse cluster, \our{} separates time window variants with and without the open (O) attribute into two separate clusters thanks to the Global Attribute Embeddings.

\subsection{Finetuning with EAL}
\label{subsec:finetuning-appendix}

\subsubsection{Finetuning to Unseen Variants}
\label{append:finetune-unseen-variants}

We conduct additional experiments on zero-shot generalization of various models and finetuning across three different settings of unseen variants in order of difficulty:

\begin{enumerate}
    \item Mixed backhauls (MB): this is the setting from \cref{sec:main-paper-eal}. We report the results in full in \cref{tab:eal-finetuning-mb-only-16} and trends over epochs in \cref{fig:finetuning-mb}.
    \item Multi-depot (MD): we add more attributes for finetuning approaches as per \cref{append:eal-modeling-details} with data generated as in \cref{sec:mtvrp-data-generation}. Results are reported in \cref{tab:eal-finetuning-md-only-32} and trends over epochs in \cref{fig:finetuning-md}.
    \item Mixed backhauls \& multi-depot (MB\&MD): this is the hardest setting, which considers as finetuning variants only the ones containing both the unseen MB and MD attributes at the same time from \cref{table:vrp_variants_asymmetric}. Full results are in \cref{tab:eal-finetuning-mb-md-16}  with trends over epochs in \cref{fig:finetuning-both}.
\end{enumerate}
We keep the same methodology as outlined in \cref{sec:main-paper-eal}, i.e., 10 epochs with 10k instances sampled for each epoch. We use \our{} models with Transformer Encoder (\textsc{RF}-TE), untrained for the scratch training and pretrained from the same checkpoints as the main experiments in \cref{subsec:experiments-main-results} for AL and EAL finetuning. Additional details on EAL modeling are available in \cref{append:eal-modeling-details}.

\begin{table}[h!]
\rebuttal{
\centering
\caption{\rebuttal{Zero-shot, retraining, and fine-tuning performance on unseen mixed backhaul (MB) variants. "$\varnothing$" are models and fine-tuning methods evaluated in zero-shot settings. EAL maintains the zero-shot performance and performs best overall.}}
\label{tab:eal-finetuning-mb-only-16}
\resizebox{\textwidth}{!}{%
\large %
\renewcommand{\arraystretch}{1.05} %
\begin{tabular}{l | cccccccccccccccc}
\toprule
 & \multicolumn{2}{c}{VRPMB} & \multicolumn{2}{c}{OVRPMB} & \multicolumn{2}{c}{VRPMBL} & \multicolumn{2}{c}{VRPMBTW} & \multicolumn{2}{c}{OVRPMBL} & \multicolumn{2}{c}{OVRPMBTW} & \multicolumn{2}{c}{VRPMBLTW} & \multicolumn{2}{c}{OVRPMBLTW} \\
\cmidrule(lr){2-3} \cmidrule(lr){4-5} \cmidrule(lr){6-7} \cmidrule(lr){8-9} \cmidrule(lr){10-11} \cmidrule(lr){12-13} \cmidrule(lr){14-15} \cmidrule(lr){16-17} 
Method & Obj. & Gap & Obj. & Gap & Obj. & Gap & Obj. & Gap & Obj. & Gap & Obj. & Gap & Obj. & Gap & Obj. & Gap \\
\midrule
HGS-PyVRP & 13.54 & * & 9.01 & * & 13.78 & * & 25.51 & * & 9.01 & * & 16.97 & * & 25.85 & * & 16.97 & * \\
OR-Tools & 14.93 & $10.27\%$ & 10.59 & $17.54\%$ & 15.42 & $11.90\%$ & 29.97 & $17.48\%$ & 10.59 & $17.54\%$ & 19.31 & $13.78\%$ & 30.44 & $17.76\%$ & 19.31 & $13.78\%$ \\
MTPOMO$^\varnothing$  & 15.04 & $11.32\%$ & 10.87 & $20.65\%$ & 15.41 & $11.97\%$ & 28.31 & $11.06\%$ & 10.85 & $20.43\%$ & 18.51 & $9.08\%$ & 28.73 & $11.27\%$ & 18.51 & $9.12\%$ \\
MVMoE$^\varnothing$  & 14.99 & $10.94\%$ & 10.85 & $20.42\%$ & 15.33 & $11.37\%$ & 28.32 & $11.10\%$ & 10.82 & $20.14\%$ & 18.55 & $9.33\%$ & 28.70 & $11.16\%$ & 18.55 & $9.30\%$ \\
\textsc{RF}-POMO$^\varnothing$  & 14.98 & $10.90\%$ & 10.84 & $20.31\%$ & 15.29 & $11.12\%$ & 28.53 & $11.94\%$ & 10.84 & $20.32\%$ & 18.62 & $9.72\%$ & 28.89 & $11.89\%$ & 18.62 & $9.71\%$ \\
\textsc{RF}-MoE$^\varnothing$  & 14.93 & $10.49\%$ & 10.76 & $19.49\%$ & 15.21 & $10.47\%$ & 28.20 & $10.63\%$ & 10.76 & $19.40\%$ & 18.45 & $8.74\%$ & 28.55 & $10.57\%$ & 18.45 & $8.72\%$ \\
\textsc{RF}-TE$^\varnothing$  & 14.88 & $10.13\%$ & 10.72 & $19.02\%$ & 15.18 & $10.32\%$ & 28.29 & $10.87\%$ & 10.72 & $19.01\%$ & 18.45 & $8.68\%$ & 28.65 & $10.82\%$ & 18.45 & $8.69\%$ \\
Train (scratch) & 15.18 & $12.13\%$ & 10.40 & $15.38\%$ & 15.48 & $12.37\%$ & 28.11 & $10.17\%$ & 10.46 & $16.08\%$ & 18.85 & $11.09\%$ & 28.69 & $10.95\%$ & 18.86 & $11.19\%$ \\
AL$^\varnothing$ & 43.15 & $221.25\%$ & 37.98 & $323.23\%$ & 32.81 & $139.84\%$ & 59.17 & $133.55\%$ & 29.15 & $224.37\%$ & 39.03 & $131.09\%$ & 66.62 & $158.21\%$ & 40.92 & $141.51\%$ \\
AL  & 14.91 & $10.10\%$ & 10.14 & $12.53\%$ & 15.12 & $9.73\%$ & 27.79 & $8.92\%$ & 10.18 & $12.95\%$ & 18.52 & $9.13\%$ & 28.33 & $9.56\%$ & 18.51 & $9.05\%$ \\
EAL$^\varnothing$  & 14.88 & $10.13\%$ & 10.72 & $19.02\%$ & 15.18 & $10.32\%$ & 28.29 & $10.87\%$ & 10.72 & $19.01\%$ & 18.45 & $8.68\%$ & 28.65 & $10.82\%$ & 18.45 & $8.69\%$ \\
EAL  & \textbf{14.59} & \textbf{7.89}$\%$ &\textbf{ 9.66} & \textbf{7.19}$\%$ & \textbf{14.78} & \textbf{7.39}$\%$ & \textbf{26.69} & \textbf{4.61}$\%$ & \textbf{9.65} & \textbf{7.13}$\%$ & \textbf{17.60} & \textbf{3.70}$\%$ & \textbf{27.13} & \textbf{4.90}$\%$ & \textbf{17.59} & \textbf{3.65}$\%$ \\
\bottomrule
\end{tabular}
}
}
\end{table}

\begin{table}[h!]
\rebuttal{
\centering
\caption{\rebuttal{Zero-shot, retraining, and fine-tuning performance on unseen multi-depot (MD) variants. "$\varnothing$" denotes models and fine-tuning methods evaluated in zero-shot settings. EAL maintains the zero-shot performance and performs best overall.}}
\label{tab:eal-finetuning-md-only-32}
\resizebox{\textwidth}{!}{%
\large %
\renewcommand{\arraystretch}{1.05} %
\begin{tabular}{l | cccccccccccccccc}
\toprule
 & \multicolumn{2}{c}{MDCVRP} & \multicolumn{2}{c}{MDOVRP} & \multicolumn{2}{c}{MDVRPB} & \multicolumn{2}{c}{MDVRPL} & \multicolumn{2}{c}{MDVRPTW} & \multicolumn{2}{c}{MDOVRPTW} & \multicolumn{2}{c}{MDOVRPB} & \multicolumn{2}{c}{MDOVRPL} \\
\cmidrule(lr){2-3} \cmidrule(lr){4-5} \cmidrule(lr){6-7} \cmidrule(lr){8-9} \cmidrule(lr){10-11} \cmidrule(lr){12-13} \cmidrule(lr){14-15} \cmidrule(lr){16-17} 
Method & Obj. & Gap & Obj. & Gap & Obj. & Gap & Obj. & Gap & Obj. & Gap & Obj. & Gap & Obj. & Gap & Obj. & Gap \\
\midrule
HGS-PyVRP & 11.89 &       * & 7.97 & *        & 11.64 & *        & 11.90 & *        & 19.33 & *         & 13.00 & *         & 8.69 & *          & 7.97 & *           \\
OR-Tools & 12.52 & $5.27\%$ & 8.16 & $2.33\%$ & 12.22 & $5.01\%$ & 12.52 & $5.24\%$ & 19.62 & $1.55\%$  & 13.09 & $0.74\%$  & 8.87 & $2.15\%$   & 8.16 & $2.33\%$    \\
MTPOMO$^\varnothing$  & 16.07 & $35.74\%$ & 10.28 & $29.06\%$ & 15.18 & $30.66\%$ & 16.30 & $37.58\%$ & 26.68 & $38.56\%$ & 17.57 & $35.67\%$ & 10.94 & $26.08\%$ & 10.28 & $29.07\%$ \\
MVMoE$^\varnothing$  & 16.02 & $35.35\%$ & 10.24 & $28.59\%$ & 15.12 & $30.13\%$ & 16.25 & $37.17\%$ & 26.67 & $38.51\%$ & 17.57 & $35.68\%$ & 10.89 & $25.56\%$ & 10.24 & $28.60\%$ \\
\textsc{RF}-POMO$^\varnothing$  & 16.03 & $35.46\%$ & 10.23 & $28.52\%$ & 15.11 & $30.10\%$ & 16.25 & $37.19\%$ & 26.60 & $38.16\%$ & 17.54 & $35.43\%$ & 10.88 & $25.48\%$ & 10.23 & $28.55\%$ \\
\textsc{RF}-MoE$^\varnothing$  & 16.01 & $35.24\%$ & 10.20 & $28.06\%$ & 15.06 & $29.69\%$ & 16.21 & $36.89\%$ & 26.60 & $38.11\%$ & 17.54 & $35.44\%$ & 10.84 & $25.02\%$ & 10.20 & $28.06\%$ \\
\textsc{RF}-TE$^\varnothing$  & 15.98 & $35.02\%$ & 10.18 & $27.82\%$ & 15.05 & $29.53\%$ & 16.20 & $36.76\%$ & 26.51 & $37.64\%$ & 17.48 & $34.96\%$ & 10.82 & $24.74\%$ & 10.18 & $27.84\%$ \\
Train (scratch) &  14.44 & 21.59$\%$ &  9.88 & 23.87$\%$ &  14.86 & 27.75$\%$ &  14.50 & 21.99$\%$ &  23.33 & 20.82$\%$ &  15.48 & 19.16$\%$ &  10.76 & 23.84$\%$ &  9.89 & 24.07$\%$ \\
AL$^\varnothing$ & 33.91 & $188.76\%$ & 25.02 & $215.12\%$ & 33.56 & $189.58\%$ & 31.06 & $164.78\%$ & 49.08 & $155.57\%$ & 31.17 & $141.42\%$ & 26.30 & $203.65\%$ & 24.12 & $203.73\%$  \\
AL  &  14.23 & 19.84$\%$ &  9.67 & 21.28$\%$ &  14.84 & 27.57$\%$ &  14.33 & 20.51$\%$ &  22.64 & 17.18$\%$ &  15.05 & 15.81$\%$ &  10.69 & 23.12$\%$ &  9.69 & 21.45$\%$  \\
EAL$^\varnothing$  & 15.98 & $35.02\%$ & 10.18 & $27.82\%$ & 15.05 & $29.53\%$ & 16.20 & $36.76\%$ & 26.51 & $37.64\%$ & 17.48 & $34.96\%$ & 10.82 & $24.74\%$ & 10.18 & $27.84\%$  \\
EAL  & \textbf{12.96} & \textbf{9.14}$\%$ & \textbf{8.64} & \textbf{8.37}$\%$ & \textbf{13.05} & \textbf{12.15}$\%$ & \textbf{12.99} & \textbf{9.31}$\%$ & \textbf{21.14} & \textbf{9.43}$\%$ & \textbf{13.81} & \textbf{6.24}$\%$ & \textbf{9.46} & \textbf{8.88}$\%$ & \textbf{8.64} & \textbf{8.33}$\%$ \\
\end{tabular}
}
\resizebox{\textwidth}{!}{%
\large %
\renewcommand{\arraystretch}{1.05} %
\begin{tabular}{l | cccccccccccccccc}
\toprule
 & \multicolumn{2}{c}{MDVRPBL} & \multicolumn{2}{c}{MDVRPBTW} & \multicolumn{2}{c}{MDVRPLTW} & \multicolumn{2}{c}{MDOVRPBL} & \multicolumn{2}{c}{MDOVRPBTW} & \multicolumn{2}{c}{MDOVRPLTW} & \multicolumn{2}{c}{MDVRPBLTW} & \multicolumn{2}{c}{MDOVRPBLTW} \\
\cmidrule(lr){2-3} \cmidrule(lr){4-5} \cmidrule(lr){6-7} \cmidrule(lr){8-9} \cmidrule(lr){10-11} \cmidrule(lr){12-13} \cmidrule(lr){14-15} \cmidrule(lr){16-17} 
Method & Obj. & Gap & Obj. & Gap & Obj. & Gap & Obj. & Gap & Obj. & Gap & Obj. & Gap & Obj. & Gap & Obj. & Gap \\
\midrule
HGS-PyVRP & 11.68 & *       & 22.03 & *        & 19.35 & *         & 8.69 & *          & 14.369 & *        & 13.00 & *         & 22.06 & *         & 14.37 & *           \\
OR-Tools & 12.22 & $4.66\%$  & 22.40 & $1.69\%$ & 19.66 & $1.58\%$ & 8.87 & $2.13\%$   & 14.49 & $0.87\%$  & 13.09 & $0.70\%$  & 22.43 & $1.70\%$  & 14.49 & $0.86\%$    \\
MTPOMO$^\varnothing$  & 15.80 & $35.54\%$ & 30.55 & $39.23\%$ & 27.13 & $40.71\%$ & 10.94 & $26.11\%$ & 19.69 & $37.62\%$ & 17.58 & $35.70\%$ & 31.09 & $41.52\%$ & 19.69 & $37.64\%$ \\
MVMoE$^\varnothing$  & 15.73 & $34.95\%$ & 30.55 & $39.22\%$ & 27.12 & $40.67\%$ & 10.90 & $25.66\%$ & 19.69 & $37.62\%$ & 17.58 & $35.74\%$ & 31.06 & $41.39\%$ & 19.69 & $37.61\%$ \\
\textsc{RF}-POMO$^\varnothing$  & 15.71 & $34.80\%$ & 30.46 & $38.80\%$ & 27.04 & $40.22\%$ & 10.89 & $25.49\%$ & 19.66 & $37.38\%$ & 17.54 & $35.43\%$ & 30.97 & $41.00\%$ & 19.66 & $37.37\%$ \\
\textsc{RF}-MoE$^\varnothing$  & 15.65 & $34.25\%$ & 30.47 & $38.87\%$ & 27.03 & $40.18\%$ & 10.84 & $25.03\%$ & 19.66 & $37.40\%$ & 17.55 & $35.45\%$ & 30.98 & $41.06\%$ & 19.66 & $37.42\%$ \\
\textsc{RF}-TE$^\varnothing$  & 15.62 & $33.98\%$ & 30.36 & $38.36\%$ & 26.93 & $39.69\%$ & 10.82 & $24.78\%$ & 19.59 & $36.95\%$ & 17.48 & $34.95\%$ & 30.86 & $40.49\%$ & 19.60 & $36.96\%$ \\
Train (scratch) &  15.05 & 28.91$\%$ &  26.43 & 20.03$\%$ &  23.41 & 21.08$\%$ &  10.77 & 24.02$\%$ &  16.86 & 17.41$\%$ &  15.50 & 19.28$\%$ &  26.52 & 20.30$\%$ &  16.88 & 17.54$\%$  \\
AL$^\varnothing$ & 32.08 & $175.25\%$ & 51.70 & $136.04\%$ & 47.65 & $147.90\%$ & 25.00 & $188.85\%$ & 32.45 & $127.08\%$ & 29.94 & $131.91\%$ & 50.14 & $128.59\%$ & 30.93 & $116.41\%$ \\
AL  &  14.95 & 28.03$\%$ &  25.81 & 17.19$\%$ &  22.70 & 17.33$\%$ &  10.70 & 23.14$\%$ &  16.47 & 14.66$\%$ &  15.07 & 15.98$\%$ &  25.84 & 17.16$\%$ &  16.48 & 14.71$\%$  \\
EAL$^\varnothing$ & 15.62 & $33.98\%$ & 30.36 & $38.36\%$ & 26.93 & $39.69\%$ & 10.82 & $24.78\%$ & 19.59 & $36.95\%$ & 17.48 & $34.95\%$ & 30.86 & $40.49\%$ & 19.60 & $36.96\%$ \\
EAL  & \textbf{13.16} & \textbf{12.70}$\%$ & \textbf{23.88} & \textbf{8.42}$\%$ & \textbf{21.18} & \textbf{9.47}$\%$ & \textbf{9.46} & \textbf{8.85}$\%$ & \textbf{15.18} & \textbf{5.61}$\%$ & \textbf{13.81} & \textbf{6.24}$\%$ & \textbf{23.94} & \textbf{8.54}$\%$ & \textbf{15.17} & \textbf{5.60}$\%$ \\
\bottomrule
\end{tabular}
}
}
\end{table}

\begin{table}[h!]
\centering
\rebuttal{
\caption{\rebuttal{Zero-shot, retraining, and fine-tuning performance on unseen variants with combined multi-depots (MD) and mixed backhauls (MB). "$\varnothing$" denotes models and fine-tuning methods evaluated in zero-shot settings. EAL finetuning maintains the zero-shot performance and performs best overall.}}
\label{tab:eal-finetuning-mb-md-16}
\resizebox{\textwidth}{!}{%
\large %
\renewcommand{\arraystretch}{1.05} %
\begin{tabular}{l | cccccccccccccccc}
\toprule
 & \multicolumn{2}{c}{MDVRPMB} & \multicolumn{2}{c}{MDOVRPMB} & \multicolumn{2}{c}{MDVRPMBL} & \multicolumn{2}{c}{MDVRPMBTW} & \multicolumn{2}{c}{MDOVRPMBL} & \multicolumn{2}{c}{MDOVRPMBTW} & \multicolumn{2}{c}{MDVRPMBLTW} & \multicolumn{2}{c}{MDOVRPMBLTW} \\
\cmidrule(lr){2-3} \cmidrule(lr){4-5} \cmidrule(lr){6-7} \cmidrule(lr){8-9} \cmidrule(lr){10-11} \cmidrule(lr){12-13} \cmidrule(lr){14-15} \cmidrule(lr){16-17} 
Method & Obj. & Gap & Obj. & Gap & Obj. & Gap & Obj. & Gap & Obj. & Gap & Obj. & Gap & Obj. & Gap & Obj. & Gap \\
\midrule
HGS-PyVRP & 10.68 & *      & 7.66 & *            & 10.71 & *         & 19.29 & *         & 7.66 & *          & 12.96 & *         & 19.31 & *         & 12.96 & *         \\
OR-Tools & 12.22 & $14.37\%$ & 8.88 & $15.83\%$  & 12.23 & $14.23\%$ & 22.39 & $16.12\%$ & 8.87 & $15.73\%$  & 14.49 & $11.79\%$ & 22.43 & $16.16\%$ & 14.49 & $11.79\%$ \\
MTPOMO$^\varnothing$  & 15.14 & $42.22\%$ & 10.91 & $42.57\%$ & 15.49 & $45.23\%$ & 28.44 & $48.01\%$ & 10.90 & $42.45\%$ & 18.56 & $43.63\%$ & 28.93 & $50.36\%$ & 18.56 & $43.65\%$ \\
MVMoE$^\varnothing$  & 15.08 & $41.67\%$ & 10.90 & $42.41\%$ & 15.40 & $44.37\%$ & 28.46 & $48.12\%$ & 10.88 & $42.13\%$ & 18.61 & $44.04\%$ & 28.89 & $50.19\%$ & 18.60 & $43.95\%$ \\
\textsc{RF}-POMO$^\varnothing$  & 15.09 & $41.78\%$ & 10.90 & $42.41\%$ & 15.37 & $44.05\%$ & 28.68 & $49.27\%$ & 10.90 & $42.37\%$ & 18.69 & $44.70\%$ & 29.08 & $51.15\%$ & 18.69 & $44.69\%$ \\
\textsc{RF}-MoE$^\varnothing$  & 15.02 & $41.08\%$ & 10.82 & $41.40\%$ & 15.29 & $43.34\%$ & 28.38 & $47.67\%$ & 10.82 & $41.36\%$ & 18.50 & $43.19\%$ & 28.77 & $49.56\%$ & 18.50 & $43.22\%$ \\
\textsc{RF}-TE$^\varnothing$  & 14.99 & $40.80\%$ & 10.77 & $40.67\%$ & 15.28 & $43.27\%$ & 28.43 & $47.93\%$ & 10.76 & $40.62\%$ & 18.49 & $43.14\%$ & 28.80 & $49.69\%$ & 18.50 & $43.17\%$ \\
Train (scratch) &  13.12 & 22.88$\%$ &  9.37 & 22.32$\%$ &  13.24 & 23.72$\%$ &  22.85 & 18.56$\%$ &  9.38 & 22.44$\%$ &  15.13 & 16.75$\%$ &  22.90 & 18.65$\%$ &  15.11 & 16.60$\%$  \\
AL$^\varnothing$ & 34.12 & $223.14\%$ & 26.36 & $245.53\%$ & 27.41 & $158.88\%$ & 48.94 & $155.28\%$ & 24.11 & $216.01\%$ & 31.53 & $144.89\%$ & 46.80 & $143.89\%$ & 30.08 & $133.48\%$ \\
AL  &  13.10 & 22.70$\%$ &  9.36 & 22.14$\%$ &  13.20 & 23.36$\%$ &  22.90 & 18.76$\%$ &  9.38 & 22.46$\%$ &  15.28 & 17.91$\%$ &  23.02 & 19.26$\%$ &  15.39 & 18.77$\%$ \\
EAL$^\varnothing$  & 14.99 & $40.80\%$ & 10.77 & $40.67\%$ & 15.28 & $43.27\%$ & 28.43 & $47.93\%$ & 10.76 & $40.62\%$ & 18.49 & $43.14\%$ & 28.80 & $49.69\%$ & 18.50 & $43.17\%$ \\
EAL  & \textbf{12.70} & \textbf{18.98}$\%$ & \textbf{8.53} & \textbf{11.35}$\%$ & \textbf{12.68} & \textbf{18.56}$\%$ & \textbf{21.41} & \textbf{11.05}$\%$ & \textbf{8.54} & \textbf{11.43}$\%$ & \textbf{13.93} & \textbf{7.41}$\%$ & \textbf{21.44} & \textbf{11.09}$\%$ & \textbf{13.91} & \textbf{7.32}$\%$ \\\bottomrule
\end{tabular}
}
}
\end{table}

\our{} models perform the best in zero-shot generalization across all experiments; moreover, EAL finetuning achieves the same zero-shot performance as the backbone  \our{} model \textsc{RF}-TE thanks to the zero-padded initialization, while AL does not due to the introduction of untrained embedding layers. Notably, experiments with multi-depots are much harder than mixed backhaul variants since they require the model to understand multiple starting (and returning) point locations and to schedule vehicle assignments to their respective depots efficiently. EAL performs the best across all variants in finetuning performance. Remarkably, EAL's performance compared to AL and retraining a model from scratch is more prominent with the increasing difficulty of the finetuning task from MB to MB+MD, indicating it is a suitable method for efficient deployment in finetuning to new tasks.

\subsubsection{Finetuning with EAL for Single-Variant Models}
\label{append:comparisons-single-variant-models-finetuning-performance}

We finetune all POMO models with the same setting as the experiment with unseen mixed backhaul and multi-depots (MB\&MD) from \cref{append:finetune-unseen-variants} with EAL. 

\begin{table}[h!]
\centering
\rebuttal{
\caption{\rebuttal{Fine-tuning performance on unseen variants of single-variant POMO models and \our{}. Finetuning a foundation model for VRPs is crucial for fast adaptation to downstream tasks.}}
\label{tab:pomo-eal-finetuning-mb-md-16}
\resizebox{\textwidth}{!}{%
\large %
\renewcommand{\arraystretch}{1.05} %
\begin{tabular}{l | cccccccccccccccc}
\toprule
 & \multicolumn{2}{c}{MDVRPMB} & \multicolumn{2}{c}{MDOVRPMB} & \multicolumn{2}{c}{MDVRPMBL} & \multicolumn{2}{c}{MDVRPMBTW} & \multicolumn{2}{c}{MDOVRPMBL} & \multicolumn{2}{c}{MDOVRPMBTW} & \multicolumn{2}{c}{MDVRPMBLTW} & \multicolumn{2}{c}{MDOVRPMBLTW} \\
\cmidrule(lr){2-3} \cmidrule(lr){4-5} \cmidrule(lr){6-7} \cmidrule(lr){8-9} \cmidrule(lr){10-11} \cmidrule(lr){12-13} \cmidrule(lr){14-15} \cmidrule(lr){16-17} 
Method & Obj. & Gap & Obj. & Gap & Obj. & Gap & Obj. & Gap & Obj. & Gap & Obj. & Gap & Obj. & Gap & Obj. & Gap \\
\midrule
HGS-PyVRP & 10.68 & *      & 7.66 & *            & 10.71 & *         & 19.29 & *         & 7.66 & *          & 12.96 & *         & 19.31 & *         & 12.96 & *         \\
OR-Tools & 12.22 & $14.37\%$ & 8.88 & $15.83\%$  & 12.23 & $14.23\%$ & 22.39 & $16.12\%$ & 8.87 & $15.73\%$  & 14.49 & $11.79\%$ & 22.43 & $16.16\%$ & 14.49 & $11.79\%$ \\
POMO\_CVRP &   13.34 & 24.97$\%$ &  9.66 & 26.01$\%$ &  13.43 & 25.50$\%$ &  25.19 & 30.84$\%$ &  9.66 & 25.97$\%$ &  25.14 & 30.39$\%$ &  17.66 & 36.50$\%$ &  17.65 & 36.43$\%$ \\
POMO\_VRPL &  13.36 & 25.14$\%$ &  9.88 & 28.97$\%$ &  13.37 & 24.99$\%$ &  28.15 & 46.43$\%$ &  9.86 & 28.70$\%$ &  28.02 & 45.58$\%$ &  20.79 & 60.98$\%$ &  20.74 & 60.53$\%$  \\
POMO\_OVRP &  13.31 & 24.62$\%$ &  9.54 & 24.45$\%$ &  13.35 & 24.77$\%$ &  26.03 & 35.27$\%$ &  9.55 & 24.63$\%$ &  26.03 & 35.07$\%$ &  18.65 & 44.21$\%$ &  18.66 & 44.30$\%$  \\
POMO\_VRPTW &  13.91 & 30.27$\%$ &  10.17 & 32.77$\%$ &  13.99 & 30.72$\%$ &  24.70 & 28.13$\%$ &  10.22 & 33.43$\%$ &  24.78 & 28.43$\%$ &  16.74 & 29.32$\%$ &  16.80 & 29.77$\%$  \\
POMO\_VRPB &  13.00 & 21.69$\%$ &  9.25 & 20.63$\%$ &  13.06 & 22.07$\%$ &  22.50 & 16.66$\%$ &  9.23 & 20.44$\%$ &  22.53 & 16.64$\%$ &  14.96 & 15.39$\%$ &  14.97 & 15.54$\%$  \\
\our{}  & \textbf{12.70} & \textbf{18.98}$\%$ & \textbf{8.53} & \textbf{11.35}$\%$ & \textbf{12.68} & \textbf{18.56}$\%$ & \textbf{21.41} & \textbf{11.05}$\%$ & \textbf{8.54} & \textbf{11.43}$\%$ & \textbf{13.93} & \textbf{7.41}$\%$ & \textbf{21.44} & \textbf{11.09}$\%$ & \textbf{13.91} & \textbf{7.32}$\%$ \\\bottomrule
\end{tabular}
}
}
\end{table}

\cref{tab:pomo-eal-finetuning-mb-md-16} shows that fine-tuning our \our{} foundation model achieves the best results, even when comparing variants that include only unseen features for both. For instance, POMO trained only on VRP with backhauls (POMO\_VRPB in the table) was trained by sampling many more (classical) backhaul features, but \our{} can still fine-tune better on MDVRPMB. Models trained on similar features as the target ones (e.g., POMO\_VRPTW being trained on time windows) can overall fine-tune better to variants that include these features (e.g., time windows) than other models. This is expected, but our foundation model performs even better. This is a strong motivation for practitioners and researchers: developing foundation models for VRPs is crucial for fast adaptation to new tasks that may arise in real-world scenarios, such as adding new constraints or attributes.


\tmlr{\section{Visualizations}}

\tmlr{We provide qualitative visualizations for the solutions of randomly picked instances from the instances of the main results of \cref{tab:main-results} for different models: MTPOMO, MVMoE, and \our{} model variants \textsc{RF}-POMO, \textsc{RF}-MoE, \textsc{RF}-TE.
}

\begin{figure}[htbp]
    \centering
    \includegraphics[width=.83\linewidth]{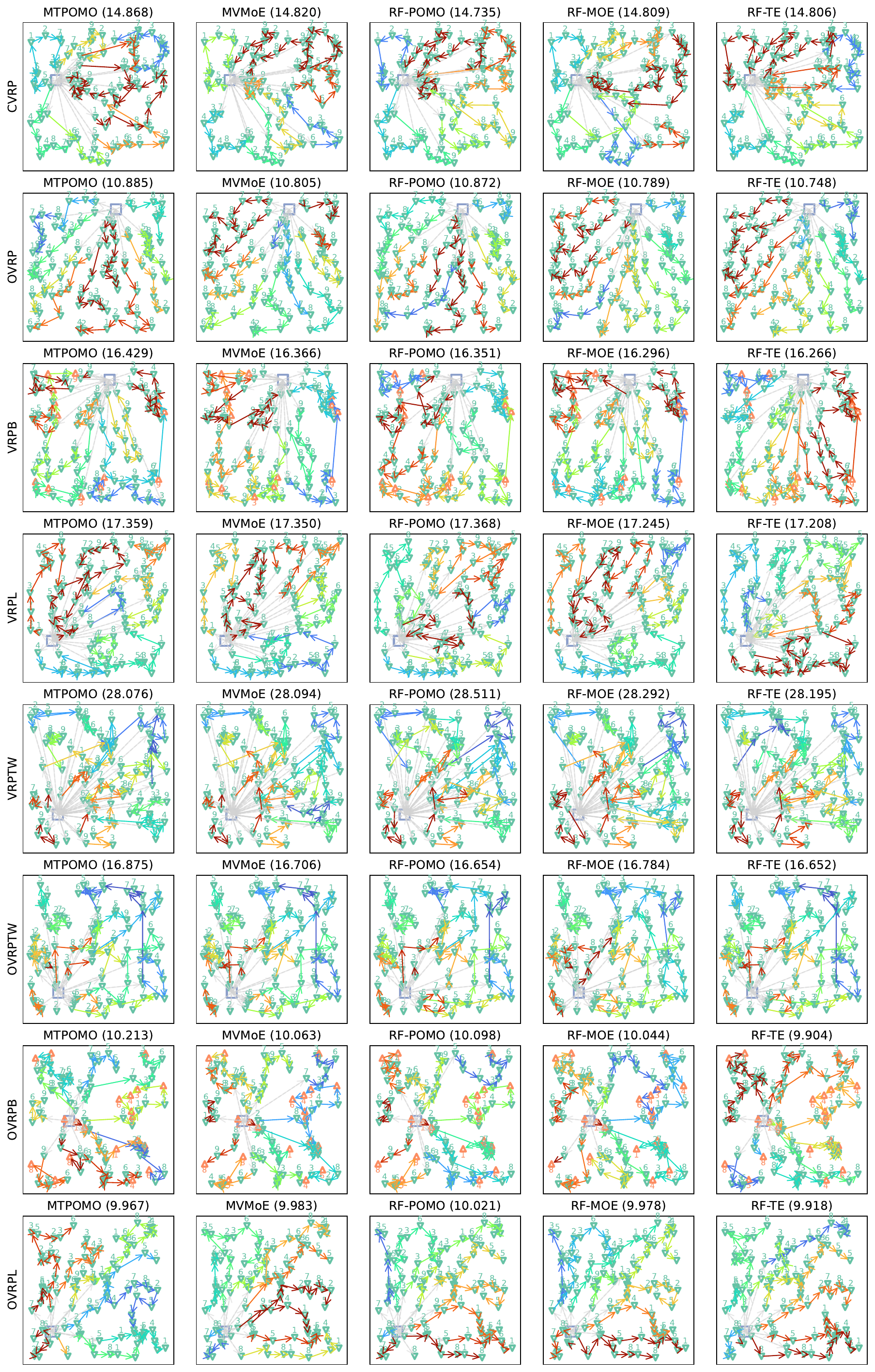}
    \caption{\tmlr{Solutions for CVRP, OVRP, VRPB, VRPL, VRPTW, OVRPTW, OVRPB, OVRPL.}}
    \label{fig:visualization-first-8}
\end{figure}

\begin{figure}[htbp]
    \centering
    \includegraphics[width=.83\linewidth]{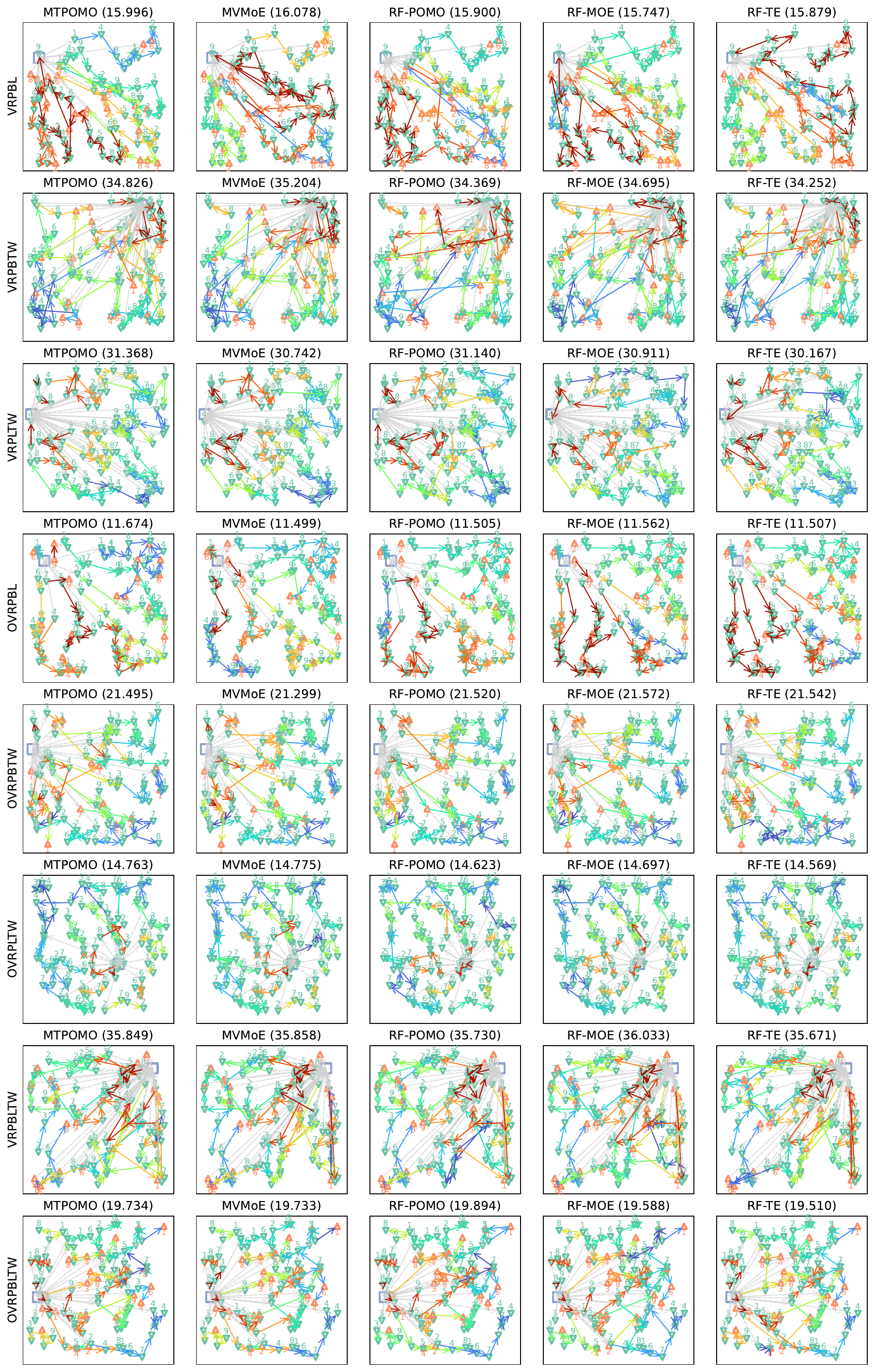}
    \caption{\tmlr{Solutions for VRPBL, VRPBTW, VRPLTW, OVRPBL, OVRPBTW, OVRPLTW, VRPBLTW, OVRPBLTW.}}
    \label{fig:visualization-last-8}
\end{figure}


\end{document}